\documentclass[twoside,11pt]{article}

\usepackage{blindtext}

\usepackage{jmlr2e}
\usepackage{cite,times,mathptm,rotating,subfigure,mathrsfs,bbm,soul,color,amsmath,algorithm,algpseudocode,adjustbox,multirow,booktabs,appendix,placeins}

\newcommand{\be}{\begin{equation}}

\newcommand{\Amat}{\mathbf{A}}

\newcommand{\Imat}{\mathbf{I}}

\newcommand{\zeromat}{\mathbf{0}}
\newcommand{\Rmat}{\mathbf{R}}

\newcommand{\Xmat}{\mathbf{X}}

\newcommand{\av}{\mathbf{a}}

\newcommand{\gv}{\mathbf{g}}

\newcommand{\cv}{\mathbf{c}}

\newcommand{\xv}{\mathbf{x}}

\newcommand{\ee}{\end{equation}}

\usepackage{lastpage}

\firstpageno{1}

\begin{document}

\title{ERICA: Quantifying Replicability of Cluster Analysis}{\,\,\, }

\author{\name Siamak K. Sorooshyari \email siamak@stanford.edu \\
       \addr Department of Statistics\\
       Stanford University\\
       Stanford, CA 94305, USA
      \AND
       \name Manuel A. Rivas \email mrivas@stanford.edu \\
       \addr Department of Biomedical Data Science\\
       Stanford University\\
       Stanford, CA 94305, USA
       \AND
       \name Robert Tibshirani \email tibs@stanford.edu \\
       \addr Department of Statistics and Biomedical Data Science\\
       Stanford University\\
       Stanford, CA 94305, USA}

\editor{\,\,\, }

\maketitle

\begin{abstract}
Despite being ubiquitous in science, clustering lacks a unified framework for quantitatively evaluating the replicability of its results. 
We present evaluating replicability via iterative clustering assignments (ERICA), a method for determining whether clusters can be identified reproducibly in a dataset. 
The pipeline computes a statistic that determines whether reproducible cluster structure is present in a dataset. 
%
Quantitative visualization methods are also introduced to characterize similarities between clusters and identify observations that may represent outliers or unstable assignments. 
Experiments on synthetic datasets demonstrate that ERICA successfully identifies reproducible cluster structure. 
%
In contrast, application of ERICA to three breast cancer gene-expression datasets reveals instances in which clustering solutions are not reproducible. 
The study underscores the importance of rigorously evaluating clustering solutions and provides a practical framework for doing so. 
\end{abstract}


\section{Introduction}

Clustering is a fundamental data mining technique for discovering patterns. 
Its versatility has made it a widely used tool for exploratory data analysis. 
It is frequently used to identify groups of similar observations and to facilitate the interpretation of outcomes and variables within a study. 
Unfortunately, rigorous questions regarding the reliability and interpretation of clustering results are often left unaddressed. 
For instance, how replicable are the findings when observations are subsampled or portions of the dataset are held-out? 
How many groups are estimated with a specified degree of confidence? 
Complicating this further, there is no universally accepted definition of what constitutes a satisfactory clustering result. 
Fortunately, the notion of stability has become a tractable approach for assessing the robustness and reproducibility of clustering solutions. 

We introduce the evaluating replicability via iterative clustering assignments (ERICA) platform for assessing clustering replicability (CR). 
The framework is model-free and is not restricted to a particular clustering algorithm or data type. 
It does not require the use of a classifier on held-out data to test its confidence in the findings. 
We also introduce a simple metric, the ERICA statistic, that measures whether a dataset's clusters can be discovered in a replicable manner. 
The pipeline and ERICA statistic provide scientists with a tool to evaluate the fidelity of structure in their data and the replicability of their findings {\it before} the results are peer-reviewed. 
A high ERICA statistic indicates replicability in the discovered structure, whereas a low value means that the inferred structure is not reliable. 
%

%
%
The proposed pipeline is sufficiently flexible to incorporate multiple clustering algorithms, thereby enabling assessment of consensus across clustering solutions. 
We use k-means and two hierarchical clustering methods that exploit different facets of the data to declare a group of points as being similar. 
A high ERICA statistic, together with consensus among methods regarding the number and identities of clusters, provides evidence for replicable latent structure. 
Conversely, inconsistent assignments across altered versions of the same dataset raise questions about whether a definitive underlying structure exists. 
Such a scenario may have arisen due to experimental issues, artifacts, noisiness of the empirical process, or the lack of an effect. 

We test ERICA on synthetic data where the ground truth is known. The results validate the ERICA pipeline on high-dimensional data with overlapping clusters. 
Three metrics are defined to assess the capabilities of the evaluation platform. 
The first two metrics measure fidelity at the cluster level. 
We average the first metric across clusters to gain a holistic account of the results, and establish the ERICA statistic as a single-number measure of whether stable structure exists in a dataset with a clustering technique. 
Next, we apply the evaluation platform to genomic data from breast cancer tumors and compare the results with outcomes from recent works. 
Our findings differ from those reported in prior studies. 
This suggests a lack of clear structure in the datasets that, if present, would lead to higher reproducibility in the number of groups (i.e. tumor types) discovered in the data. 
This is not surprising in light of the replicability crisis in science. 

There are caveats with our analysis that motivate ongoing and future work. 
For instance, our instantiation of ERICA does not involve a search over the parameter space. 
We have considered a limited number of algorithms and give equal weight to each of their results. 
This is not a comprehensive account because the strategies optimize different objective functions and yield disparate solutions for identical inputs. 
The remainder of the paper is organized as follows. 
In Section 2 we discuss the importance of replicability analysis in clustering and outline a series of questions addressed by ERICA. 
Section 3 introduces the definitions as well as the ERICA statistic. 
We also introduce simple yet elegant methods for visualizing cluster replicability at the dataset, cluster, and individual data-point levels. 
The ERICA algorithm is presented in Section 4. 
The simulation results of Section 5 validate the performance of the evaluation platform on several versions of synthetic datasets where the ground truth is known. This is followed by using ERICA on three gene expression datasets for breast cancer subtype validation. 
We compare the ERICA statistic and the evaluation pipeline to prior works in Section 6 and conclude in Section 7.

\section{A Necessity for Replicability}

The replicability crisis in science has been discussed as an unfortunate phenomenon \citep{ioannidis2005}. 
Replicability is essential for establishing the reliability of scientific findings. 
Relatively recent canonical works such as \citet{baker2016}, \citet{openscience2015}, and \citet{begley2012} have reported that most scientific findings are non-replicable or are 
perceived to lack reproducibility. 
This is concerning since the rate of research progress is strongly dependent on communities quickly integrating and expanding upon findings. 
Algorithmic techniques must be developed to identify results that are not reliable before they are accepted and compromise the reputation of the researcher and the trainees' careers. 
From a funding perspective, alarmingly low rates of replicability within a scientific field cast doubt in the minds of donors, agencies, and governments as to whether grant money should be allocated. 

A large number of decisions are made from unsupervised learning methods that incorporate pattern recognition and clustering. 
The areas are frequently viewed in unison as they rely on identifying groups of objects that are highly similar, and dissimilar to other groups. 
Replicability analysis for such ubiquitous procedures will contribute to addressing the aforementioned crisis. 
Prior to claiming a discovery, it is highly desirable to have a procedure and quantitative metrics that address questions such as whether a dataset contains meaningful structure and whether the resulting findings are replicable. 
%
Studies such as \citet{liu2022stability} and \citet{bendavid2006} have analyzed stability as the best means of immediately determining the validity of a clustering solution. 
Given a dataset, we leverage iterative statistical analysis and present a statistic to assess the replicability of the relationships that may exist. 
Poor replicability in the grouping of observations in a dataset reflects negatively on the expected effect, hypothesis, experiment, or the data collection. 
Without additional experimentation or data collection, a researcher would reassess the aforementioned factors. 
This will serve as a preventive measure for the introduction of data and conclusions that may not be repeatable by other researchers. 

\subsection{Replicability Analysis in Clustering}

The absence of a ground truth in unsupervised learning leads to difficulties in assessing the goodness of an algorithm. 
Replicability has been studied as a means of evaluating the efficacy of conjectured structure in data \citep{liu2022stability, bendavid2006, hennig2007}. 
It is necessary but not sufficient in assessing whether the findings are reliable. 
If the results are to be taken seriously, the consistency of discovered relationships should hold with relatively minor changes to the dataset. 
From an algorithmic angle, recent works have suggested different measures of similarity between partitions of the data to quantify replicability \citep{masoero2023}. 
%
%
Improving robustness in clustering solutions will help address aspects of the replicability crisis.
The rationale follows that reliable relationships among observations will persist despite perturbations of the variables that comprise the observations or several of the observations themselves. 
It is therefore valuable to have a statistic that quantifies the reproducibility of clustering results. 

\subsection{Verification methods for clustering integrity}

The bootstrap (BS) is perhaps the most utilized technique to study the confidence in the output of a clustering algorithm \citep{jain1987bootstrap}. 
Monte Carlo techniques have also been employed to develop the notion of a gap statistic (GS) \citep{tibshirani2001gap}, while k-fold cross-validation has been used for estimating the number of groups via the prediction strength (PS) \citep{tibshirani2005prediction}. 
There exists a limited discussion on the appropriateness of bootstrapping versus Monte Carlo subsampling (MCSS) for evaluating clustering \citep{moeller2006, mucha2015}. 
Sampling with replacement is often considered inappropriate for evaluating whether a finite dataset with an unknown distribution contains well-defined groups. 
Works such as \citet{dudoit2002}, \citet{abul2003}, and \citet{lange2004} have considered the combination of sampling without replacement (WOR) and the use of classifiers to evaluate clustering results. 

We believe MCSS to be more applicable to our formulation and view bootstrapping as an extension of this work since the BS is a specific case. 
The main thrust of the pipeline will be more sensible for sampling WOR and thus keeping a portion of the dataset as "held-out." 
In \citet{dangl2020} data splitting, bootstrapping, and subsetting were applied to cluster data from a finite mixture of multivariate Gaussians. 
It was noted that data splitting, where there is no overlap in the samples used for determining the clusters and validation of the clusters, provided a more reliable estimate of the number of groups than bootstrapping or subsetting the data into non-disjoint portions. 
The analysis was restricted to k-means clustering. 
The authors in \citet{lange2004} also deemed BS as not appropriate for clustering assessment and subsampling was used instead. 
Another motivation for our use of MCSS is a methodology used in \citet{masoero2023}. 
The authors referred to their technique as a BS, whereas their use of sampling WOR more closely resembles MCSS.

\subsection{Goals of a replicability evaluation platform}

At the end of an analysis it is necessary to have quantitative metrics answer a series of questions. 
The first and most important is whether a dataset has a stable clustering. 
In other words, do there exist groups of points that are more similar to each other than to those in other groups? 
Different clustering algorithms will provide very different results when provided with the same dataset. 
It is important to be aware of this by not discounting the lack of structure in light of results from one technique. Conversely, it may be misleading to base the findings of a study on the results of a single technique. 
Thus, a second question is whether there is consensus reached about the dataset's clustering across techniques. 
The most natural third question is what is the most likely number of groups. 
Clustering is not a well-defined problem since the definition of a cluster is ambiguous, and there is usually more than one credible answer to how many groups exist in a dataset. Although this is not unanimously accepted, it is increasingly being acknowledged \citep{jain2010, masulli2015}. 

\begin{figure}[h]
\begin{center}
\epsfig{figure=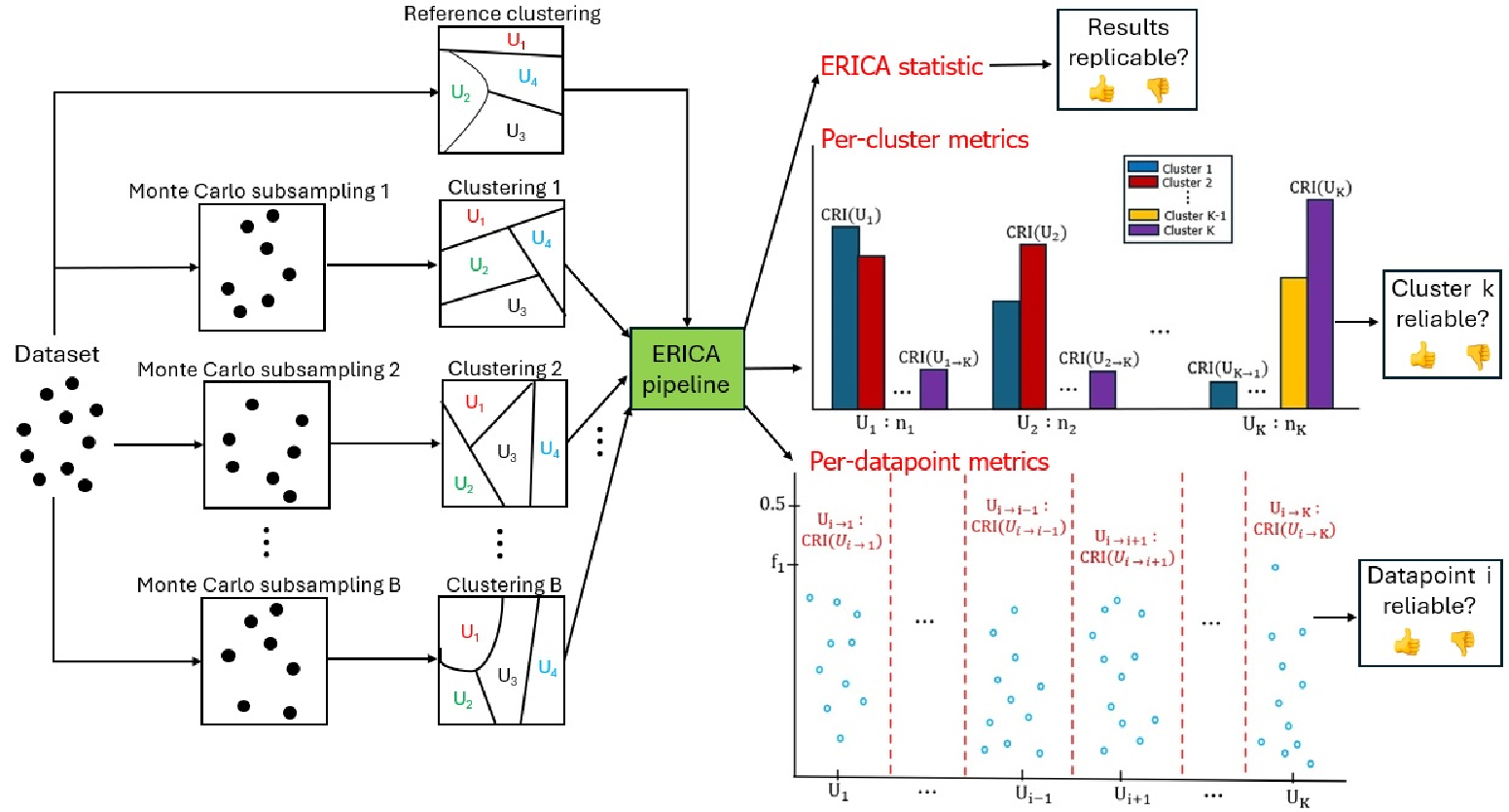, height=3.4in}
\end{center}
\caption{A high-level view of the ERICA platform. Data is clustered in the uppermost branch to form centroids that will serve as a reference for the cluster identities. The dataset also undergoes $B$ iterations of MCSS with the results being grouped via the same clustering technique. In the ERICA pipeline, the identities of the grouped data points across the $B$ iterations are registered with respect to the reference and collected to form the CLAM. The matrix is then processed to obtain metrics that quantify the replicability in cluster assignments.} 
\label{highlevel_fig1}
\end{figure}

The three questions follow each other naturally and are fundamental for the assessment of a replicability study. 
Nevertheless, further crucial inquiries remain. 
A fourth item to investigate is which groups (if any) are most confidently believed to exist in the dataset. 
We anticipate that the answer will be valuable to an experimentalist. 
Rather than treating cluster discovery as a binary outcome, it is insightful to also have a measure of its fidelity. 
A fifth question stems from further interrogating the prospective relationships in the dataset by quantifying which clusters are more similar or "closer" to each other. 
This information is valuable when assessing whether an experiment has preserved the relationships among clusters. 
Alternatively, this information may constitute a discovery and lead to new findings. 
In a clinical trial, noting two groups as more similar to each other than to constituents may foreshadow the subjects in the second group experiencing the same outcome as subjects in the first group. 
Lastly, it is important to investigate how the results have been affected by individual data points. Namely, identifying the data points that do not definitively belong to a group. 
The answer may identify observations that are artifacts and should be discarded. 
Alternatively, such data points may be important rare-event cases that have arisen because of the protocol, instrumentation, or unknown factors and require a separate investigation. 
The six questions are formalized as follows:\\ 
\makebox[0.5em][l]{} 1) Can the same clusters be consistently found in the dataset?\\
\makebox[0.5em][l]{} 2) Is there consensus among the methods in the assignment of observations to clusters?\\
\makebox[0.5em][l]{} 3) What is the most likely number of clusters?\\
\makebox[0.5em][l]{} 4) Which clusters exhibit the highest replicability?\\
\makebox[0.5em][l]{} 5) What clusters share the greatest level of similarity?\\ 
\makebox[0.5em][l]{} 6) Which data points most definitively belong to a cluster, or alternatively do not clearly belong to any cluster?  
\\
The questions proceed from the largest to the smallest scale. 
They first scrutinize the entire dataset for the presence of structure, then examine the degree and properties of the structure, and progress to the assessment of individual observations that may be anomalous with respect to the identified groups. 
While this progression is by design, it does not need to be followed when using the ERICA evaluation platform. 
Figure \ref{highlevel_fig1} provides an overview of the components and steps involved in the presented analysis. 
Each component will be expanded upon in the subsequent sections as we introduce the relevant symbols, metrics, and visualizations. 

\section{A Replicability Evaluation Platform}

The purpose of the ERICA pipeline is to evaluate CR for a dataset. 
This is not equivalent to the more ambitious and often infeasible task of assessing clustering accuracy. 
In fact, correctness is not well-defined in this context due to an absence of ground truth. 
Nevertheless, CR is a necessary but not sufficient condition for pattern recognition \citep{bendavid2006, vonluxburg2010}. 
%
We are not solely concerned with model selection; rather, we seek to answer fundamental questions about the dataset. 
The definitions and metrics presented below enable us to answer these questions. 
Specifically, a cluster assignment matrix (CLAM) is introduced as a computed entity for evaluating the replicability of a grouping. 
By considering several clustering strategies that have diverse objectives, we investigate whether the data contain structure that can be identified reproducibly. 

\subsection{CLAM definitions and the ERICA statistic}

Before presenting the definitions, we describe the pipeline in Figure \ref{highlevel_fig1}. 
The top branch of the pipeline takes a dataset $\mathcal{X}$ and clusters it in its entirety. 
This provides the reference cluster identities used later in the registration process. 
Then, a random but large portion of the dataset is clustered to form clustering boundaries as shown in Figure \ref{highlevel_fig1}. 
The held-out data points are then assigned to the clusters that were formed. 
This process is repeated independently $B$ times. 
During each trial, we record the cluster identity assigned to every left-out data point. 
At this stage, we count across the $B$ trials how many times each data point was assigned to each of the $K$ clusters. 
%
%
Under a stable clustering solution, data points will be assigned predominantly to the same cluster across trials. 
Conversely, a dataset with clusters that are poorly distinguished or weakly separated will have data points with assignments that are distributed more uniformly across clusters. 
This is the premise of the ERICA evaluation platform. 

To formalize, a fraction of $\mathcal{X}$ is sampled without replacement and used as the input to a clustering algorithm. 
The resultant cluster boundaries or grouping rule is applied to the held-out (i.e. unselected) samples from $\mathcal{X}$.
The identities of the held-out data points and the clusters that they have been assigned to are known and recorded by incrementing an appropriate entry in the $t \times K$ CLAM. 
Our notation will largely follow that of \citet{masoero2023} since it has adopted a lucid, general presentation. 
Define the partitioned dataset as $\mathcal{X} = \left[\Xmat, \Xmat' \right]$ with $t$ data points, where the columns are partitioned via $\Xmat = \left[\xv_{1}, \ldots, \xv_{n} \right]$ and $\Xmat' = \left[\xv'_{1}, \ldots, \xv'_{m} \right]$ with $t=n+m$. 
The columns of $\Xmat$ are $p$-dimensional observations that comprise $P\%$ of the $t$ data points in $\mathcal{X}$. 
The columns of $\Xmat'$ may be part of the collected data, new data points, or perturbed versions of $\xv_{i} \in \Xmat$.  
For observation $\xv_i$, the index $I_i$ will be used to mark its global identity within the dataset, and the function $\mathcal{I}(.)$ returns the identity of the data point. 
In other words, for $\xv_{i} : i=1, \ldots, t$ we have $\mathcal{I}(\xv_{i}) = I_{i}$. 

A crucial variable $K$ will be used to denote the number of clusters declared by a clustering technique $\mathcal{A}$. 
This is one of the most important facets of the analysis since all downstream results depend on the assignment. 
The properties of a clustering function $\psi(.;\mathcal{A}, \Xmat)$ have been defined in \citet{masoero2023}. 
With $[k] \triangleq \{1, \ldots, k\}$, we review that $\psi(.;\mathcal{A}, \Xmat) : \mathbb{R}^{p} \rightarrow [K]$ is a learned operation that is shaped by $\Xmat$. 
The clusters of data points will be denoted by $U_{1}, \ldots, U_{K}$ and the assignment of $\xv_{i}$ to $U_j$ will be denoted via $I_{i} \rightarrow U_{j}$. 
Thus, $U_{1}, \ldots, U_{K}$ are disjoint subsets that form a partition of $\Xmat$. 
These subsets are then updated via $U_{i} \leftarrow U_{i} \cup \{\xv' \in \Xmat' : \psi(\xv'; \mathcal{A}, \Xmat) = i\}$ to contain the data points in $\Xmat'$ that have been assigned to the same partition and thus have the same cluster label. 
In the identification of clusters in a space, $\cv_{k}$: $k=1, \ldots, K$ will be the centroid of the set $U_{k}$. 

We define $\Amat(i,j)$ as the number of times that data point $I_{i}$ has been assigned to cluster $U_{j}$ over $B$ independent trials. 
In terms of our notation this equates to $\#\{I_{i} \rightarrow U_{j} \}$ for $i=1, \ldots, t$ and $j=1, \ldots, K$ over $B$ independent trials. 
The entries $\{\Amat(i,j)\}$ will form the $t \times K$ matrix $\Amat$ that will be referred to as the CLAM. 
For $i= 1, \ldots, t$ data points we designate 
\be
{\rm Sum}(i) = \sum_{j=1}^{K} \Amat(i,j) 
\label{summet1}
\ee
with
\be
{\rm Max}_{c}(i,1) = \underset{j}{\mathrm{max}} \,\, \Amat(i,j) \quad {\rm and} \quad {\rm Idx}_{c}(i,1) = \underset{j}{\mathrm{argmax}} \,\, \Amat(i,j). 
\label{maxmet1}
\ee
Equation (\ref{summet1}) gives the number of times that an observation is assigned to any of the $K$ clusters. 
It should be apparent that $0 \leq {\rm Sum}(i) \leq B$. 
The importance of (\ref{maxmet1}) lies in identifying the maximum number of times that data point $i$ is assigned to a cluster and the identity (i.e. index) of the corresponding cluster, respectively. 
The computation of
\begin{eqnarray}
{\rm Max}_{c}(i,2) &=& \underset{j \, \neq \, {\rm Idx}_{c}(i,1)}{\mathrm{max}} \Amat(i,j) \nonumber \\
{\rm Idx}_{c}(i,2) &=& \underset{j \, \neq \, {\rm Idx}_{c}(i,1)}{\mathrm{argmax}} \Amat(i,j) 
\end{eqnarray}
is also important while noting that, more generally, for $k=1,2 \ldots, f$ we have
\begin{eqnarray}
{\rm Max}_{c}(i,k) &=& \underset{j \, \neq \, {\rm Idx}_{c}(i,1), \, \ldots, \, {\rm Idx}_{c}(i, k-1)}{\mathrm{max}} \Amat(i,j) \label{MaxandIdxeq1} \\ 
{\rm Idx}_{c}(i,k) &=& \underset{j \, \neq \, {\rm Idx}_{c}(i,1), \, \ldots, \, {\rm Idx}_{c}(i, k-1)}{\mathrm{argmax}} \Amat(i,j) . 
\label{MaxandIdxeq2}
\end{eqnarray}
In the above $f \leq K$ is viewed as a termination point on the number of groups that an observation is assigned to. 
We use $\{{\rm Max}_{c}(i,k)\}$ to compute the entries of a $t \times K$ normalized sorted frequency matrix $\Amat_{S}$, and specify the index matrix $\Amat_{I}$ via
\be
\Amat_{S}(i,k) = \frac{{\rm Max}_{c}(i,k)}{{\rm Sum}(i)} \quad  {\rm and} \quad  \Amat_{I}(i,k) = {\rm Idx}_{c}(i,k). \label{EqAsandAimatrix1}
\ee
%
At this point, the analysis requires defining the sets
\begin{eqnarray}
C_{k,k} &=& \{ i : \Amat_{I}(i,1) = k \} \label{Eqsetckk1} \\ 
V_{k,k} &=& \{\Amat_{S}(i \in C_{k,k}, 1) \} \quad {\rm for} \quad k=1, \ldots, K . \label{Eqsetvkk1} 
\end{eqnarray} 
The sets $C_{k,k}$ contain the indices of the data points that are assigned to cluster $k$ the most number of times across $B$ MCSS trials. 
Similarly, the sets $V_{k,k}$ contain the normalized frequencies of the points that have been assigned to cluster $k$ the most number of times. 
We use $X_k$ to denote the count of data points assigned the most number of times to cluster $k$ - it should be apparent from the notation that $X_{k} = |C_{k,k}| = |V_{k,k}|$. 
Now consider the points assigned the most number of times to $U_k$, and for $k,j = 1, \ldots, K, k \neq j$ define the sets 
\begin{eqnarray}
C_{k,j} &=& \{\Amat_{I}(i,j) \,\, | \,\, i \in C_{k,k}, \Amat_{I}(i,j) \neq 0 \} \label{Eqsetckj1} \\ 
V_{k,j} &=& \{\Amat_{S}(i,j) \,\, | \,\, i \in C_{k,k}, \Amat_{I}(i,j) \neq 0 \} \label{Eqsetvkj1} 
\end{eqnarray}
as the indices of those data points that have also been assigned to $U_j$, and the frequencies at which they have been assigned to $U_j$, respectively. 
These definitions will be instrumental in our presented statistic, the operation of ERICA, and visualizations of replicability. 

A number of metrics have already been used to assess the replicability of a clustering technique. 
We define the clustering replicability index (CRI) for cluster $k$ as 
\be
{\rm CRI}(U_{k}) = \frac{\sum_{x \in V_{k,k}}^{} x}{X_{k}} . 
\label{CRIEq1} 
\ee
Thus, of the points that have been assigned the most number of times to $U_k$, CRI$(U_{k})$ represents the fraction of times they have been assigned to $U_k$. 
On a per-cluster basis, this metric quantifies how consistently the data points assigned to a cluster are reassigned to the same cluster across MCSS trials. 
We note that $1/K \leq CRI(U_{k}) \leq 1$, and complete stability for the $k$th group would entail $CRI(U_{k})=1$. 
\begin{definition}[ERICA Statistic]
For a dataset $\mathcal{X}$ and clustering technique $\mathcal{A}$ we present
\be
{\rm CRI} \,\, = \,\, \frac{1}{K} \sum_{k=1}^{K} {\rm CRI}(U_{k}) 
\label{WCRIEq2a} 
\ee
as a measure of confidence in a clustering result. 
The constituent values $\{CRI(U_{k})\}$ provide per-cluster resolution. 
This quantity will be referred to as the ERICA statistic. 
\end{definition}
\noindent
It is not difficult to observe that $\max \{\rm ERICA \,\, statistic\} = 1$ corresponds to complete stability for all data points across the MCSS iterations. 
It is natural to use the pipeline to characterize the "closeness" or similarity among groups. 
We define 
\be
{\rm CRI}(U_{k \rightarrow j}) = \frac{\sum_{x \in V_{k,j}}^{} x}{X_{k}} 
\label{Eqspilov1}
\ee
to quantify the fraction of times that the points, which have been assigned the most number of times to $U_{k}$, are assigned to $U_{j}: j \neq k$. 
In effect, this is viewed as a spillover between the canonical clusters or groups believed to exist in the dataset. 
By construction we note that
\be
{\rm CRI}(U_{k}) + \sum_{j \neq k}^{} {\rm CRI}(U_{k \rightarrow j}) = 1. 
\nonumber
\ee
The spillover between $U_{i}$, $U_{j}$, and $U_{k}$ reflects the closeness between the groups with $CRI(U_{i \rightarrow j}) > CRI(U_{i \rightarrow k})$ indicating that cluster $i$ is believed to be more similar to $j$ than to $k$. 
In measuring replicability on a per-cluster basis $CRI(U_{i}) > CRI(U_{j})$ indicates that $U_i$ contains data points that are more separated from other clusters, whereas $U_j$ contains points that overlap with other clusters. 
In comparing the metrics to baseline values that would be attained under the null condition of the unlabeled data having no structure, it is useful to consider a worst-case analysis. 
If each observation is randomly assigned to a cluster, we shall have a uniform distribution of points allocated to groups with $(\ref{MaxandIdxeq1})$ providing 
\be
{\rm Max}_{c}(i,1) = {\rm Max}_{c}(i,2) = \ldots = {\rm Max}_{c}(i,K) = \frac{B}{K} . 
\nonumber
\ee
This would also result in $\Amat_{S}(i,k) =1/K$ in (\ref{EqAsandAimatrix1}) and $X_{k}= t/K$ since each of the $t$ data points have been assigned uniformly to $K$ clusters across the $B$ MCSS iterations. 
The above assumes that each observation has appeared $B$ times during the pipeline. 
While this may not occur in practice, it does not invalidate the worst-case analysis. 
The uniform assignment will yield CRI($U_{k})=1/K \,\, \forall \,\, k$ and an ERICA statistic of $1/K$ indicating that a clustering has not found replicable structure in a dataset. 
This will occur either because the dataset lacks separable groups or the clustering algorithm has failed to identify existing structure. 

\subsection{Additional metrics for evaluating CR}
 
In a similar manner, a weighted CRI (WCRI) metric is defined as
\be
{\rm WCRI}(U_{k}) = \frac{X_{k}{\rm CRI}(U_{k})}{t}. 
\label{WCRIEq1} 
\ee
The weighted metric takes into account the fraction of the $t$ data points that have been most frequently assigned to a given group. 
Thus, clusters that are assigned a greater number of points at a higher frequency than other groups would exhibit larger WCRI values. 
For a holistic account of whether the dataset contains reproducible clusters, the mean replicability across the identified groups is computed via
\be
{\rm WCRI} = \frac{1}{K} \sum_{k=1}^{K} {\rm WCRI}(U_{k}) 
\label{WCRIEq2b} 
\ee
where $\max \{WCRI\} = 1$ corresponds to complete stability for all data points across the MCSS iterations. 
Note that the above is a special case of 
\be
{\rm WCRI^{*}} = \frac{1}{K}\sum_{k=1}^{K} p_{k} X_{k}
\ee
with the natural constraints that $\sum_{k=1}^{K} X_{k} = t$ and $0 \leq p_{k} \leq 1$. 
Lastly, the total WCRI (TWCRI) is defined as the non-cluster-normalized sum ${\rm TWCRI}(U_{k}) = K \times {\rm WCRI}(U_{k})$. 
Similar to the ERICA statistic, it is evident that higher values for the metrics reflect a higher replicability of assignment. 
The uniform assignment will yield WCRI$(U_{k})=1/K^{2} \,\, \forall \,\, k$, WCRI $=1/K^{2}$, and TWCRI $=1/K$. 
%
These values serve as useful baselines when deciding whether a clustering solution exhibits meaningful replicability. 
The ERICA statistic and the other metrics are relatively easy to compute and model-free because they do not require distributional assumptions on the data. 
However, they are not exhaustive and another goal of the presentation is to scrutinize their efficacy under different conditions. 

\begin{figure*}[!h]
\centering
\subfigure{
\includegraphics[width=0.60\linewidth]{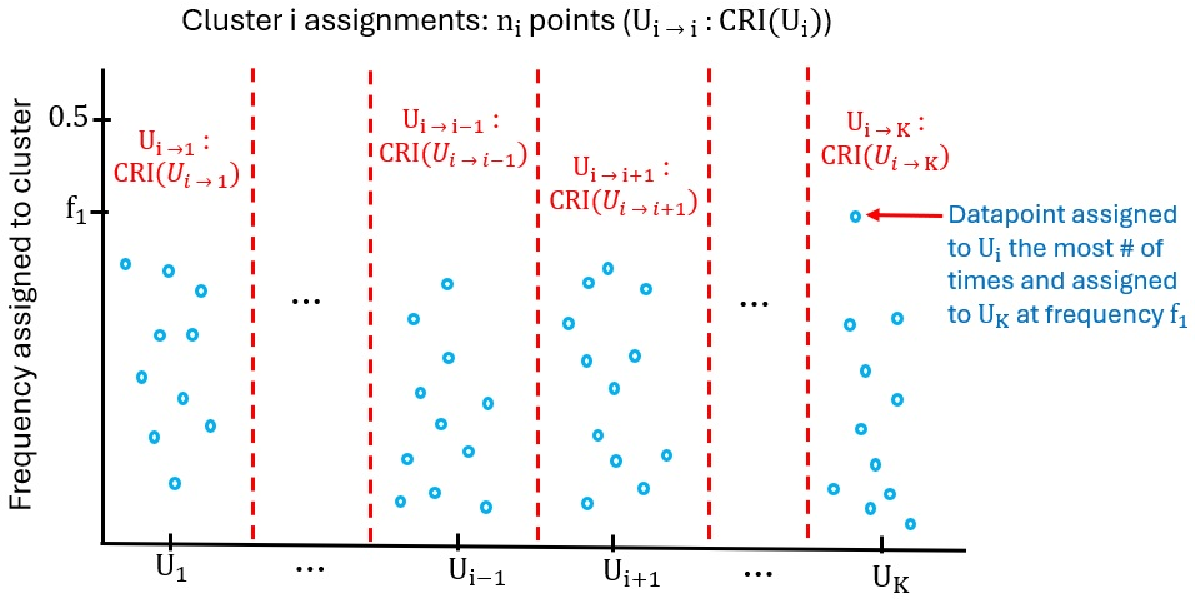}
}
\subfigure{
\includegraphics[width=0.72\linewidth]{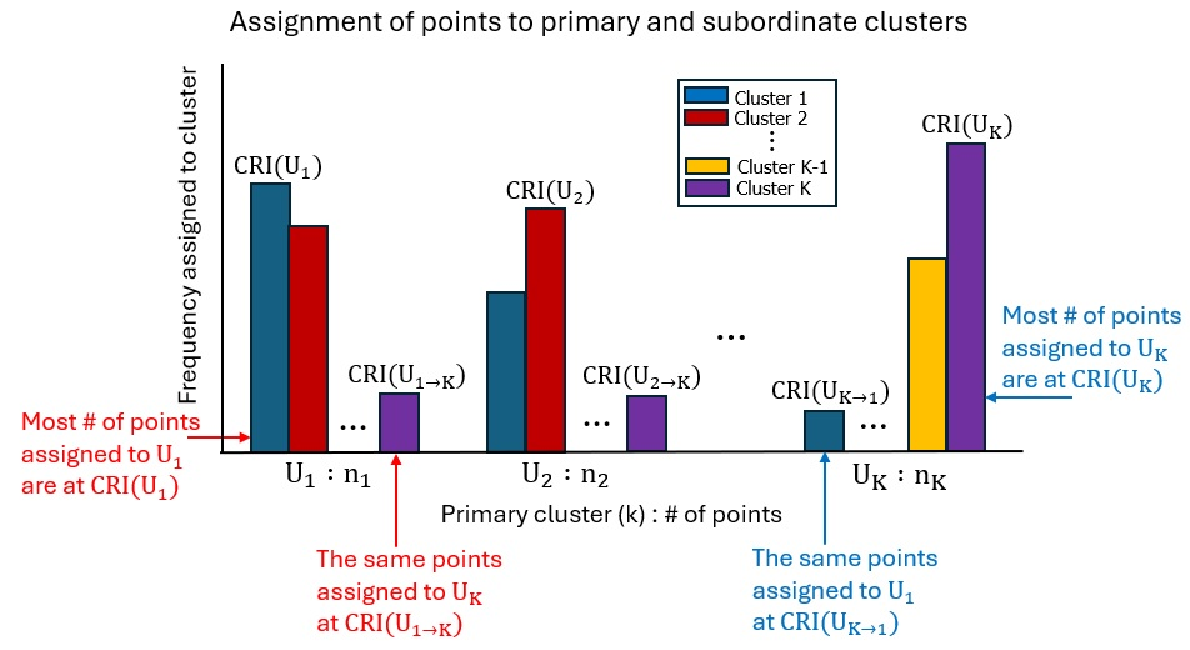}
}
\caption{The ERICA pipeline provides visual outputs of replicability at the level of individual clusters and data points. 
In the PCSP (top), the abscissa represents bins corresponding to the clusters $U_{k} : k \neq i$, to which the data points were assigned over the $B$ MCSS iterations. The ICAH (bottom) provides a cross-cluster summary of the replicability as well as the closeness of the clusters in terms of the rate at which the data points that are primarily assigned to them appear in other groups.}\label{figPCSPICAHclust1}
\end{figure*}

\subsection{Visualizing the results}

The notion of similarity may be subjective or diminished when it is condensed into a few metrics, thus illustrations are a necessary aid for scientists and experimentalists. 
While the structure and properties of a dataset are crucial to quantify, it is also important to provide qualitative accounts. 
Specifically, 
we seek to visualize which groups are most similar to each other and the identity of the data points that are frequently assigned to a cluster. 
We introduce a per-cluster scatter plot (PCSP) in Figure \ref{figPCSPICAHclust1} (top) as a means of showing the association of every data point that has been primarily assigned to cluster $i$ but is also believed to belong to cluster $j$. 
It is common for unstable points in a dataset to sit at the boundary of overlapping clusters \citep{liu2022stability}. Such points are not necessarily outliers, and are visualized in a PCSP by having relatively high frequencies in bins that correspond to secondary clusters $j \neq i$. 
The closeness of the point believed to belong in $U_i$ to group $j$ is quantified in the PCSP. 
For $U_i$, Figure \ref{figPCSPICAHclust1} (top) shows data points in the set $C_{i,i}$ with the frequency that they are assigned to $U_{j} : j \neq i$.  
This is a means of isolating observations (on a per-cluster level) and their tendency to be associated with other clusters. 
In a PCSP, the abscissa represents bins for clusters $U_{j} : j \neq i$. 
Within a bin, the abscissa of the point has no relevance, but the ordinate is the frequency of the assignment. 
Each point corresponds to a unique $p$-dimensional observation. 
By construction, no data point in a PCSP will exceed a value of 0.5 since that is the highest frequency that will be assigned to a cluster that is not its primary assignment. 
While Figure \ref{figPCSPICAHclust1} shows the 
CRI metric ${\rm CRI}(U_{i})$ and the spillover values ${\rm CRI}(U_{i \rightarrow j})$ of $U_{i}$ as summary parameters, it is important to note that in general $CRI(U_{i \rightarrow j}) \neq CRI(U_{j \rightarrow i})$. 
\begin{figure}[h]
\begin{center}
\epsfig{figure=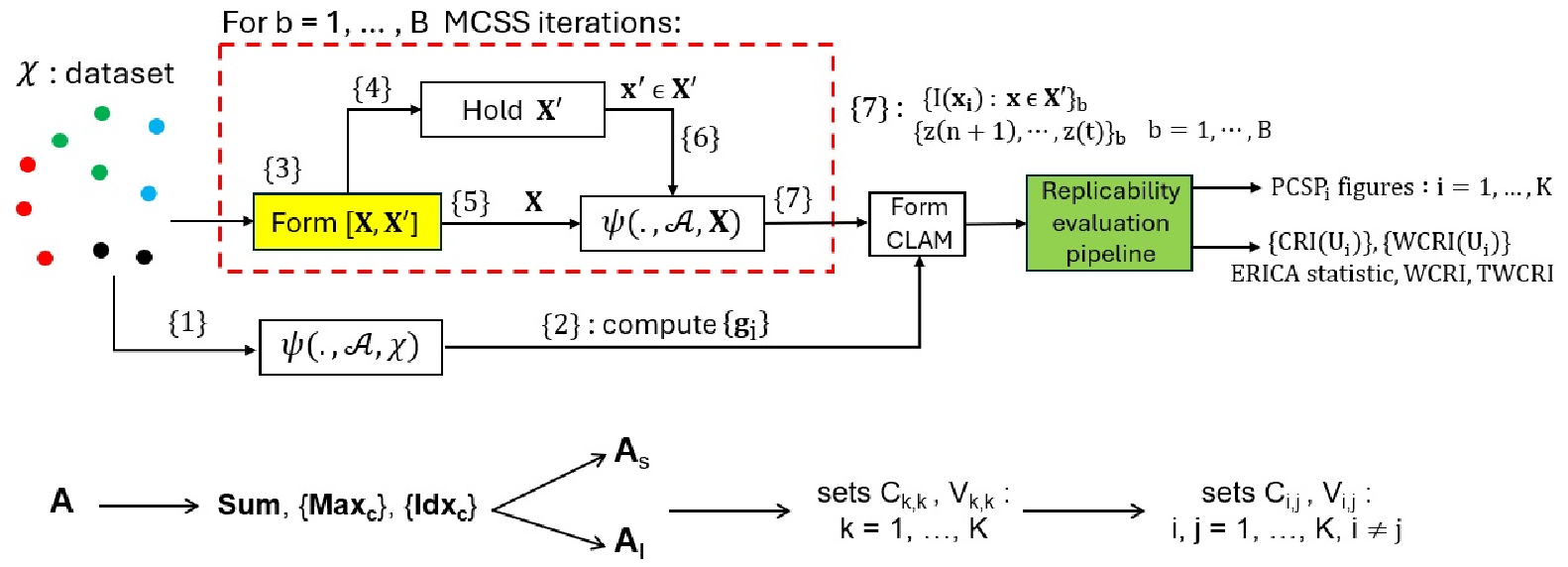, height=2.3in}
\end{center}
\caption{A schematic of the ERICA computational workflow for dataset $\mathcal{X}$ and clustering technique $\mathcal{A}$. Beginning with dataset $\mathcal{X}$, labels \{1\}-\{7\} represent the sequence of operations (top). The flow of processing in the replicability evaluation pipeline (bottom).} 
\label{systemmod_fig1}
\end{figure}

We also present the inter-cluster assignment histogram (ICAH) as a cluster-level summary of the frequency of inter-group assignments. 
The ICAH contains $K$ groups of bar plots, each consisting of $K$ bars. 
More specifically, the $i$th group will show $CRI(U_{i \rightarrow 1}), \ldots, CRI(U_{i}), \ldots, CRI(U_{i \rightarrow K})$, and the primary cluster for each group will have the largest CRI value. 
The ICAH provides the cross-cluster summary of the replicability seen in the assignments, as well as the closeness of the clusters in terms of the rate at which the data points that are primarily assigned to them appear in other clusters. 
%
%
The abscissa reports the number of data points most frequently assigned to each cluster. 
The assignments satisfy the natural relation $n_{1} + n_{2} +\ldots + n_{K} = t$. 
While Figure \ref{figPCSPICAHclust1} shows a PCSP and ICAH made with the CRI metric, analogous figures are possible for the WCRI metric by applying the scalings via $(\ref{WCRIEq1})$. 
%
The dynamics of ERICA are demonstrated via an illustrative example in Appendix A that consists of $t=10$ and $p=2$. 
The points in the dataset $\mathcal{X}$ are assumed to belong to the four clusters shown in Figure \ref{toyexamplefig1a}, and the choice of $K=4$ has been made in $\mathcal{A}$. 
The resulting four PCSPs and ICAH are shown in Figure \ref{figToyComposite1} after $B=8$ MCSS iterations.

\section{Cluster Determination and Registration}

The illustrative example in Appendix A assumes a clustering assignment as the starting point along with the identity of the clusters maintained across the MC iterations. 
Naturally, $\mathcal{A}$ imposes restrictions in terms of its required parameters as well as the information that it will provide.\footnote{Additional information may be provided by $\mathcal{A}$ in the case of other clustering techniques that may be incorporated in future versions of ERICA. For example, affinity propagation will provide exemplars.} 
Additionally, cluster registration is necessary because we must have a consistent reference to identify clusters across MC iterations. 
These are two crucial facets of the ERICA evaluation platform and are discussed below as part of Algorithm 1. 
A detailed view of the components and sequence of steps encompassed by the algorithm is provided by the top diagram in Figure \ref{systemmod_fig1}. 
The steps \{1\} and \{2\} entail clustering the entire dataset and computing $K$ reference groups for the registration portion of the evaluation platform. Steps \{3\} and \{4\} encompass the subsampling, and step \{5\} consists of obtaining the clusters by inputting $\Xmat$ to the clustering $\mathcal{A}$. Subsequently, \{6\} applies the clustering function $\psi(.,\mathcal{A}, \Xmat)$ to the held-out data $\Xmat'$. The outputs at step \{7\} are used to form a CLAM that is then input to the replicability evaluation pipeline. The detailed computations are provided via Algorithm 1. The learned function $\psi$ varies across the $B$ MCSS iterations and differs from the function used to cluster the entire dataset at step $\{1\}$. Nevertheless, to reduce the notation, we do not use the more rigorous $\psi_{b}(.,\mathcal{A}, \Xmat)$ and $\psi^{R}(.,\mathcal{A}, \mathcal{X})$ with "R" denoting reference. 
This pipeline will be applied to each dataset considered in this work. 
The bottom diagram in Figure \ref{systemmod_fig1} illustrates the flow of processing used to derive and utilize the various quantities in the replicability evaluation pipeline. 
\begin{algorithm}
\caption{Evaluating Replicability via Iterative Clustering Assignments (ERICA).}\label{Pipeline1}
\begin{algorithmic}[1]
\State \textbf{Input:} Dataset $\mathcal{X} = \{\xv_{i} : i = 1, \ldots, t\}$, clustering technique $\mathcal{A}$, CLAM matrix $\Amat = \zeromat_{t \times K}$.
\State Cluster $\mathcal{X}$ with $\mathcal{A}$ 
\State \textbf{Compute:} Centroids for the $K$ clusters $\{\gv_{i} \}$. 
\State \textbf{Compute:} $L_2$-norms of the cluster centroids, $g_{i} \triangleq ||\gv_{i}||_{2} : i=1, \ldots, K$. 
\State Sort cluster indices based on centroid norms s.t. for $U_{i}$: $g_{i} \leq g_{j}$ with $j \in \{i+1, \ldots, K\}$. 
\For{$b = 1$ to $B$}
    \State Take a random $P \%$ of data points to be the columns of $\Xmat$ where $\mathcal{X}=\left[\Xmat, \Xmat' \right]$
    \State Cluster $\Xmat$ with $\mathcal{A}$ to attain $\psi(.; \mathcal{A}, \Xmat)$ and $U_{1}, \ldots, U_{K}$ with $U_{i}=\{\xv \in \Xmat : \psi(\xv; \mathcal{A}, \Xmat) = i\}$
    \State \textbf{Compute:} Centroids $\{\cv_{1}, \ldots, \cv_{K}\}$ of the clusters. 
    \State Apply the learned clustering function $\psi(.;\mathcal{A}, \Xmat)$ to the $m$ held-out points comprising $\Xmat'$. 
     \State Record $\{I(\xv_{i}) : \xv \in \Xmat' \}$ as the identities of the held-out data points.
        \For{$j = 1$ to $K$}
         \State Assign $U_{j} = U_{i^{*}} : i^{*} = \arg\min_{i} ||\gv_{i} - \cv_{j}||_{2}$ \quad \% for cluster registration 
        \EndFor 
    \State Update $\Amat$ via $\Amat(i,j)= \Amat(i,j) + \mathbbm{1} (I_{i} \rightarrow U_{j}) \,\, \forall \,\, i, j$. \quad \% this forms the CLAM
\EndFor
\State \textbf{Compute:} ${\rm Sum}(i) = \sum_{j=1}^{K} \Amat(i,j) : i=1, \ldots, t$. 
\For{$i = 1$ to $t$}
\For{$k = 1$ to $K$}
\State \textbf{Compute:} ${\rm Max}_{c}(i,k)$ and ${\rm Idx}_{c}(i,k)$ via (\ref{MaxandIdxeq1}) and (\ref{MaxandIdxeq2}), respectively.  
\EndFor
\EndFor
\State Populate $\Amat_{S}$, $\Amat_{I}$ via $\Amat_{S}(i,k)={\rm Max}_{c}(i,k)/{\rm Sum}(i)$ and $\Amat_{I}(i,k) = {\rm Idx}_{c}(i,k)$
\For{$i = 1$ to $K$}
\State Form sets $C_{k,k} = \{ i : \Amat_{I}(i,1) = k \}$ and $V_{k,k} = \{\Amat_{S}(i \in C_{k,k}, 1)\}$  
\EndFor
\For{$k = 1$ to $K$}
\For{$j = k+1$ to $K$}
\State Form the sets: $C_{k,j}$ and $V_{k,j}$ via (\ref{Eqsetckj1}) and (\ref{Eqsetvkj1}), respectively.  
\EndFor
\EndFor
\State \textbf{Compute:} \{CRI($U_{i}$)\}, \{WCRI($U_{i}$)\}, ERICA statistic, WCRI, TWCRI via (\ref{CRIEq1})-(\ref{WCRIEq2b}). 
\State \Return ERICA statistic, WCRI, TWCRI, \{CRI($U_{i}$)\}, \{WCRI($U_{i}$)\}, ${\rm PCSP}_{i}$ : $i=1, \ldots, K$.
\end{algorithmic}
\end{algorithm}

\subsection{Considered clustering techniques}

It is important to quantitatively assess the extent to which the specification of $\mathcal{A}$ determines whether clusters exist in a dataset and if they are replicable findings. 
K-means is perhaps the most widely used technique for separating points into groups, and the majority of CR studies have focused on this method. 
This highlights the difficulty of the clustering problem, especially in realistic scenarios where the dataset is not well-structured and the number of clusters is not known a priori. 
K-means seeks centroids to partition the space and is inherently biased to (convex) spherical clusters \citep{jain2010}. 
The technique requires the number of clusters to be specified a priori, and is not robust to noisy data, especially outliers. 
Furthermore, it is sensitive to the initial conditions used to run the algorithm, specifically the initial placement of the centroids. 
This has remained an issue despite advancements such as k-means++. 

Hierarchical clustering (HC) does not require a representation for each cluster, instead it assumes that they have a nested structure via each group being composed of smaller groups. 
We consider the agglomerative subclass of HC, specifically with the Ward linkage function (HC-WL) and the single linkage function (HC-SL). 
Although the $K$ value is not explicitly required to perform HC, the point at which a resulting hierarchical tree, or dendrogram, is cut will determine the number of clusters. 
Thus, $K$ must be specified to attain the cluster identities that have been assigned to data points upon the conclusion of the algorithm. 

\citet{masoero2023} reported that the choice of clustering algorithm significantly affected replicability estimates. 
At first glance, this is expected since different methods exploit distinct structure with various objective functions that may give rise to highly disparate partitions in a dataset. 
However, this observation is not obvious, and perhaps surprising. 
This is because we are considering replicability rather than accuracy. 
In the case that the clusters decided by $\psi_{1}(.;\mathcal{A}_{1}, \Xmat)$ and $\psi_{2}(.;\mathcal{A}_{2}, \Xmat)$ are very different, it is not intuitive that the consistency in the decisions of the two techniques must vary. 
Nevertheless, it is important to be cognizant that any one method may give an inadequate picture, and hence our rationale for seeking consensus across multiple clustering techniques. 

\subsection{A cluster registration algorithm} 

Across MCSS iterations, a clustering technique will form different groups based on the data that is provided. 
It is crucial to maintain consistency between what the clustering has called "cluster $k$" (i.e. $U_{k}$) in iteration $i$ and what it has called cluster $k$ in iteration $j$. 
Failure to do so would render the findings of a CR study meaningless, if not misleading. 
The effects of cluster registration on CR have been discussed in studies such as \citet{vanderlaan2003} with the authors considering the BS rather than MCCS. 
The method in \citet{vanderlaan2003} pertains to HC, and entails forming a matrix of pairwise distances between the clusters in each bootstrap and a reference set of clusters. 
The pairs that are closest in distance are matched as corresponding clusters. 
The works \citet{lange2004}, \citet{hofmans2015}, and \citet{minaeibidgoli2014} address cluster registration by permuting the group identities attained at each iteration of their analysis to maximize the similarity between the groups determined for the entire dataset (or a suitable fixed reference). 
This is effective in providing an upper bound on the CR, but is too optimistic. 
The analysis is also computationally burdensome since it requires considering a large number of possibilities. 

In the cluster registration algorithm we consider a reference partition obtained by clustering $\mathcal{X}$ into $K$ groups with centroids $\{ \gv_{i}\}$. 
This baseline set of cluster centers is, for convenience, sorted and indexed based on the increasing $L_2$ norm of their centroids $\{\gv_{i}\}$ with $g_i = ||\gv_{i}||_{2}$ such that $U_{i}$: $g_{i} \leq g_{j}$ and $j \in \{i+1, \ldots, K\}$. 
At each MC iteration $b = 1, \ldots, B$ the clustering is then applied to the held-out data $\Xmat' = [ \xv'_{1}, \ldots, \xv'_{m} ]$ to return $K$ cluster centers $\cv_{1}, \ldots, \cv_{K}$. 
The difference between the centroids $\{\cv_{i}\}$ and the reference centroids $\{\gv_{i}\}$ is computed, and we make the greedy assignment
\be
U_{j} = U_{i^{*}} : i^{*} = \arg\min_{i} ||\gv_{i} - \cv_{j}||_{2} \quad \quad {\rm for} \quad j=1, \ldots, K. 
\label{Eqgreedy1}
\ee
\begin{figure}[t]
\centering

\subfigure[]{
  \includegraphics[width=0.37\linewidth]{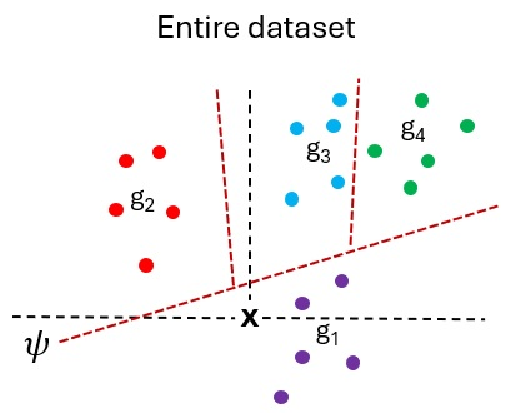}
  \label{fig:1a}
}
\hspace{0.06\linewidth}
\subfigure[]{
  \includegraphics[width=0.37\linewidth]{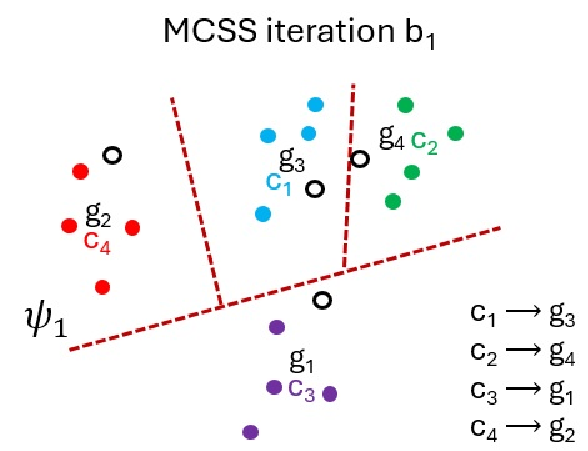}
  \label{fig:1b}
}

\vspace{0.2in}

\subfigure[]{
  \includegraphics[width=0.37\linewidth]{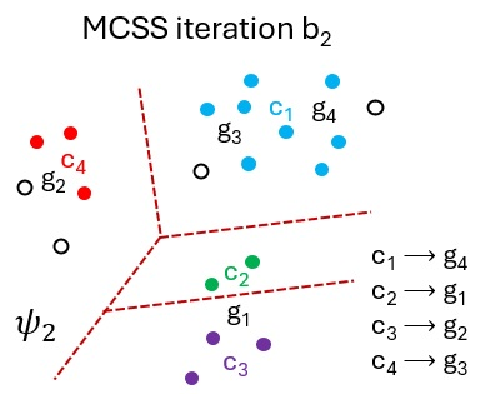}
  \label{fig:1c}
}
\hspace{0.06\linewidth}
\subfigure[]{
  \includegraphics[width=0.37\linewidth]{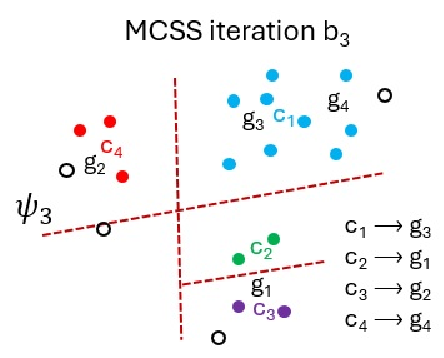}
  \label{fig:1d}
}

\caption{An example of ERICA cluster registration with $t=20$, $p=2$, and $K=4$. The choice $P=80\%$ corresponds to $m=4$ randomly held-out data points (shown as hollow circles) in each MCSS iteration. The locations of the labels $\{\gv_{i}\}$ indicate the centroids of the reference clusters obtained by clustering the entire dataset. The locations of the colored labels $\{\cv_{i}\}$ indicate the centroids computed for the newly formed clusters after the data points are held-out. The assignments (via "$\rightarrow$") denote the cluster registration decisions made by ERICA.}
\label{clusterreg_fig1}
\end{figure}
\begin{figure}[h]
\begin{center}
\epsfig{figure=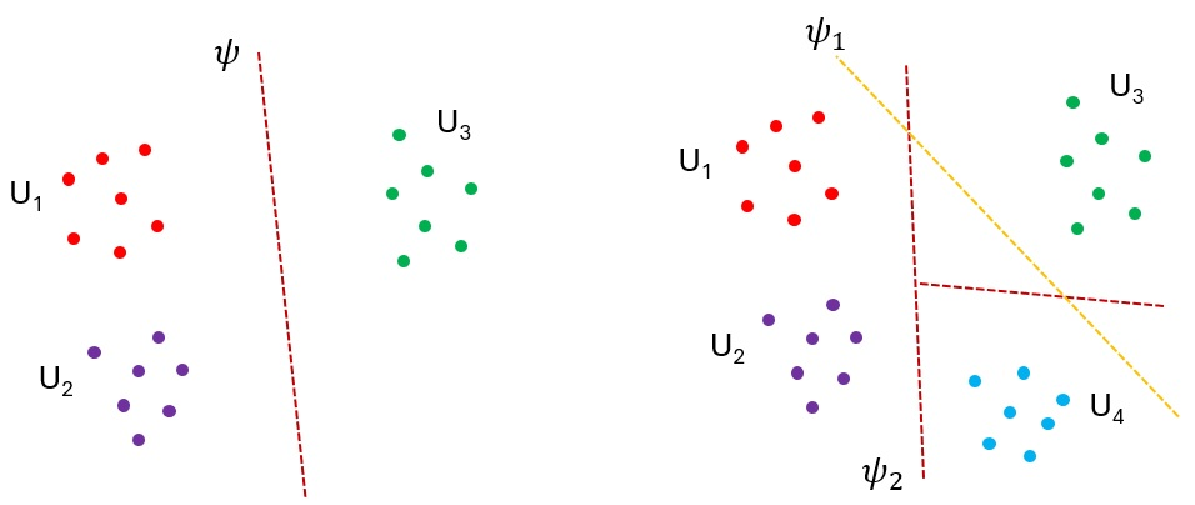, height=2.05in}
\end{center}
\caption{The under-clustering phenomenon. A data-generating process produces fully separable clusters. Although $K=3$ clusters are identifiable in the left figure, it is conceivable to select the parsimonious solution $K=2$ via the groups $U_{1} \cup U_{2}$ and $U_{3}$. On the right, the introduction of a fourth separable group $U_{4}$ may lead to selecting $K=2$ with $U_{1} \cup U_{2} \cup U_{4}$ and $U_{3}$ or the $K=3$ outcome of $U_{1} \cup U_{2}$, $U_{3}$, and $U_{4}$. In either case, under-clustering leads to a more conservative clustering solution with higher replicability if the experiment were repeated.}
\label{underclustfig3}
\end{figure}
Thus, every cluster at MC iteration $b$ is assigned a unique index (i.e. cluster identity) based on the reference centroid that its centroid is closest to. 
This provides a consistent interpretation of the groups across the $B$ iterations. 
Clustering registration is more involved for hierarchical clustering since there is no direct notion of a centroid. 
Nevertheless, by sorting the $K$ groups in ascending order according to the $L_2$ norm of their centers, it is possible to attain a labeling of the clusters. 
An example of the dynamics associated with the ERICA cluster registration algorithm is shown in Figure \ref{clusterreg_fig1}. 
The entire dataset is clustered with the colors denoting the cluster assignment. 
In Figure \ref{clusterreg_fig1}(a), the location of the labels $\{\gv_{i}\}$ indicate the centroids for each cluster. The ERICA cluster registration algorithm for MCSS iteration $b=b_{1}$ is shown in Figure \ref{clusterreg_fig1}(b) with the location of the colored letters $\{\cv_{i}\}$ representing the centroids computed for the newly formed clusters after the data points are held-out. The assignments (via "$\rightarrow$") denote the cluster registration decisions that arise with the considered registration technique. 
%
In Figure \ref{clusterreg_fig1}(c), a greedy registration does not lead to a globally optimal solution in terms of the distances between the computed and the reference centroids. For instance $\cv_3$ is registered as cluster 2, whereas its registration to cluster 1 would have been more appropriate.
A similar scenario arises at $b=b_{3}$ in Figure 
\ref{clusterreg_fig1}(d) where data points corresponding to $\cv_{2}$ are not registered to their globally optimal cluster, i.e. 4. 
We have not yet discussed the specification of $K$. This important facet is elaborated on below.

\subsection{Choosing the proper number of clusters} 

The literature on discovering the number of groups in a dataset is vast. 
We deliberately refer to the appropriate number of clusters rather than the correct number of clusters since we are in an unsupervised learning domain without a ground truth. 
Nevertheless, the discovery of the appropriate number of clusters $K^{*}$ will be undertaken for a synthetic dataset.
This will allow us to test the efficacy of our algorithm for discovering $K^{*}$ when there is a priori knowledge akin to a ground truth. 
Depending on the clustering technique, various factors may lead the data to be under- or over-clustered. 
The latter has been discussed recently as being rampant and detrimental \citep{kimes2017, liu2008, mcshane2002, grabski2023}. 
While a concern, this will be mitigated by considering several clustering techniques in parallel as well as different metrics that provide varying views on a dataset. 
Since unsupervised learning has a degree of subjectivity, the collective results can be used to examine whether consensus exists across methods. 
Conversely, a lack of agreement may be a sign of distortion, artifacts, or the absence of structure in a dataset. 

The under-clustering problem has received less attention than over-clustering, and is frequently studied in the same vein \citep{andrews2018, ghasemian2019}. 
Notwithstanding, it is natural for clustering metrics to under-cluster by favoring a conservative solution. 
This phenomenon was recently noted with several techniques and the adjusted Rand index (ARI) in \citet{masoero2023}. 
An illustration of increased certainty with a fewer number of clusters is shown in Figure \ref{underclustfig3}. 
Aside from the trivial solution of $K=1$, the scenario on the left shows under-clustering to be likely with an outcome of $K=2$ with $U_{1} \cup U_{2}$ being one cluster. 
In the second illustration, both of the clustering rules result in under-clustering. 
From a replicability perspective, $\psi_2$ would be expected to yield a more desirable solution while exhibiting less under-clustering than $\psi_1$. 
In tying this to ERICA, for the specification of $K$ clusters, define $\mathcal{M}_{K}$ as one of the presented metrics, i.e. ERICA statistic, WCRI, or TWCRI. 
It is not difficult to construct a case such as the first scenario in Figure \ref{underclustfig3} where $U_{1} \cup U_{2}$ are more reliably separated from $U_{3}$ than $U_{1}$, $U_{2}$, and $U_{3}$ will be separated from each other. 
Thus, we would expect that $\mathcal{M}_{2} > \mathcal{M}_{3}$. 
While analytically sound, this leads to missed structure. 
The addition of $U_{4}$ to the dataset shown via the second scenario exacerbates this issue as we expect that $\mathcal{M}_{2} > \max \{\mathcal{M}_{3}, \mathcal{M}_{4}\}$. 

To mitigate under-clustering, we propose Algorithm 2. 
The rationale is to introduce conditions to encourage a solution that selects a higher number of replicable clusters rather than settling for the safest solution. 
Understanding Algorithm 2 requires recognition that the presence of "NA" for a cluster-level metric (e.g. $\exists \,\, k : CRI(U_{k}) = NA$) arises when $X_{k}=0$ indicating that no observation was assigned most frequently to $U_{k}$. 
%
Aside from rendering the metric undefined for $U_{k}$, this also indicates that no observation had $U_{k}$ as its primary cluster, raising doubt as to whether the cluster should exist.
In Algorithm 2, the first if-condition disqualifies $K$ values for which there exists at least one cluster with no data points assigned to it at the highest frequency. 
The second if-condition favors cases where a solution with a larger $K$ has led to increased replicability than its immediate predecessor. 
This may be viewed akin to the "jump" computed in \citet{sugar2003} and used to mitigate the overly conservative selection of $K^{*}$. 
The aforementioned work drew upon a distortion measure and sought the largest jump whereas we introduce replicability measures and consider the last positive jump. 
We are ready to define the number of clusters discovered by ERICA. 
\begin{definition}[ERICA number of clusters, $K^{*}$]
For $K$ clusters, consider a partition $\pi_{K} = U_{1}, \ldots, U_{K}$ and define the set of partitions $\pi = \{\pi_{K} : K=K^{\rm min}, \ldots, K^{\rm max} \,\,\, s.t. \,\,\, |U_{i}| > 0 : i=1, \ldots, K\}$. 
For $n \leq K^{\rm max} - K^{\rm min} +1$, define the potentially smaller set of partitions $\overline{\pi} = \{\pi_1, \ldots, \pi_n : |\pi_1| <, \ldots, < |\pi_n|\}$. 
We define
\be
K^{*} \,\, = \,\, \max\{K^{min}, |\pi_{i}| \,\,\, s.t. \,\,\, \mathcal{M}(\pi_{i}) \geq \mathcal{M}(\pi_{i-1}) \}. 
\ee
where $\mathcal{M}(.)$ denotes the chosen replicability metric, e.g. the ERICA statistic. 
\end{definition} 
\noindent
It should be noted that the Algorithm 2 portion of ERICA leverages the premise that $\mathcal{M}_{k} > \mathcal{M}_{k-1}$ indicates the addition of a cluster has increased replicability. 
\begin{table}[ht]
    \centering
    \caption{Panels summarizing the CRI (top) and WCRI (bottom) values obtained using ERICA across $M$ datasets when $K$ is the primary parameter. A mean CRI value (the ERICA statistic) serves as the summary metric for each entry. The mean and total WCRI values are reported to incorporate the relative number of observations in a cluster into the summary metrics. The sub-entries for $U_{i}: i = 1, \ldots, K^{\max}$ report the CRI and WCRI metrics for the individual clusters.}
\begin{adjustbox}{width=0.95\textwidth}
    \begin{tabular}{lcccccc}
        \toprule
        & K = 2 & K = 3 & \dots & K = $K^{max}$ \\
        \midrule
        Case 1 & ERICA statistic & ERICA statistic & \ldots & ERICA statistic \\
              & ($CRI(U_{1})$, $CRI(U_{2})$) & ($CRI(U_{1})$, $CRI(U_{2})$, $CRI(U_{3})$) & \ldots & ($CRI(U_{1})$, \ldots, $CRI(U_{K^{max}})$) \\
         \,\, \quad \vdots &  & & \vdots &  \\
        Case M & ERICA statistic & ERICA statistic & \ldots & ERICA statistic \\
              & ($CRI(U_{1})$, $CRI(U_{2})$) & ($CRI(U_{1})$, $CRI(U_{2})$, $CRI(U_{3})$) & \ldots & ($CRI(U_{1})$, \ldots, $CRI(U_{K^{max}})$) \\
        \bottomrule
    \end{tabular}
    \end{adjustbox}
    \label{template_for_mytable1}
\vspace{0.2in}
\begin{adjustbox}{width=0.95\textwidth}
    \begin{tabular}{lcccccc}
        \toprule
        & K = 2 & K = 3 & \dots & K = $K^{max}$ \\
        \midrule
        Case 1 & WCRI, TWCRI & WCRI, TWCRI & \ldots & WCRI, TWCRI \\
              & ($WCRI(U_{1})$, $WCRI(U_{2})$) & ($WCRI(U_{1})$, $WCRI(U_{2})$, $WCRI(U_{3})$) & \ldots & ($WCRI(U_{1})$, \ldots, $WCRI(U_{K^{max}})$) \\
         \,\, \quad \vdots &  & & \vdots &  \\
        Case M & WCRI, TWCRI & WCRI, TWCRI & \ldots & WCRI, TWCRI \\
              & ($WCRI(U_{1})$, $WCRI(U_{2})$) & ($WCRI(U_{1})$, $WCRI(U_{2})$, $WCRI(U_{3})$) & \ldots & ($WCRI(U_{1})$, \ldots, $WCRI(U_{K^{max}})$) \\
        \bottomrule
    \end{tabular}
    \end{adjustbox}
    \label{template_for_mytable12}
\end{table}

\section{Simulation Results}

An algorithm for evaluating and quantifying CR has been presented. 
ERICA is now applied to a synthetic dataset consisting of high-dimensional mixture of Gaussians where there is a ground truth. 
We provide evidence for the utility of the presented approach by first evaluating its performance on variants of the synthetic dataset that are challenging to cluster. 
This is due to the clusters overlapping and the distribution of data points per cluster being uniform. 
Nevertheless, it serves to evaluate the platform's efficacy since a ground truth is known. 
Subsequently, we consider $\mathcal{X}$ comprised of gene expression data from breast cancer tumors.
The objective will be to identify the samples as belonging to subtypes that are phenotypically associated with different severity. 
The findings will illustrate how reliably cancer subtypes are distinguished in three datasets. 
In the analysis we shall use $P=80\%$ which is analogous to five-fold cross-validation (CV). 

Equation (\ref{toeplitzEq1}) is a Toeplitz correlation structure that has been motivated for time series analysis and spatial data. 
Algorithm 1 is applied to the Gaussian mixture dataset under different dimensionality, dispersion, and correlation structures. 
Naturally, $K=4$ remains the ground truth since there will be four components for all cases. 
Each analysis will consist of running the ERICA evaluation platform in Figure \ref{highlevel_fig1} for $B=200$ MCSS iterations and computing the ERICA statistic, WCRI, and TWCRI. 
The panels in Table \ref{template_for_mytable1} are then populated to provide insight into the per-cluster replicability. 
This has been done with the three clustering techniques and is shown via tabulation of Tables \ref{Kmeanstable1} to \ref{ACsingle_pbiggern_results2}. 
The bold entries of each row correspond to the $K$ that was selected as most appropriate via Algorithm 2.

\subsection{Synthetic dataset: mixture of Gaussians}

The presence of a high-dimensional Gaussian mixture is a prevalent scenario in nature, engineering, and biology. 
Motivation of this model as synthetic data for clustering is found in 
\citet{liu2008} and references within, as well as newer works such as \citet{golalipour2021} and \citet{dalmaijer2022}. 
%
We consider the dataset consisting of four components via 
\be
\xv_{i} \sim \frac{1}{4} \sum_{j=1}^{4} N(\av_{j}, \Rmat)
\label{Eqdisteq1}
\ee
with $\av_{1} = \bf{1}$, $\av_{2} = 4 \bf{1}$, $\av_{3} = 7 \bf{1}$, and $\av_{4} = 10 \bf{1}$ where $\bf{1}$ denotes the all 1s vector. 
A standard model will entail the features being mutually independent and identically distributed via $\Rmat = \Imat$. 
We also consider a high-variance case with $\Rmat = 10\Imat$ (referred to as $\uparrow$ variance), as well as correlation between features where the components of $\Rmat$ are specified by
\be
\Rmat(i, j) = \left\{ \begin{array}{rcl} 1, &  i=j, \\ 
0.9, &   |i-j|=1, \\
0.8, &   |i-j|=2, \\
0.7, &   |i-j|=3, \\
0.6, &   |i-j|=4, \\
0.5, &   |i-j|=5, \\
0.4, &   |i-j|=6, \\
0.3, &   |i-j|=7, \\
0.2, &   |i-j|=8, \\
0.1, &   |i-j|=9, \\
0.001, &  |i-j|=10, \\
0, & \mbox{otherwise.}   \\ \end{array}\right.
\label{toeplitzEq1}
\ee

\begin{table}[ht]
    \centering
    \caption{The ERICA statistic for the synthetic datasets generated from a four-component Gaussian mixture. The statistic was computed for three clustering techniques (k-means, HC-WL, and HC-SL) with $B=200$ MCSS iterations and $P=80\%$. $K^*$ denotes the most appropriate number of clusters according to the ERICA statistic.}
    \begin{adjustbox}{width=0.6\textwidth}
    \begin{tabular}{lccc}  
        \toprule
        \cmidrule{1-4}
        Dataset & Clustering & ERICA statistic & $K^*$ \\
        \midrule
        t = 10,000, p = 100 & K-means & 0.994 & 4 \\
         & HC-WL, HC-SL & 1 , 1 & 4, 4 \\
        t = 10,000, p = 200 & K-means & 0.993 & 4 \\
         & HC-WL, HC-SL & 1, 1 & 4, 4 \\
        t = 10,000, p = 400 & K-means & 0.994 & 4 \\
         & HC-WL, HC-SL & 1, 1 & 4, 4 \\
        t = 10,000, p = 100, Toeplitz & K-means & 0.997 & 4  \\
         & HC-WL, HC-SL & 0.999, 0.907 & 4, 5 \\
        t = 10,000, p = 200, Toeplitz & K-means & 0.998 & 4 \\
         & HC-WL, HC-SL & 1, 0.827 & 4, 6 \\
        t = 10,000, p = 400, Toeplitz & K-means & 0.997 & 4 \\
         & HC-WL, HC-SL & 1, 1 & 4, 4 \\
        t = 10,000, p = 100, $\uparrow$ variance & K-means & 0.966 & 4 \\
         & HC-WL, HC-SL & 0.999, 0.662 & 4, 5 \\
        t = 10,000, p = 200, $\uparrow$ variance & K-means & 0.945 & 4 \\
         & HC-WL, HC-SL & 1, 0.507 & 4, 5 \\
        t = 10,000, p = 400, $\uparrow$ variance & K-means & 0.941 & 4 \\
         & HC-WL, HC-SL & 1, 0.995 & 4, 4 \\
        t = 100, p = 1,000 & K-means & 0.997 & 4 \\
         & HC-WL, HC-SL & 0.995, 0.998 & 4, 4 \\
        t = 200, p = 1,000 & K-means & 0.99 & 4 \\
         & HC-WL, HC-SL & 1, 1 & 4, 4 \\
        t = 100, p = 1,000, Toeplitz & K-means & 0.986 & 4 \\
         & HC-WL, HC-SL & 0.996, 0.998 & 4, 4 \\
        t = 200, p = 1,000, Toeplitz & K-means & 0.993 & 4 \\
         & HC-WL, HC-SL & 1, 1 & 4, 4 \\
        t = 100, p = 1,000, $\uparrow$ variance & K-means & 0.992 & 2 \\
         & HC-WL, HC-SL & 0.995, 0.998 & 4, 4 \\
        t = 200, p = 1,000, $\uparrow$ variance & K-means & 0.931 & 4 \\
         & HC-WL, HC-SL & 1, 1 & 4, 4 \\
        \bottomrule
    \end{tabular}
    \end{adjustbox}
    \label{winningCRI_1}
\end{table}
In Table \ref{winningCRI_1}, with $t > p$, we note that ERICA returns the correct number of clusters for all cases when k-means or HC-WL are used. 
Even the lowest ERICA statistic (0.941 for $t=10,000, p=400, \uparrow$ variance with k-means) in these cases suggests a high degree of replicability. 
With the HC-SL, there are two cases with the Toeplitz correlation structure and the high variance scenarios where either $K^{*}=5$ or 
$K^{*}=6$ are declared. 
These four scenarios correspond to the four lowest ERICA statistics via 0.507 ($t=10,000, p=200, \uparrow$ variance), 0.662 ($t=10,000, p=100, \uparrow$ variance), 0.827 ($t=10,000, p=100,$ Toeplitz), and 0.907 ($t=10,000, p=200,$ Toeplitz). 
It is expected that the higher variance and presence of correlation will give rise to lower ERICA statistics. 
The case of $p > t$ in Table \ref{winningCRI_1} shows ERICA providing the correct number of clusters for all cases except with k-means and $t=100, p=1,000, \uparrow$ variance where there is under-clustering via $K^{*}=2$. 
The ERICA statistics are near the maximal value for all considered cases including the $K^{*}=2$ scenario. 
This is likely because the higher variance increases the overlap between clusters, leading k-means to consistently (ERICA statistic = 0.992) partition the data into the same two groups. 
%


Alternatively, it is possible to evaluate the replicability as well as the clustering assignments via a consensus among the ERICA statistic and the two derivative metrics. 
This is done by introducing the quantities $a$, $b$, and $c$ in the $(a, b, c)$ tuple to denote the number of times that a $K$ value has been selected as most appropriate according to the ERICA statistic, WCRI, and TWCRI metrics, respectively. 
For simplicity we allocate equal weight to each of the three clustering techniques, thus each entry $a, b, c$ can take values in $\{0, 1, 2, 3\}$. 
In the case of $t > p$, Table \ref{Fintalsynth1_res1} shows that the ERICA statistic and TWCRI metrics return the correct number of clusters irrespective of the dimensionality, correlation structure, or variance. 
There are several cases where the ERICA statistic and TWCRI metrics over-cluster via assignments of $K^{*}=5$ or $6$. 
With a few exceptions (i.e. $p = 400$ and $p = 100, \uparrow$ variance) the WCRI metric favors the parsimonious solution of $K^{*}=2$ according to the three clustering techniques. 

We examine more specific facets of the results that have been provided by the individual clustering techniques. 
The observation that the ERICA statistic and TWCRI values in Tables \ref{Kmeanstable1} and \ref{Kmeanstable12} are rather close to their maximal value of one indicates that k-means has declared the datasets to exhibit four clusters at a highly replicable degree. 
The ERICA statistic and TWCRI for the three dimensionalities are lower when the variance of the features has increased. 
For instance, with $K=4$, in comparing $p=400$ to $p=400, \uparrow$ variance, we note a 5.33\% decrease in ERICA statistic and a 5.34\% decrease in TWCRI. 
This is expected since the increased variance should obfuscate the clusters and thus degrade the consistency of the assignments. 
In Tables \ref{Kmeanstable1} and \ref{Kmeanstable12} it is encouraging that for all of the datasets the metrics monotonically decrease as $K$ increases further from four. 
The same trends and similar numerical results are noted with HC-WL via Tables \ref{ACWardresults1} and \ref{ACWardresults2}. 
For the majority of the cases the ERICA statistic reaches the maximal value of stability when $K=4$. 
The corresponding WCRI values are not exactly one because of numerical rounding used to maintain a consistent number of significant digits. 
Nevertheless, the point remains that the use of Algorithms 1 and 2 with HC-WL leads to a robust result of $K^{*}=4$ despite the correlation and increased variability. 
The findings from Tables \ref{ACSLresults1} and \ref{ACSLresults12} with HC-SL are similar to k-means and HC-WL with the important exception of the degradation brought on by correlation and higher variance. 
When $p=100$ and $200$, the HC-SL replicability is affected by the noisiness of the dataset and the correlation among the features. 
In fact, in both cases the ERICA statistic and TWCRI values have dropped to a point where the results do not indicate four clusters as being the most likely structure for the data. 
In comparing the HC-SL result when $p=100$ to $p=100, \uparrow$ variance, a 33.8\% decrease is noted in the ERICA statistic. 
It is interesting that the ERICA statistic and TWCRI metrics show the replicability with HC-SL to recover at the higher dimensionality of $p=400$. 
This may be attributed to data points that were at cluster boundaries in the lower dimensional datasets becoming less close with additional signal-bearing dimensions. 
The TWCRI metric led to $K^{*}=6$ for a dataset with Toeplitz structure and one with high variance. 
The WCRI results indicate the parsimonious solution by HC-SL since $K^{*}=2$ for all of the datasets except the case $p = 100, \uparrow$ variance. 

We now discuss the results when considering the consensus of the introduced metrics for cases with $p > t$. 
Table \ref{Fintalsynth1_res1} also contains summary results of Tables \ref{Kmeans_pbiggern_table1} to \ref{ACsingle_pbiggern_results2}. 
The ERICA statistic and TWCRI metrics provide the correct number of clusters across all cases aside for $t=100, \uparrow$ variance for which $K=2$ is deemed most appropriate. 
With the exception of $t=100$, Toeplitz, the WCRI metric under-clusters via $K^{*}=2$ across the datasets. 
Tables \ref{Kmeans_pbiggern_table1} to \ref{ACsingle_pbiggern_results2} indicate that the ERICA statistic and TWCRI are near their maximal value of one, thus the clustering techniques are rather certain of the number of clusters found. 
With the exception of the cases where k-means was applied, the ERICA statistic and TWCRI metrics have not been lowered with the increased variance in any of the scenarios. 
We attribute this to the $p > t$ condition providing degrees of freedom to separate the data points generated from the clusters despite the added overlap among the groups. 
The columns in Table \ref{Fintalsynth1_res1} are very concentrated. 
With a few exceptions, the ground truth value or the parsimonious case of two clusters are selected. 
The concentration indicates a degree of consensus which informs a user that clusters exist and may be reliably found. 
Furthermore, it is important that the analysis rarely over-clusters. 
Such an effect is especially true with a higher number of features (i.e. $p > 100$), and holds in the Toeplitz and increased variance scenarios. 
This is desirable because it reduces the introduction of artificial clusters that would be unreproducible in an ensuing trial. 
\begin{table}[ht]
    \centering
    \caption{The ERICA statistic for the breast cancer datasets. The statistic was computed for three clustering techniques (k-means, HC-WL, and HC-SL) with $B=200$ MCSS iterations and $P=80\%$. $K^*$ denotes the most appropriate number of clusters according to the ERICA statistic.}
    \begin{adjustbox}{width=0.55\textwidth}
    \begin{tabular}{lccc}  
        \toprule
        \cmidrule{1-4}
        Dataset & Clustering & ERICA statistic & $K^*$ \\
        \midrule
        Mainz (full) & K-means & 0.752 & 2 \\
         & HC-WL, HC-SL & 0.643, 0.912 & 6, 2 \\
        Mainz (3G) & K-means & 0.819 & 6 \\
         & HC-WL, HC-SL & 0.769, 0.817 & 6, 3 \\
        Transbig (full) & K-means & 0.877 & 2 \\
         & HC-WL, HC-SL & 0.595, 0.68 & 7, 7 \\
        Transbig (3G) & K-means & 0.787 & 7 \\
         & HC-WL, HC-SL & 0.793, 0.721 & 7, 8 \\
        VDX (full) & K-means & 0.539 & 5 \\
         & HC-WL, HC-SL & 0.871, 0.687 & 2, 6 \\
        VDX (3G) & K-means & 0.828 & 6 \\
         & HC-WL, HC-SL & 0.783, 0.669 & 6, 7 \\
        \bottomrule
    \end{tabular}
    \end{adjustbox}
    \label{winningCRI_breast1}
\end{table}

\begin{figure*}[!h]
\centering
\subfigure{
\includegraphics[width=0.85\linewidth]{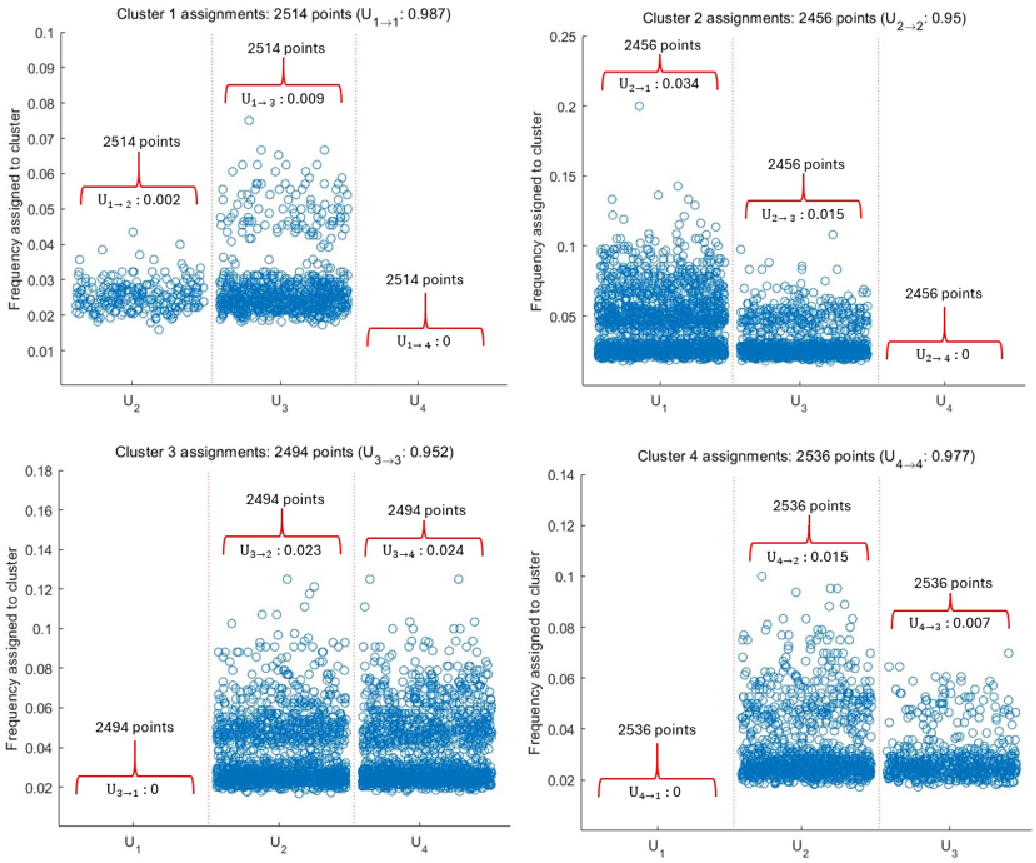}
}
\subfigure{
\includegraphics[width=0.45\linewidth]{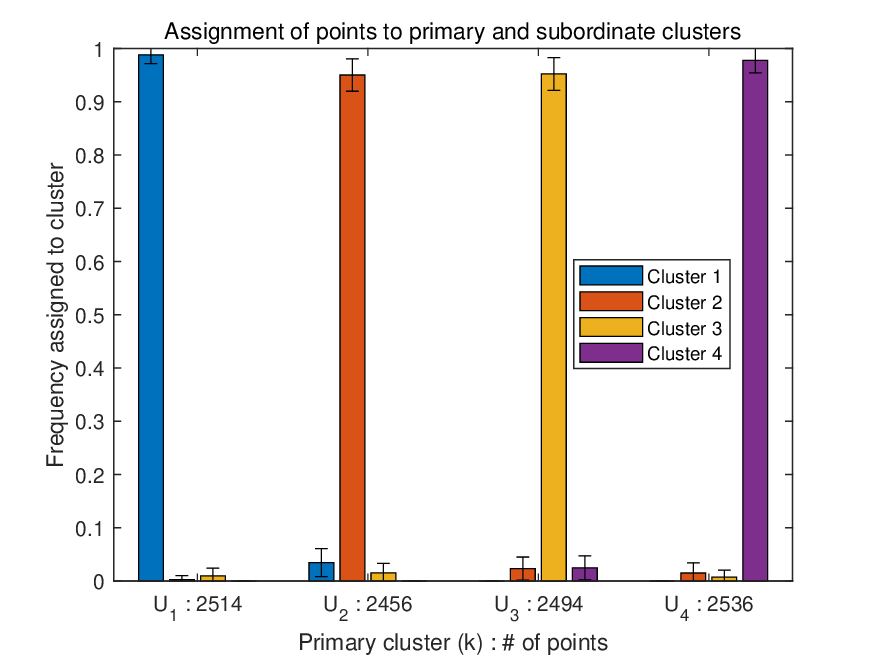}
}
\caption{The PCSPs and ICAH for the Gaussian mixture dataset with $t=10,000$, $p=100, \uparrow$ variance, and k-means with $K=4$. The CRI values and cross-cluster spillovers are illustrated by the four PCSPs (top). The ICAH (bottom) provides a cluster-level view of the replicability, indicating that $U_1$ and $U_4$ are more stable than the other two groups. Only limited spillover is observed between each primary cluster and its three secondary clusters.} \label{figPCSP_GaussMix4_100dim_highvar_1}
\end{figure*}

The proximity of the clusters and the data points that may be deemed unstable by not being definitively assigned to the same group are visualized via the PCSPs and ICAH. 
In Figure \ref{figPCSP_GaussMix4_100dim_highvar_1} the PCSPs and ICAH are shown for the case of $t=10,000, p=100, \uparrow$ variance where k-means was used with $K = 4$. 
We note that $U_{1}$ is the most replicable of the clusters as noted via its CRI ($CRI(U_{1})=0.987$). 
The data points assigned at the highest frequency to cluster 1 are closest to cluster 3 ($CRI(U_{1 \rightarrow 3}) = 0.009$) and furthest from cluster 4 ($CRI(U_{1 \rightarrow 4}) = 0)$. 
The asymmetry of the spillover in cluster assignments is apparent in Figure \ref{figPCSP_GaussMix4_100dim_highvar_1} since $U_{4 \rightarrow 2} > U_{2 \rightarrow 4} =0$. 
Lastly, we comment that the replicability analysis selects the most appropriate clustering algorithm for the data \citep{liu2022stability, masoero2023, jain1987bootstrap}. 
For Gaussian mixtures, we expect k-means and HC-WL to perform reasonably well even with the features' higher variance. 
Despite the increase in overlap among the groups, they remained spherical and were equally-sized. 
Conversely, we did not expect HC-SL to perform well in the presence of increased variability since the shortest distance between points is more easily affected. 
This resonates with the loss in replicability with HC-SL being larger than that seen with the other methods (Table \ref{winningCRI_1}).

\subsection{Breast cancer gene expression data}

The same steps are followed in the application of ERICA to several breast cancer datasets. 
We consider the three gene expression datasets studied in \citet{masoero2023} and \citet{haibekains2012}. 
The biological details of the data are provided in the aforementioned works as well as their references. 
Seminal studies have used clustering to establish four molecular subtypes of breast cancer \citep{perou2000, sorlie2001, parker2009}. 
ERICA is run separately on the full and three gene version of the Mainz, Transbig, and VDX datasets. 
In each case the three clustering techniques are used with $K \in \{2, \ldots, 8 \}$ and $B=200$. 
The cluster-level values for the metrics are provided in Tables \ref{mainzfull_res1} to \ref{Transbigfull_res2}. 
The ERICA statistics in Table \ref{winningCRI_breast1} indicate that for the Mainz dataset $K^{*}=2$ via k-means and HC-SL. 
The most likely number of clusters is six with HC-WL. 
For Transbig, we have $K^{*}=2$ with k-means, and $K^{*}=7$ via HC-WL and HC-SL. 
When considering the VDX dataset, we note that $K^{*}=5$ when using k-means, while two and six clusters are the most appropriate with HC-WL and HC-SL, respectively. 
It is important that for all three datasets, the highest ERICA statistic (0.912 for Mainz, 0.877 for Transbig, 0.871 for VDX) is observed with two clusters. 
For the 3G version of the Mainz dataset, the results in Table \ref{winningCRI_breast1} indicate that $K^{*}=6$ with k-means and HC-WL. 
The use of HC-SL returns three groups as the most likely number of clusters for Mainz (3G). 
When applied to Transbig (3G), the ERICA statistic indicates $K^{*}=7$ with k-means and HC-WL, while $K^{*}=8$ with the use of HC-SL. 
Application of ERICA to the VDX (3G) dataset returns $K^{*}=6$ with k-means and HC-WL, while $K^{*}=7$ with HC-SL. 
Considering the highest ERICA statistic for each dataset leads to six (0.819 with k-means), seven (0.793 with HC-WL), and six (0.828 with k-means) clusters for Mainz (3G), Transbig (3G), VDX (3G), respectively.

Results are presented with the approach of evaluating replicability via a consensus among the ERICA statistic and the derivative metrics. 
We choose to not include the WCRI in this analysis because of its propensity to under-cluster and lead to a conservative solution. 
This was seen with the synthetic data results in Table \ref{Fintalsynth1_res1}. 
The summarized results in Table \ref{Fintalparmig1_res1} indicate that when considering the full set of features ($p=22,283$), the metrics provide a different account of the number of clusters across the three datasets. 
For the Mainz data, the two metrics indicate $K^{*}=2$ while $K=6$ was also implicated. 
The Transbig results also favor $K^{*}=2$ via the TWCRI while the ERICA statistic indicates seven groups. 
The findings are similar in the case of VDX. 
The ERICA statistic indicates two, five, and six clusters according to HC-WL, k-means, and HC-SL, respectively. 
The TWCRI is more conclusive in stating $K^{*}=2$. 
Collectively, the three datasets suggest two replicable tumor types when considering the full set of genes in the ERICA evaluation platform. 
%
%
The 3G features support solutions with a larger number of clusters. 
For Mainz (3G), the ERICA statistic indicates six groups while the TWCRI favors the $K^{*}=2$ solution as the most replicable across the clusterings. 
Interestingly, k-means selects for six clusters across the two considered metrics. 
The results for Transbig (3G) show that the ERICA statistic favors seven tumor types. 
The TWCRI metric indicates $K^{*}=$ 5, 7, and 8 as being the most replicable for HC-WL, k-means, and HC-SL, respectively. 
The findings are more categorical for VDX (3G) as $K^{*}=6$ according to the ERICA statistic and TWCRI metric.
\begin{figure}[t]
\centering

\begin{tabular}{cc}
\includegraphics[width=0.45\textwidth]{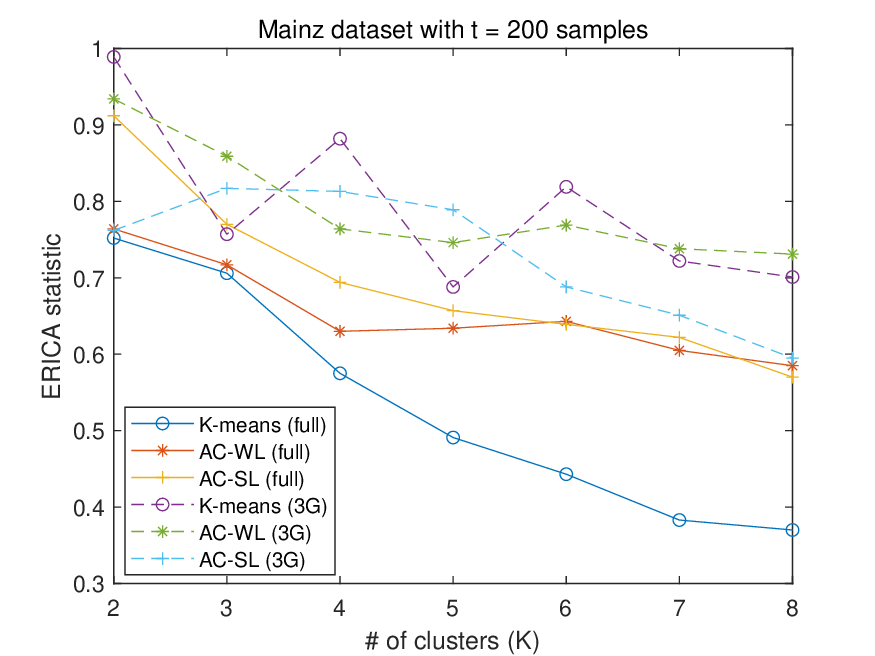} &
\includegraphics[width=0.45\textwidth]{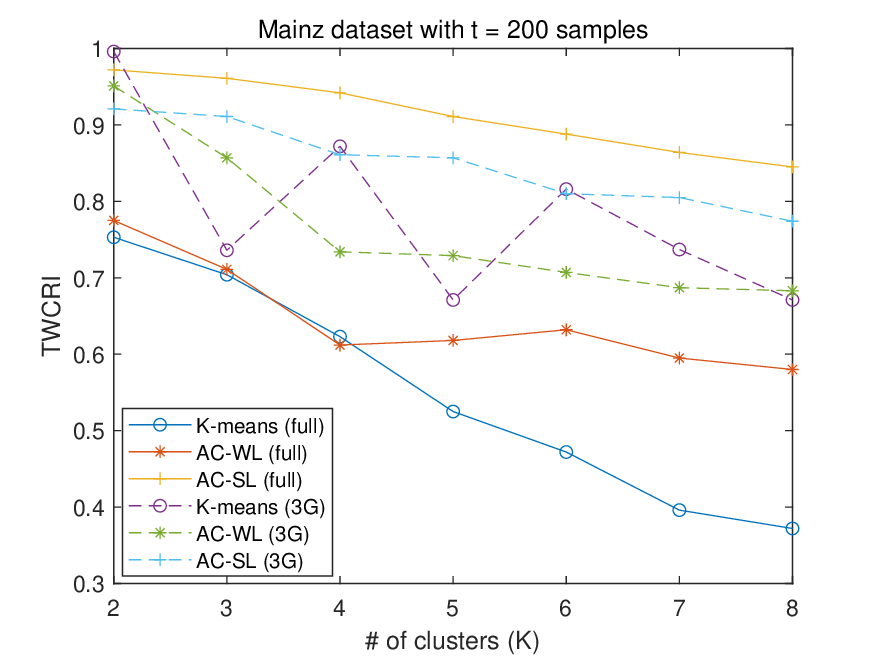} \\[0.1in]

\includegraphics[width=0.45\textwidth]{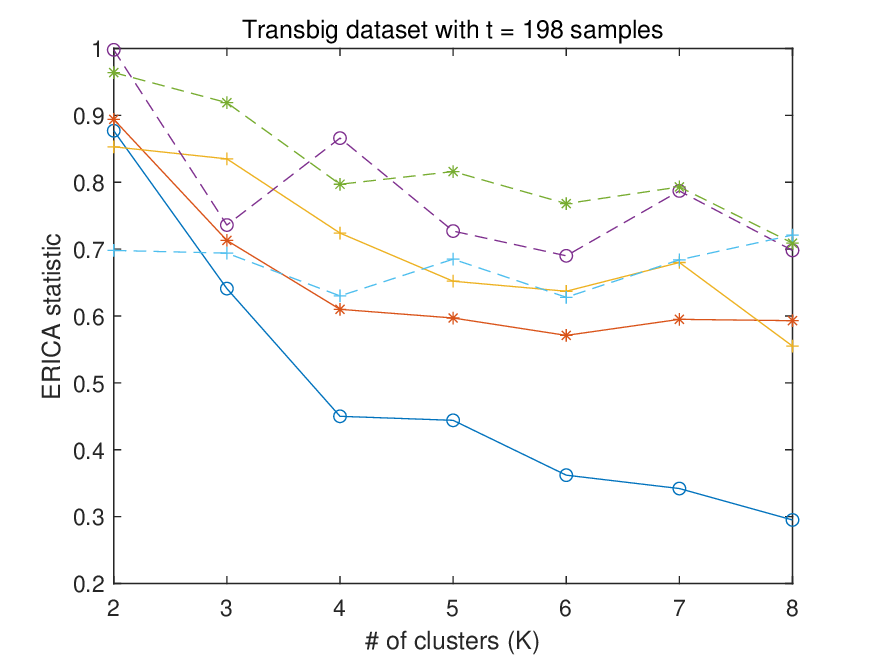} &
\includegraphics[width=0.45\textwidth]{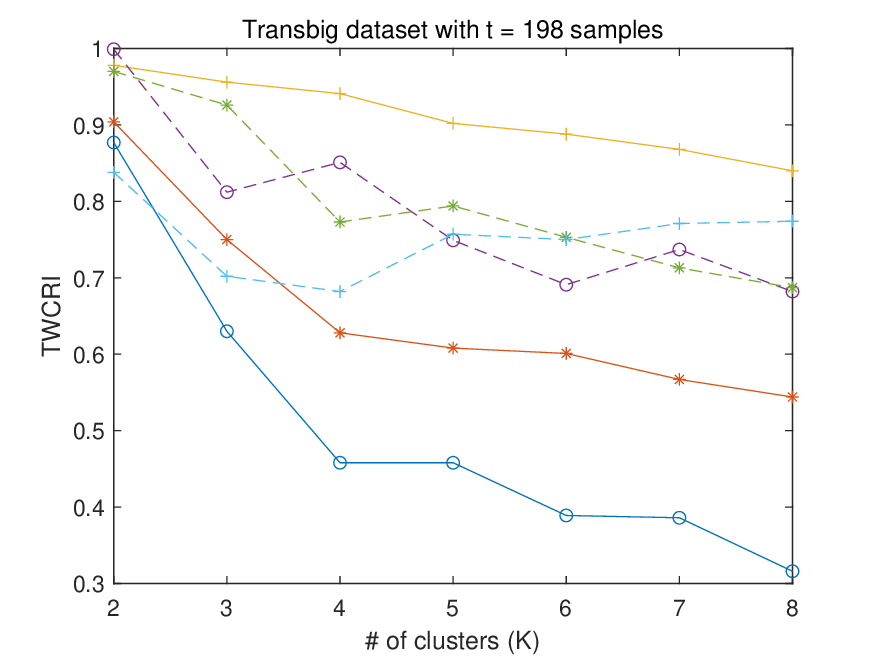} \\[0.1in]

\includegraphics[width=0.45\textwidth]{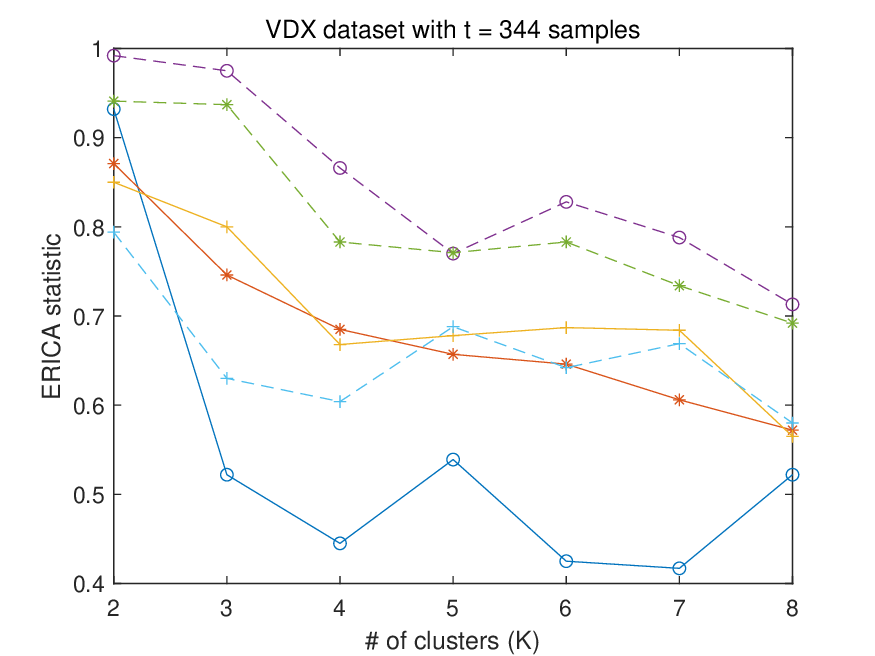} &
\includegraphics[width=0.45\textwidth]{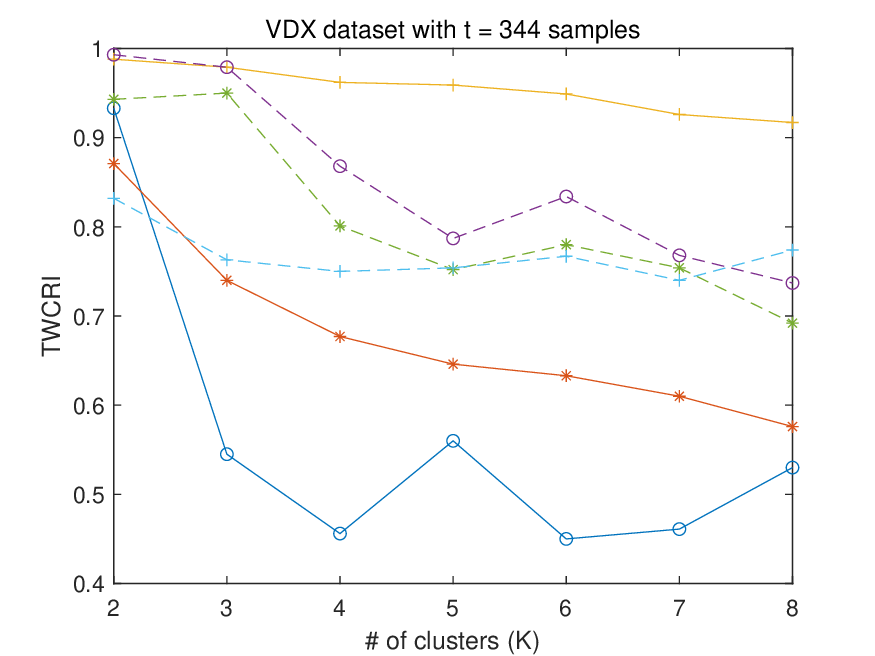}
\end{tabular}

\caption{Application of ERICA to breast cancer data using the full gene set $p=22,283$ (full) and $p=3$ genes (3G) for each sample. The number of replicable clusters is determined using the ERICA statistic and TWCRI. Except with HC-SL, all three datasets exhibit higher metric values when using 3G rather than the full gene set.}
\label{figmetricsParmigdata1}

\end{figure}

\begin{figure*}[!h]
\centering
\subfigure{
\includegraphics[width=0.85\linewidth]{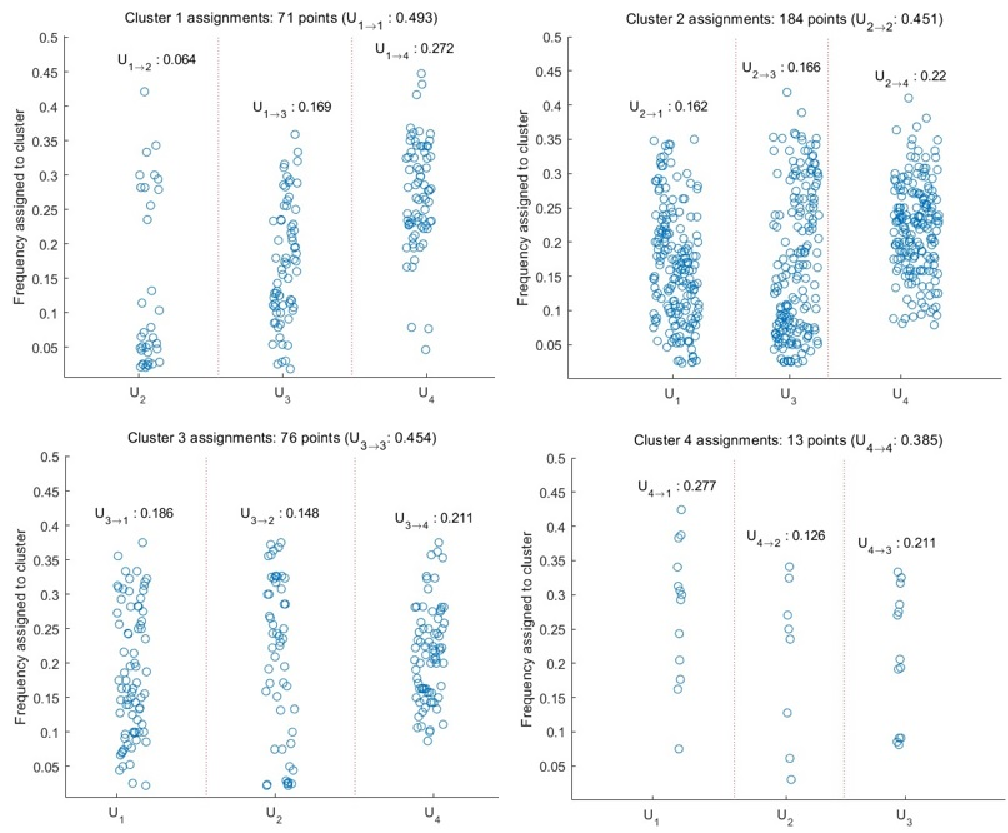}
}
\subfigure{
\includegraphics[width=0.43\linewidth]{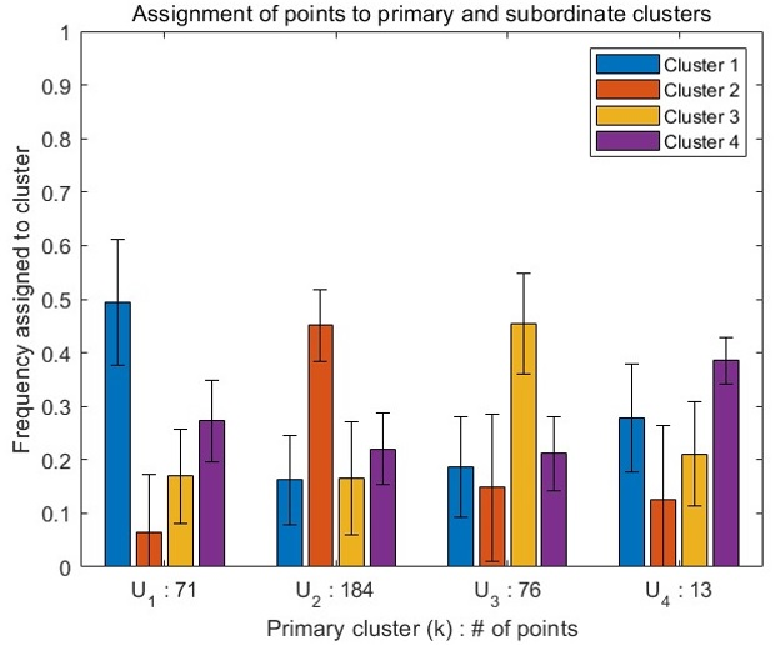}
}
\caption{Application of ERICA to the VDX (full) dataset with $p=22,283$ and $t=344$. We used k-means with $K=4$ and $B=200$ MCSS iterations. The PCSPs (top) illustrate spillover between clusters and their relative proximity while the ICAH (bottom) shows $U_4$ as the most tenuous of the four clusters. In addition to having the fewest points assigned to it and the lowest CRI, $U_{4}$ exhibits larger spillover CRIs to its secondary clusters than the other primary clusters.}\label{figPCSP_VDX_full_1}
\end{figure*}

By plotting the metrics as a function of $K$, Figure \ref{figmetricsParmigdata1} provides a collective comparison between the replicability attained with all of the genes and 3G. 
For Mainz, Transbig, and VDX, with the exception of HC-SL, the groupings are more stable with 3G than the full set. 
This is seen across $K \in \{2, \ldots, 8 \}$ rather than only at the $K^{*}$, and indicates the efficacy of the 3G assignment yielding more tenable groups that are replicated at a higher rate. 
The PCSPs and ICAH for the VDX dataset with k-means and $K=4$ are shown in Figure \ref{figPCSP_VDX_full_1}. 
We note that $U_{1}$ is the most replicable cluster according to the CRI $(CRI(U_{1})=0.493)$.
The points primarily assigned to $U_1$ are closest to (i.e. most commonly assigned to) the secondary group $U_4$. 
While consisting of the fewest number of primary data points, $U_4$ is the least stable cluster $(CRI(U_{4})=0.385)$ with the highest proportion of its points assigned to $U_1$ (i.e. $U_{4 \rightarrow 1}=0.277$). 
In light of the PCSP of $U_4$ and the ICAH values associated with this group, it is reasonable to 
question if this is a stable and meaningful cluster. 
When provided with Figure \ref{figPCSP_VDX_full_1}, the data analyst (or collector) should reconsider a number of factors. 
Firstly, whether to favor fewer (e.g. three) groups existing in the data. 
Secondly, the clustering technique $\mathcal{A}$ may not have been an appropriate choice for $\mathcal{X}$. 
Thirdly, the experimental conditions, components, or equipment that gave rise to the results may need to be evaluated. 
This is not a comprehensive list of the factors to question when viewing such results. 
For instance, a sufficient amount of data might not have been collected to represent the phenomenon that the scientist is expecting. 

It is informative to compare the PCSPs and ICAH attained for the synthetic dataset (Figure \ref{figPCSP_GaussMix4_100dim_highvar_1}) and those for the VDX scenario (Figure \ref{figPCSP_VDX_full_1}). 
There are a few caveats with the comparison. 
Aside from Gaussian mixture model clustering, k-means is the theoretically best $\mathcal{A}$ that could have been used for the synthetic data. 
Whereas from Table \ref{winningCRI_breast1} we note that k-means did not yield the highest ERICA statistic for VDX (full). 
More importantly, despite the clusters' overlap, it is known that $K=4$ is a ground truth for the synthetic dataset. 
It is believed that there are four breast cancer subtypes, but this does not constitute a ground truth and the analysis remains exploratory since there is inter-subject variability. 
%
The synthetic data scenario was a challenging clustering task with the groups overlapping in every dimension and a uniform number of points were assigned to them so that no one or two clusters could dominate. 
Nevertheless, a comparison of the ICAHs in Figure \ref{figPCSP_GaussMix4_100dim_highvar_1} and Figure \ref{figPCSP_VDX_full_1} reveals a large disparity in the dominance of the four primary clusters over the secondary groups. 
This speaks to the complexity and challenge of the biological data despite there being beliefs about the underlying structure. 
Previous clustering work has not reported a clear consensus about the number of breast cancer molecular subtypes, and analyzed robustness of classifiers for either three or four subtypes \citep{haibekains2012}. 
The CR analysis in \citet{haibekains2012} discovered that in general $K^{*}=3$ with the 3G subset while $K^{*}=4$ for PAM50 (50 genes). 
This indicates the features as highly heterogeneous, and the results being very sensitive to the number and choice of genes. 
Our results in Table \ref{winningCRI_breast1} and Table \ref{Fintalparmig1_res1} do not reflect the prior findings. 
Namely, a conservative interpretation of our results suggests two tumor subtypes, whereas a more exploratory interpretation supports six or possibly seven subtype groups.

\section{Relation to Prior Works and Future Avenues}

Clustering has been an active area of research since the 1960s, resulting in an extensive body of literature. 
%
%
The assessment of replicability began relatively recently and has been more limited in scope. 
The PS method uses k-fold cross-validation to estimate the number of groups. 
The metric considers the pairwise co-memberships to avert the issue of requiring an identity for the clusters containing the observations. 
%
Rather than restricting attention to pairwise co-membership, ERICA associates each observation with a cluster across MCSS iterations. 
%
%
%
Our PCSPs identify observations that are not consistently assigned to their primary clusters across MCSS iterations. 
The PS can be calculated for individual observations, but doing so requires a separate computation. 
In ERICA, the replicability of individual observations is aggregated into global replicability statistics. 
The GS requires the specification of a reference distribution so $B$ reference datasets must be generated. 
The ensuing calculation of the GS is dependent on the reference, and the metric was primarily intended for well-separated clusters. 
Neither PS nor GS provides an account of cluster closeness. 

The presented replicability metrics are computed differently from the silhouette introduced in \citet{rousseeuw1987}. 
The silhouette is calculated based on the average dissimilarity of data points within their assigned cluster and their dissimilarity with points in the second-best cluster assignment. 
Similar to the CRI metrics, the silhouette provides per-observation and per-cluster resolution. 
By definition, the metric identifies the second-best cluster for a data point, but it does not provide a ranking of clusters for data points as attained via the ICAH that is derived from a CLAM. 
This is important if one is interested in the least appropriate cluster for an observation. 
The silhouette metric has been used to derive a clustering algorithm \citep{batool2021}. 
Its use as well as the implementation of silhouette-based clustering techniques in the ERICA platform is a future investigation. 

Our consideration of replicability on a per-cluster basis has been motivated by works such as \citet{smolkin2003} and \citet{hennig2007}. 
The authors of \citet{smolkin2003} also perform MCSS, but measure per-cluster stability via the proportion of iterations for which the groups overlap. 
While their sensitivity measure is at the per-cluster level, it does not directly provide the stability of individual data points. 
In \citet{hennig2007} the author considers the Jaccard coefficient between clusters formed via the entire dataset and those formed on bootstrap samples. 
The ERICA statistic is different from a Jaccard coefficient as we compute the occurrence and cluster assignment of individual data points to evaluate replicability. 
The Clest method \citep{dudoit2002} shares several similarities to ERICA by performing iterative sampling WOR to assess CR. 
However, the technique requires a classification step, and the generation of null hypotheses to determine cluster significance. 
Clest scrutinizes the number of groups in the data, but not the grouping of the data points or the proximity of the prospective clusters. 

The metrics considered in works such as \citet{masoero2023} address the stability of the clustering solution, but do not account for replicability on a per-cluster basis. 
The authors study which clusters are the most and least replicable, but this was done by an algorithm to study "local replicability" rather than via metrics that are aggregated into more comprehensive measures in one pipeline. 
Furthermore, they do not provide a quantitative measure of the proximity of the discovered clusters. 
Another recent study \citep{hennig2022} has considered metrics in addition to what has been used in \citet{masoero2023}. 
The author performs a comprehensive analysis of clustering techniques on a wide set of data to assess the comparative performances of the algorithms. 
The nine clustering techniques considered, however, are not as diverse as those considered in \citet{masoero2023}. 
Also, accuracy rather than replicability was the focus of the investigation since the metrics were computed using known cluster labels that accompanied the datasets. 

In this work, ERICA has consisted of three clustering techniques with the results of each technique weighted equally for the purpose of making statements about the structure in the evaluated datasets. 
%
%
Developing principled weighting schemes for combining clustering methods is a future avenue of this research. 
Furthermore, the effects of the parameter settings should be explored over a broader range. 
For instance, via $P=80\%$ we have used an 80-20 split to form the cluster boundaries at each MCSS iteration, and then group the held-out data. 
Different values of $P$ may lead to different conclusions.  

\begin{algorithm}
\caption{Cluster number ($K^{*}$) selection with ERICA.}\label{Pipeline2}
\begin{algorithmic}[1]
\State \textbf{Input:} Metric for considered $K$ values $\{\mathcal{M}_{K} : K = 2, \ldots, K^{\max}\}$. Cluster-level metrics for considered $K$ values $\{\mathcal{M}_{2}(k): k=1,2\}, \, \ldots, \, \{\mathcal{M}_{K^{\max}}(k): k=1, \ldots, K^{\max}\}$. 
\State $K^{*} \gets 2$  \quad \% initialize 
\For{$K = 3$ to $K^{\max}$}
\If{${\rm NA} \notin \{\mathcal{M}_{K}(k)\}$} \quad \% is violated if $\exists \,\, k \geq 1 : X_{k}=0$
\If{$\mathcal{M}_{K} \geq \mathcal{M}_{K-1}$} 
    \State $K^{*} \gets K$ 
\EndIf 
\EndIf 
\EndFor
\State \Return $K^{*}$.
\end{algorithmic}
\end{algorithm}

\section{Conclusion and Discussion}

Although a ubiquitous practice, used to make decisions, discover structure, and verify results, clustering remains a heuristic that is often not rigorously scrutinized. 
There are several reasons for this, including the lack of a framework for prospective evaluation and the absence of appropriate quantitative metrics. 
Furthermore, it is an unsupervised learning technique with no ground truth. 
ERICA is a framework that analyzes whether a dataset contains groups of observations that are reproducibly discerned. 
%
%
The clustering techniques constitute one component of the evaluation pipeline. 
Through the introduction of several metrics, ERICA provides a quantitative means of determining how many groups of observations can be reliably discovered in a dataset. 
The ERICA statistic provides a single-number measure of the most replicable assignment and whether it is a reliable clustering solution. 
For every grouping, measures of similarity among the discovered clusters are available, enabling the identification of groups that are highly similar to one another. 
The evaluation platform also provides data-point-level resolution through the CLAM. 
This reveals observations that are unstable (i.e. not replicable), potentially indicating artifacts or observations located near the boundaries of multiple clusters. 
Given the heterogeneity of solutions produced by different clustering algorithms, multiple techniques can be incorporated into the evaluation platform to assess whether consensus emerges among methods for datasets containing reproducibly clustered observations. 

We evaluated ERICA on a challenging synthetic dataset containing overlapping clusters under a range of conditions to assess its operation and performance. 
The results were consistent with the known ground truth and the expected trends present in the generated data. 
This enabled us to apply ERICA to genomic data from breast cancer tumors, for which the ground truth is unknown. 
Some of the observed trends were consistent with those reported in previous studies. 
However, many of our findings differ from prevailing beliefs about the genomic data and 
raise questions regarding the reproducibility of the clustering structure reported in these datasets. 
While these findings are noteworthy, they represent an initial step. 
The groundwork has been laid for future applications of ERICA, including large-scale studies of clustering replicability and structure in complex datasets. 
The code for ERICA is publicly available at \url{https://github.com/sorooshyari/ERICA}. 



\acks{The authors thank David Donoho, Jerome Friedman, and Brad Efron for very helpful discussions and comments.}


\newpage

\appendix
\section{An Illustrative Example of ERICA}
\label{app:theorem}


\renewcommand{\thetable}{\thesection.\arabic{table}}
\renewcommand{\thefigure}{\thesection.\arabic{figure}}

\setcounter{table}{0}
\setcounter{figure}{0}

\begin{figure}[h]
\begin{center}
\epsfig{figure=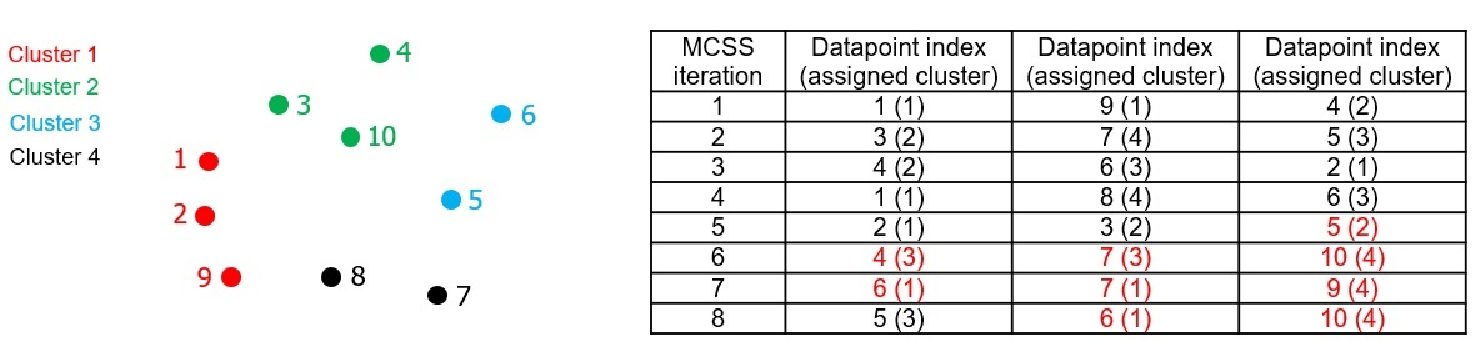, height=1.35in}
\end{center}
\caption{A clustering example with $t=10$, $p=2$, and $K=4$ groups. The dataset and its ground truth labels are shown (left). Applying MCSS across $B=8$ iterations together with an arbitrary clustering algorithm yields the results shown in the table (right). With $P=70\%$, the clustering results for the $m=3$ held-out data points are shown. The red entries in the table correspond to mis-clustered observations.}
\label{toyexamplefig1a}
\end{figure}
Consider a dataset of $t=10$ points shown in Figure \ref{toyexamplefig1a}. 
MCSS is performed for $B=8$ iterations, with clustering technique $\mathcal{A}$ applied to $P = 70 \%$ of the dataset at each iteration. 
We assume that the ground-truth cluster identities are known, as shown in Figure \ref{toyexamplefig1a}. 
Using Algorithm 1, $\mathcal{A}$ is applied for $B=8$ MC iterations with $n=7$ observations used to determine the cluster boundaries. 
The clustering results for the remaining $30\%$ of the data that are held-out ($m=3$) are shown in Figure \ref{toyexamplefig1a} for each MCSS iteration. 
The red entries in the table correspond to clustering errors identified using the ground truth, although knowledge of the ground truth is not required by ERICA.

From the group assignments and the identities of the held-out observations, the $10 \times 4$ CLAM can be constructed as 
\be
\Amat  = 
\begin{bmatrix}
2 & 2 & 0 & 0 & 0 & 2 & 1 & 0 & 1 & 0 \\
0 & 0 & 2 & 2 & 1 & 0 & 0 & 0 & 0 & 0 \\
0 & 0 & 0 & 1 & 2 & 2 & 1 & 0 & 0 & 0 \\
0 & 0 & 0 & 0 & 0 & 0 & 1 & 1 & 1 & 2 
\end{bmatrix}^{\top}
\ee
%
and compute
\begin{eqnarray}
{\rm Sum}(i) &=& \sum_{k=1}^{4} \Amat(i,k) \,\, : \,\, i = 1,2, \ldots, 10 \\ 
{\rm \textbf{Sum}} &=& [2 \quad 2 \quad 2 \quad 3 \quad 3 \quad 4 \quad 3 \quad 1 \quad 2 \quad 2]. \nonumber
\end{eqnarray}
Via (\ref{MaxandIdxeq1}) and (\ref{MaxandIdxeq2}) we shall have
\begin{eqnarray}
{\rm \textbf{Idx}}_{c}(1) &=& [1 \quad 1 \quad 2 \quad 2 \quad 3 \quad 1 \quad 1 \quad 4 \quad 1 \quad 4]\nonumber\\ 
{\rm \textbf{Max}}_{c}(1) &=& [2 \quad 2 \quad 2 \quad 2 \quad 2 \quad 2 \quad 1 \quad 1 \quad 1 \quad 2]\nonumber\\ 
{\rm \textbf{Idx}}_{c}(2) &=& [0 \quad 0 \quad 0 \quad 3 \quad 2 \quad 3 \quad 3 \quad 0 \quad 4 \quad 0]\nonumber\\ 
{\rm \textbf{Max}}_{c}(2) &=& [0 \quad 0 \quad 0 \quad 1 \quad 1 \quad 2 \quad 1 \quad 0 \quad 1 \quad 0]\nonumber\\ 
{\rm \textbf{Idx}}_{c}(3) &=& [0 \quad 0 \quad 0 \quad 0 \quad 0 \quad 0 \quad 4 \quad 0 \quad 0 \quad 0]\nonumber\\ 
{\rm \textbf{Max}}_{c}(3) &=& [0 \quad 0 \quad 0 \quad 0 \quad 0 \quad 0 \quad 1 \quad 0 \quad 0 \quad 0]\nonumber \\
{\rm \textbf{Idx}}_{c}(4) &=& \bf{0}\nonumber \\ 
{\rm \textbf{Max}}_{c}(4) &=& \bf{0}.
\end{eqnarray} 
At this point, the normalized sorted frequency matrix and index matrix can be formed as 
\be
\Amat_{S}(i, j) = \frac{{\rm Max}_{c}(i,j)}{{\rm Sum}(i)} = 
\begin{bmatrix}
1 & 0 & 0 & 0\\
1 & 0 & 0 & 0\\
1 & 0 & 0 & 0\\
2/3 & 1/3 & 0 & 0\\
2/3 & 1/3 & 0 & 0\\
1/2 & 1/2 & 0 & 0\\
1/3 & 1/3 & 1/3 & 0\\
1 & 0 & 0 & 0\\
1/2 & 1/2 & 0 & 0\\
1 & 0 & 0 & 0
\end{bmatrix}
\ee
and
\be
\Amat_{I}(i, j) 
= \left[
  \begin{array}{cccc}
     &  &     &    \\
    {\rm \textbf{Idx}}_{c}(1)    &  {\rm \textbf{Idx}}_{c}(2)  &  {\rm \textbf{Idx}}_{c}(3) &  {\rm \textbf{Idx}}_{c}(4)  \\
    &  &     &    
  \end{array}
\right] 
= 
\begin{bmatrix}
1 & 0 & 0 & 0\\
1 & 0 & 0 & 0\\
2 & 0 & 0 & 0\\
2 & 3 & 0 & 0\\
3 & 2 & 0 & 0\\
1 & 3 & 0 & 0\\
1 & 3 & 4 & 0\\
4 & 0 & 0 & 0\\
1 & 4 & 0 & 0\\
4 & 0 & 0 & 0
\end{bmatrix}.
\ee
In the above matrices, entries with index value "0" denote null (i.e. NA) entries. 

\begin{figure}[t]
\centering

\subfigure[]{
  \includegraphics[width=0.8\linewidth]{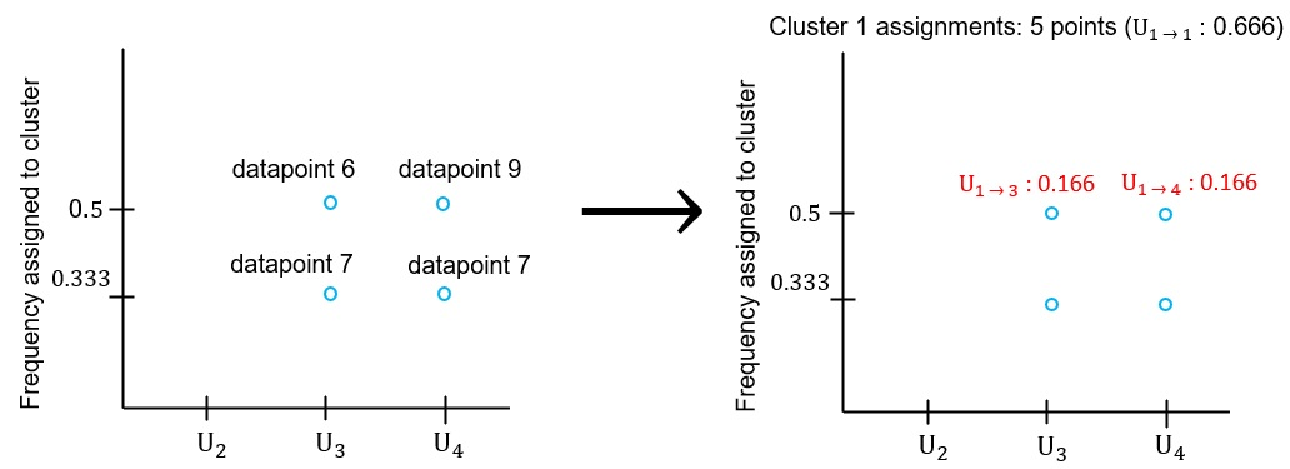}
  \label{fig:14a}
}

\vspace{0.2in}

\subfigure[]{
  \includegraphics[width=0.8\linewidth]{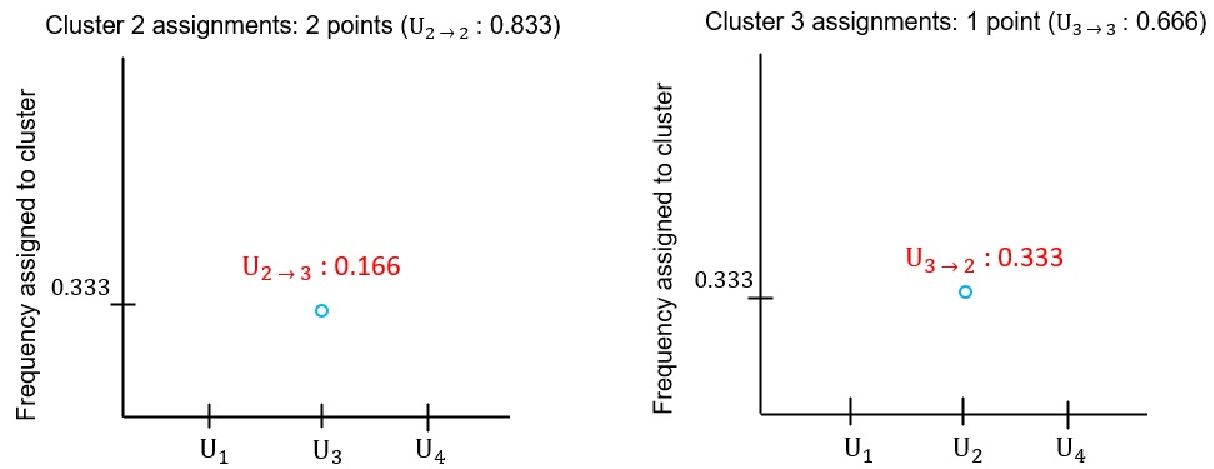}
  \label{fig:14b}
}

\vspace{0.2in}

\subfigure[]{
  \includegraphics[width=0.45\linewidth]{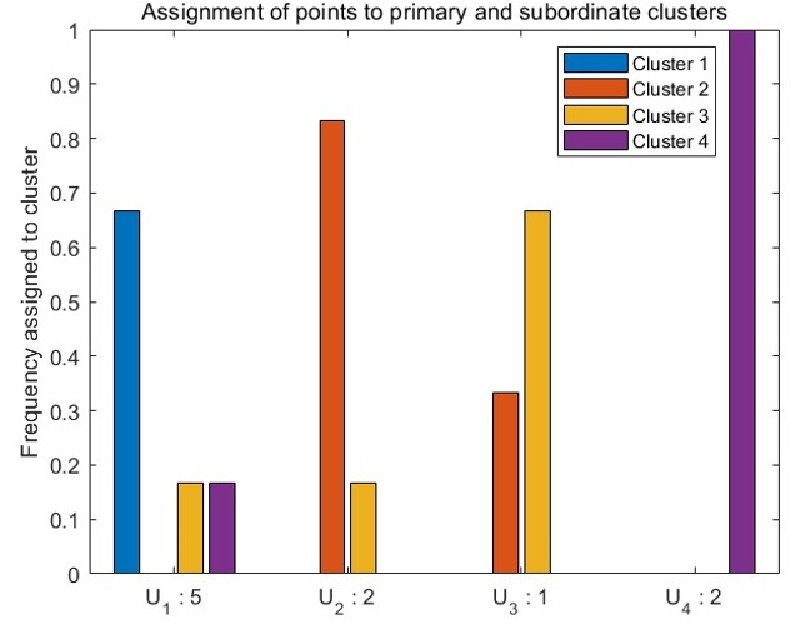}
  \label{fig:14c}
}

\caption{Formation of PCSPs and ICAH. (a) The PCSP corresponding to the data points assigned most frequently to cluster 1. (b) The PCSP of cluster 2 is constructed by noting $V_{2,3}=\{1/3\}$ and $CRI(U_{2 \rightarrow 3})=0.166$. Similarly, the PCSP of cluster 3 is constructed from $V_{3,2}=\{1/3\}$ and $CRI(U_{3 \rightarrow 2})= 1/3$. (c) The ICAH provides a holistic view of the clustering replicability across the identified groups.}
\label{figToyComposite1}
\end{figure}

Furthermore, the analysis could have been terminated at $f=3$ without altering any of the conclusions, rather than proceeding to $f=K=4$. 
We now form the sets $C_{k,j}, V_{k,j}: k,j=1,\ldots, 4$ according to (\ref{Eqsetckk1})-(\ref{Eqsetvkj1}). 
The observations that have been assigned the most number of times to cluster 1 are reflected via $C_{1,1}$. 
It is not difficult to observe that 
\begin{eqnarray}
C_{1,1} &=& \{i: \Amat_{I}(i,1)=1\} = \{1, 2, 6, 7, 9\} \\
V_{1,1} &=& \{\Amat_{S}(i \in C_{1,1}, 1)\} = \{1, 1, 1/2, 1/3, 1/2\} 
\end{eqnarray}
and we proceed to form the sets $C_{1,j}: j=2,3,4$. 
For $C_{i,j} : i=1,2$, the columns $j=2,3,4$ of $\Amat_{I}$ reflect that data points $t=1,2$ have not been assigned to any other cluster. 
With $i=6$, however, we note that via column $j=2$ we have $\Amat_{I}(6,2)=3$ at frequency $\Amat_{S}(6,2)=1/2$. 
Similarly, with $i=7$ examination of columns $j=2,3,4$ of $\Amat_{I}$ yield the following assignments: for $j=2$ we have $\Amat_{I}(7,2)=3$ at $\Amat_{S}(7,2)=1/3$, while for $j=3$ we have $\Amat_{I}(7,3)=4$ at $\Amat_{S}(7,3)=1/3$. 
Lastly, with $i=9$ we note that for $j=2$ we have $\Amat_{I}(9,2)=4$ at $\Amat_{S}(9,2)=1/2$. 
Collecting these assignments yields the following sets
\begin{eqnarray}
C_{1,2} &=& \{\varnothing\} \quad\quad \,\,\,\,\, V_{1,2} = \{\varnothing\}\\
C_{1,3} &=& \{6, 7\} \quad\quad V_{1,3} = \{1/2, 1/3\}\\
C_{1,4} &=& \{7, 9\} \quad\quad V_{1,4} = \{1/3, 1/2\} . 
\end{eqnarray}
We are ready to form the PCSP for cluster 1. 
The set $V_{1,1}$ contains the 5 observations most frequently assigned to cluster 1, yielding $CRI(U_{1}) = (1 + 1 + 1/2 + 1/3 + 1/2)/5 = 2/3$. 
This value reflects the assignment $U_{1 \rightarrow 1}$ shown in Figure \ref{figToyComposite1}(a) alongside what we refer to as the spillover assignments $U_{1 \rightarrow j \neq 1}$. 
The latter quantifies the frequency of observations that were attributed the most number of times to cluster 1 but were also assigned (i.e. spilled-over) to other clusters, namely $U_3$ and $U_4$ for this example. 
In the PCSP, the frequency $CRI(U_{1 \rightarrow 3}) = CRI(U_{1 \rightarrow 4}) = 0.166$ follows from the calculation in $(\ref{Eqspilov1})$ for $V_{1,3}$ and $V_{1,4}$, respectively. 

We proceed to cluster 2 and note that 
\begin{eqnarray}
C_{2,2} &=& \{i: \Amat_{I}(i,1)=2\} = \{3, 4\} \label{EqC221} \\
V_{2,2} &=& \{\Amat_{S}(i \in C_{2,2}, 1) \} = \{1, 2/3\}, 
\end{eqnarray}
which enables us to construct the sets $C_{2,j}, V_{2,j}: j=1,3,4$. 
With $i=3$ in (\ref{EqC221}) we examine columns $j=2, 3, 4$ of $\Amat_{I}$ and since data point 3 has not been assigned to any other cluster, it is not included in $C_{2,1}$, $C_{2,3}$, or $C_{2,4}$. 
For $i=4$, the same columns of $\Amat_I$ indicate that via $j=2$ we have $\Amat_{I}(4,2)=3$ at frequency $\Amat_{S}(4,2)=1/3$. 
This leads to the assignments
\begin{eqnarray}
C_{2,1} &=& \{\varnothing\} \quad\quad \,\,\,\,\, V_{2,1} = \{\varnothing\}\\
C_{2,3} &=& \{4\} \quad\quad V_{2,3} = \{1/3\}\\
C_{2,4} &=& \{\varnothing\} \quad\quad \,\,\,\,\, V_{2,4} = \{\varnothing\} 
\end{eqnarray}
and the rather simplistic PCSP for cluster 2 shown in Figure \ref{figToyComposite1}(b). 
The cluster 2 replicability index is determined from $CRI(U_{2}) = (1 + 2/3)/2=0.833$ while the spillover frequency for $U_{2 \rightarrow 3}$ follows from $CRI(U_{2 \rightarrow 3}) = (1/3)/2 = 0.166$. 
For cluster 3 it is noted 
\begin{eqnarray}
C_{3,3} &=& \{i: \Amat_{I}(i,1)=3\} = \{5\} \\
V_{3,3} &=& \{\Amat_{S}(i \in C_{3,3}, 1) \} = \{2/3\}. 
\end{eqnarray}
Examining columns $j=2,3,4$ of $\Amat_I$, for $i=5$ we observe that $\Amat_{I}(5,2)=2$ at frequency $\Amat_{S}(5,2)=1/3$. 
In this case we have the sets
\begin{eqnarray}
C_{3,2} &=& \{5\} \quad\quad\quad\quad\quad \, V_{3,2} = \{1/3\}\\
C_{3,1} &=& C_{3,4} = \{\varnothing\} \quad\quad V_{3,1} = V_{3,4} = \{\varnothing\} 
\end{eqnarray}
and the PCSP in Figure \ref{figToyComposite1}(b). 
The cluster 3 replicability is computed from $CRI(U_{3})=(2/3)/1 = 2/3$, and the spillover occurs into one cluster via $CRI(U_{3 \rightarrow 2}) = (1/3)/1= 1/3$ since there is one data point that has been assigned to cluster 3 at the highest frequency. 
This also illustrates that, in general, $CRI(U_{i \rightarrow j}) \neq CRI(U_{j \rightarrow i})$. 
Lastly, in the case of cluster 4 we note that 
\begin{eqnarray}
C_{4,4} &=& \{i: \Amat_{I}(i,1)=4\} = \{8, 10\} \\
V_{4,4} &=& \{\Amat_{S}(i \in C_{4,4}, 1) \} = \{1, 1\} 
\end{eqnarray}
and $CRI(U_{4}) = (1 + 1)/2 = 1$. 
None of the observations most frequently assigned to cluster 4 were assigned to a different cluster, thus 
\be
C_{4,1} = C_{4,2} = C_{4,3} = \{\varnothing\} \quad\quad V_{4,1} = V_{4,2} = V_{4,3} = \{\varnothing\}. 
\ee
There is no point in viewing a PCSP for cluster 4 since it would be unoccupied. 
The CRI values provide a per-cluster measure of the certainty with which observations belong to their hypothesized cluster. 
The four PCSPs provide a cluster-level visualization of uncertain cluster memberships and the extent of spillover between clusters.
Viewing the collection of data points in a PCSP further illustrates the degree of prospective overlap with other groups. 
The ICAH in Figure \ref{figToyComposite1}(c) provides a summary of the clustering replicability attained across the $K=4$ groups in this example. 
Averaging the per-cluster CRI values yields an ERICA statistic of 0.791. 


\section{Per-cluster Replicability Results for Gaussian Mixture Data}
\label{app:theorem22}

The following tables report the metrics obtained by applying ERICA to different Gaussian mixture datasets. 
In every case we used $B=200$ MCSS iterations and $P=80\%$. 
The format of the panels is described in Table \ref{template_for_mytable1}. The bold entries represent the metric (ERICA statistic, WCRI, or TWCRI) associated with the most appropriate number of clusters ($K^{*}$) in the dataset. 

\renewcommand{\thetable}{\thesection.\arabic{table}}
\renewcommand{\thefigure}{\thesection.\arabic{figure}}

\setcounter{table}{0}
\setcounter{figure}{0}

\begin{table}[ht]
    \centering
    \caption{Values of the ERICA statistic with $t=10,000$. K-means clustering was used.}
\begin{adjustbox}{width=0.95\textwidth}
    \begin{tabular}{lcccccc}
        \toprule
        & K = 2 & K = 3 & K = 4 & K = 5 & K = 6 \\
        \midrule
        p = 100 & 1 & 0.816 & \textbf{0.994} & 0.866 & 0.696 \\
              & (1, 1) & (0.983, 0.64, 0.825) & (0.997, 0.987, 0.995, 0.997) & (0.872, NA, 0.875, 0.859, 0.86) & (0.751, NA, 0.77, 0.736, 0.496, 0.731) \\
        p = 200 & 1 & 0.792 & \textbf{0.993} & 0.843 & 0.638 \\
              & (1, 1) & (0.964, 0.628, 0.785) & (0.994, 0.992, 0.989, 1) & (0.877, 0.875, 0.85, NA, 0.772) & (0.76, 0.75, 0.681, 0.48, 0.519, 0.643) \\
        p = 400 & 1 & 0.769 & \textbf{0.994} & 0.86 & 0.682 \\
              & (1, 1) & (0.95, 0.602, 0.756) & (0.994, 1, 0.985, 1) & (0.889, 0.871, 0.84, NA, 0.843) & (NA, 0.762, 0.755, 0.727, 0.466, 0.701) \\
        p = 100, Toeplitz & 1 & 0.814 & \textbf{0.997} & 0.872 & 0.664 \\
              & (1, 1) & (0.997, 0.659, 0.787) & (1, 0.994, 1, 0.997) & (0.844, 0.887, NA, 0.87, 0.887) & (0.751, 0.502, 0.722, 0.503, 0.746, 0.76) \\
        p = 200, Toeplitz & 1 & 0.8 & \textbf{0.988} & 0.76 & 0.645 \\
              & (1, 1) & (0.769, 0.67, 0.963) & (0.992, 0.982, 0.979, 1) & (0.872, 0.671, 0.53, 0.879, 0.848) & (0.721, 0.625, 0.539, 0.505, 0.745, 0.74) \\
        p = 400, Toeplitz & 1 & 0.815 & \textbf{0.997} & 0.863 & 0.663 \\
              & (1, 1) & (0.968, 0.661, 0.817) & (0.997, 1, 0.994, 1) & (0.863, 0.874, 0.868, 0.849, NA) & (0.747, 0.704, 0.505, 0.769, 0.714, 0.544) \\
        p = 100, $\uparrow$ variance & 1 & 0.878 & \textbf{0.966} & 0.821 & 0.64 \\
              & (1, 1) & (0.973, 0.665, 0.997) & (0.987, 0.95, 0.952, 0.977) & (0.876, 0.812, 0.82, NA, 0.778) & (0.735, 0.515, 0.752, 0.704, 0.479, 0.659) \\
        p = 200, $\uparrow$ variance & 1 & 0.874 & \textbf{0.945} & 0.835 & 0.667 \\
              & (1, 1) & (0.963, 0.661, 0.999) & (0.966, 0.91, 0.941, 0.963) & (0.853, NA, 0.823, 0.844, 0.822) & (0.759, NA, 0.463, 0.68, 0.733, 0.703) \\
        p = 400, $\uparrow$ variance & 1 & 0.879 & \textbf{0.941} & 0.831 & 0.632 \\
              & (1, 1) & (0.972, 0.666, 0.999) & (0.967, 0.91, 0.922, 0.966) & (0.892, NA, 0.794, 0.828, 0.813) & (0.733, 0.484, 0.46, 0.69, 0.725, 0.7) \\
        \bottomrule
    \end{tabular}
    \end{adjustbox}
    \label{Kmeanstable1}
\end{table}

\begin{table}[ht]
    \centering
    \caption{WCRI and TWCRI values with $t=10,000$. K-means clustering was used.}
\begin{adjustbox}{width=0.95\textwidth}
    \begin{tabular}{lcccccc}
        \toprule
        & K = 2 & K = 3 & K = 4 & K = 5 & K = 6 \\
        \midrule
        p = 100 & \textbf{0.5}, 1 & 0.259, 0.779 & 0.248, \textbf{0.992} & 0.216, 0.865 & 0.149, 0.745 \\
              & (0.497, 0.503) & (0.247, 0.288, 0.244) & (0.25, 0.242, 0.248, 0.252) & (0.219, NA, 0.214, 0.214, 0.218) & (0.188, NA, 0.189, 0.183, 9.9e-5, 0.185) \\
        p = 200 & \textbf{0.5}, 1 & 0.254, 0.764 & 0.247, \textbf{0.991} & 0.21, 0.84 & 0.149, 0.745 \\
              & (0.497, 0.503) & (0.244, 0.263, 0.257) & (0.249, 0.243, 0.246, 0.253) & (0.22, 0.214, 0.211, NA, 0.195) & (0.191, 0.184, 0.169, 2.4e-4, 0.006, 0.154) \\
        p = 400 & 0.5, 1 & 0.246, 0.739 & \textbf{0.248}, \textbf{0.992} & 0.214, 0.858 & 0.146, 0.734 \\
              & (0.497, 0.503) & (0.249, 0.262, 0.228) & (0.249, 0.245, 0.245, 0.253) & (0.223, 0.213, 0.209, NA, 0.213) & (NA, 0.191, 0.185, 0.181, 4.6e-5, 0.177) \\
        p = 100, Toeplitz & \textbf{0.5}, 1 & 0.262, 0.786 & 0.249, \textbf{0.996} & 0.217, 0.869 & 0.146, 0.734 \\
              & (0.497, 0.503) & (0.252, 0.267, 0.267) & (0.251, 0.244, 0.249, 0.252) & (0.212, 0.217, NA, 0.216, 0.224) & (0.188, 0.006, 0.168, 0.002, 0.182, 0.192) \\
        p = 200, Toeplitz & \textbf{0.5}, 1 & 0.259, 0.779 & 0.246, \textbf{0.987} & 0.163, 0.817 & 0.116, 0.701 \\
              & (0.497, 0.503) & (0.281, 0.254, 0.244) & (0.249, 0.241, 0.244, 0.253) & (0.219, 0.163, 0.001, 0.219, 0.215) & (0.181, 0.131, 0.018, 6.5e-4, 0.184, 0.187) \\
        p = 400, Toeplitz & \textbf{0.5}, 1 & 0.261, 0.783 & 0.248, \textbf{0.995} & 0.215, 0.861 & 0.121, 0.728 \\
              & (0.497, 0.503) & (0.243, 0.295, 0.245) & (0.25, 0.245, 0.247, 0.253) & (0.216, 0.214, 0.216, 0.215, NA) & (0.187, 0.171, 7.5e-4, 0.191, 0.175, 0.004) \\
        p = 100, $\uparrow$ variance & \textbf{0.5}, 1 & 0.276, 0.83 & 0.241, \textbf{0.965} & 0.205, 0.82 & 0.118, 0.709 \\
              & (0.497, 0.503) & (0.26, 0.316, 0.254) & (0.248, 0.233, 0.237, 0.247) & (0.22, 0.199, 0.204, NA, 0.197) & (0.184, 5.1e-5, 0.184, 0.175, 0.001, 0.165) \\
        p = 200, $\uparrow$ variance & \textbf{0.5}, 1 & 0.276, 0.828 & 0.235, \textbf{0.943} & 0.208, 0.834 & 0.143, 0.716 \\
              & (0.497, 0.503) & (0.262, 0.313, 0.253) & (0.242, 0.223, 0.234, 0.244) & (0.214, NA, 0.202, 0.21, 0.208) & (0.19, NA, 1.3e-4, 0.166, 0.182, 0.178) \\
        p = 400, $\uparrow$ variance & \textbf{0.5}, 1 & 0.277, 0.831 & 0.234, \textbf{0.939} & 0.207, 0.831 & 0.118, 0.71 \\
              & (0.497, 0.503) & (0.259, 0.319, 0.253) & (0.243, 0.223, 0.229, 0.244) & (0.224, NA, 0.195, 0.206, 0.206) & (0.184, 4.8e-5, 9.2e-5, 0.169, 0.18, 0.177) \\
        \bottomrule
    \end{tabular}
    \end{adjustbox}
    \label{Kmeanstable12}
\end{table}

\begin{table}[ht]
    \centering
    \caption{Values of the ERICA statistic with $t=10,000$. Hierarchical clustering with Ward linkage (HC-WL) was used.}
\begin{adjustbox}{width=0.95\textwidth}
    \begin{tabular}{lcccccc}
        \toprule
        & K = 2 & K = 3 & K = 4 & K = 5 & K = 6 \\
        \midrule
        p = 100 & 0.89 & 0.905 & \textbf{1} & 0.88 & 0.787 \\
              & (0.887, 0.894) & (0.981, 0.735, 1) & (1, 1, 1, 1) & (0.893, 0.919, NA, 0.839, 0.872) & (0.809, NA, 0.822, NA, 0.722, 0.796) \\
        p = 200 & 0.907 & 0.887 & \textbf{1} & 0.901 & 0.8 \\
              & (0.897, 0.918) & (0.953, 0.708, 1) & (1, 1, 1, 1) & (NA, 0.875, 0.917, 0.909, 0.903) & (NA, 0.774, 0.827, 0.812, NA, 0.788) \\
        p = 400 & 0.908 & 0.891 & \textbf{1} & 0.902 & 0.81 \\
              & (0.897, 0.919) & (0.971, 0.702, 1) & (1, 1, 1, 1) & (NA, 0.889, 0.903, 0.919, 0.898) & (NA, 0.785, 0.823, 0.841, NA, 0.791) \\
        p = 100, Toeplitz & 0.889 & 0.902 & \textbf{0.999} & 0.875 & 0.681 \\
              & (0.86, 0.918) & (0.993, 0.714, 0.999) & (1, 0.999, 0.999, 1) & (0.923, 0.813, NA, 0.893, 0.872) & (0.82, 0.701, 0.51, 0.795, 0.745, 0.519) \\
        p = 200, Toeplitz & 0.89 & 0.899 & \textbf{1} & 0.895 & 0.688 \\
              & (0.879, 0.901) & (0.971, 0.728, 1) & (1, 1, 1, 1) & (0.833, NA, 0.929, 0.923, 0.896) & (0.703, 0.521, 0.846, 0.811, 0.5, 0.751) \\
        p = 400, Toeplitz & 0.875 & 0.914 & \textbf{1} & 0.897 & 0.799 \\
              & (0.846, 0.905) & (0.999, 0.743, 1) & (1, 1, 1, 1) & (0.857, 0.918, NA, 0.911, 0.903) & (0.763, NA, 0.819, NA, 0.807, 0.808) \\
        p = 100, $\uparrow$ variance & 0.863 & 0.915 & \textbf{0.999} & 0.902 & 0.81 \\
              & (0.891, 0.836) & (0.991, 0.757, 0.999) & (0.999, 0.999, 0.999, 0.999) & (0.895, 0.925, NA, 0.917, 0.871) & (0.794, NA, 0.832, NA, 0.82, 0.795) \\
        p = 200, $\uparrow$ variance & 0.912 & 0.842 & \textbf{1} & 0.901 & 0.8 \\
              & (0.904, 0.921) & (0.758, 0.878, 0.89) & (1, 1, 1, 1) & (NA, 0.875, 0.917, 0.909, 0.903) & (NA, 0.774, 0.827, 0.812, NA, 0.788) \\
        p = 400, $\uparrow$ variance & 0.921 & 0.82 & \textbf{1} & 0.902 & 0.804 \\
              & (0.919, 0.923) & (0.728, 0.869, 0.865) & (1, 1, 1, 1) & (0.889, NA, 0.903, 0.919, 0.898) & (0.785, NA, 0.823, 0.841, NA, 0.787) \\
        \bottomrule
    \end{tabular}
    \end{adjustbox}
    \label{ACWardresults1}
\end{table}

\begin{table}[ht]
    \centering
    \caption{WCRI and TWCRI values with $t=10,000$. Hierarchical clustering with Ward linkage (HC-WL) was used.}
\begin{adjustbox}{width=0.95\textwidth}
    \begin{tabular}{lcccccc}
        \toprule
        & K = 2 & K = 3 & K = 4 & K = 5 & K = 6 \\
        \midrule
        p = 100 & \textbf{0.444}, 0.889 & 0.288, 0.866 & 0.249, \textbf{0.998} & 0.219, 0.879 & 0.196, 0.785 \\
              & (0.44, 0.449) & (0.257, 0.356, 0.253) & (0.251, 0.245, 0.249, 0.253) & (0.224, 0.225, NA, 0.209, 0.221) & (0.203, NA, 0.201, NA, 0.18, 0.201) \\
        p = 200 & \textbf{0.453}, 0.906 & 0.283, 0.849 & 0.249, \textbf{0.998} & 0.224, 0.899 & 0.199, 0.798 \\
              & (0.445, 0.461) & (0.266, 0.33, 0.253) & (0.251, 0.245, 0.249, 0.253) & (NA, 0.219, 0.225, 0.226, 0.229) & (NA, 0.194, 0.203, 0.202, NA, 0.199) \\
        p = 400 & \textbf{0.453}, 0.907 & 0.282, 0.848 & 0.249, \textbf{0.998} & 0.225, 0.9 & 0.202, 0.808 \\
              & (0.445, 0.462) & (0.26, 0.335, 0.253) & (0.251, 0.245, 0.249, 0.253) & (NA, 0.223, 0.221, 0.229, 0.227) & (NA, 0.197, 0.202, 0.209, NA, 0.2) \\
        p = 100, Toeplitz & \textbf{0.444}, 0.888 & 0.285, 0.856 & 0.249, \textbf{0.998} & 0.218, 0.874 & 0.127, 0.763 \\
              & (0.427, 0.461) & (0.253, 0.35, 0.253) & (0.251, 0.245, 0.249, 0.253) & (0.232, 0.199, NA, 0.222, 0.221) & (0.206, 0.169, 0.002, 0.198, 0.188, 6.2e-4) \\
        p = 200, Toeplitz & \textbf{0.444}, 0.889 & 0.286, 0.86 & 0.249, \textbf{0.998} & 0.223, 0.894 & 0.129, 0.774 \\
              & (0.436, 0.453) & (0.259, 0.348, 0.253) & (0.251, 0.245, 0.249, 0.253) & (0.209, NA, 0.228, 0.23, 0.227) & (0.175, 5.7e-4, 0.207, 0.202, 1e-4, 0.19) \\
        p = 400, Toeplitz & \textbf{0.437}, 0.875 & 0.29, 0.871 & 0.249, \textbf{0.998} & 0.224, 0.896 & 0.199, 0.797 \\
              & (0.42, 0.455) & (0.251, 0.367, 0.253) & (0.251, 0.245, 0.249, 0.253) & (0.215, 0.225, NA, 0.227, 0.229) & (0.191, NA, 0.201, NA, 0.201, 0.204) \\
        p = 100, $\uparrow$ variance & \textbf{0.431}, 0.862 & 0.292, 0.877 & 0.249, \textbf{0.998} & 0.225, 0.9 & 0.202, 0.808 \\
              & (0.443, 0.419) & (0.253, 0.371, 0.253) & (0.251, 0.245, 0.249, 0.253) & (0.225, 0.227, NA, 0.228, 0.22) & (0.199, NA, 0.204, NA, 0.204, 0.201) \\
        p = 200, $\uparrow$ variance & \textbf{0.456}, 0.912 & 0.275, 0.825 & 0.249, \textbf{0.998} & 0.224, 0.899 & 0.199, 0.798 \\
              & (0.449, 0.463) & (0.346, 0.254, 0.225) & (0.251, 0.245, 0.249, 0.253) & (NA, 0.219, 0.225, 0.226, 0.229) & (NA, 0.194, 0.203, 0.202, NA, 0.199) \\
        p = 400, $\uparrow$ variance & \textbf{0.46}, 0.92 & 0.266, 0.799 & 0.249, \textbf{0.998} & 0.225, 0.9 & 0.201, 0.807 \\
              & (0.456, 0.464) & (0.346, 0.234, 0.219) & (0.251, 0.245, 0.249, 0.253) & (0.223, NA, 0.221, 0.229, 0.227) & (0.197, NA, 0.202, 0.209, NA, 0.199) \\
        \bottomrule
    \end{tabular}
    \end{adjustbox}
    \label{ACWardresults2}
\end{table}

\begin{table}[ht]
    \centering
    \caption{Values of the ERICA statistic with $t=10,000$. Hierarchical clustering with single linkage (HC-SL) was used.}
\begin{adjustbox}{width=0.95\textwidth}
    \begin{tabular}{lcccccc}
        \toprule
        & K = 2 & K = 3 & K = 4 & K = 5 & K = 6 \\
        \midrule
        p = 100 & 0.848 & 0.864 & \textbf{1} & 0.906 & 0.918 \\
              & (0.846, 0.85) & (0.799, 0.882, 0.912) & (1, 1, 1, 1) & (0.958, 0.92, 0.873, NA, 0.873) & (0.969, 0.955, NA, 0.908, NA, 0.842) \\
        p = 200 & 0.826 & 0.817 & \textbf{1} & 0.87 & 0.849 \\
              & (0.835, 0.818) & (0.878, 0.67, 0.905) & (1, 1, 1, 1) & (0.893, NA, 0.85, 0.874, 0.864) & (0.889, NA, NA, 0.819, 0.831, 0.859) \\
        p = 400 & 0.854 & 0.889 & \textbf{1} & 0.814 & 0.771 \\
              & (0.862, 0.847) & (0.961, 0.892, 0.816) & (1, 1, 1, 1) & (0.87, 0.914, 0.835, 0.88, 0.571) & (0.873, 0.86, 0.834, 0.864, 0.428, NA) \\
        p = 100, Toeplitz & 0.894 & 0.765 & 0.773 & \textbf{0.907} & 0.851 \\
              & (0.892, 0.897) & (0.844, 0.709, 0.742) & (0.999, 0.975, 0.547, 0.572) & (0.945, 0.949, 0.88, 0.885, 0.88) & (0.953, 0.909, 0.904, 0.888, 0.557, 0.898) \\
        p = 200, Toeplitz & 0.787 & 0.875 & 1 & 0.823 & \textbf{0.827} \\
              & (0.77, 0.805) & (0.976, 0.654, 0.996) & (1, 1, 1, 1) & (0.889, 0.875, 0.875, 0.55, 0.929) & (0.884, 0.675, 0.894, 0.87, 0.684, 0.958) \\
        p = 400, Toeplitz & 0.805 & 0.828 & \textbf{1} & 0.884 & 0.841 \\
              & (0.794, 0.817) & (0.893, 0.696, 0.896) & (1, 1, 1, 1) & (0.885, 0.919, 0.924, 0.789, 0.904) & (0.899, 0.869, 0.617, 0.878, 0.868, 0.919) \\
        p = 100, $\uparrow$ variance & 0.999 & 0.741 & 0.658 & \textbf{0.662} & 0.623 \\
              & (0.999, NA) & (NA, 0.999, 0.483) & (0.512, 0.998, 0.463, 0.66) & (0.512, 0.998, 0.415, 0.741, 0.644) & (0.536, 0.998, 0.481, 0.5, 0.677, 0.551) \\
        p = 200, $\uparrow$ variance & 0.826 & 0.526 & 0.505 & \textbf{0.507} & 0.46 \\
              & (0.835, 0.818) & (0.424, 0.429, 0.727) & (0.378, 0.499, 0.739, 0.404) & (0.324, 0.555, 0.473, 0.753, 0.432) & (0.464, 0.445, 0.355, 0.364, 0.652, 0.485) \\
        p = 400, $\uparrow$ variance & 0.806 & 0.899 & \textbf{0.995} & 0.801 & 0.759 \\
              & (0.841, 0.772) & (1, 0.708, 0.991) & (0.994, 0.99, 1, 0.999) & (0.865, 0.894, 0.81, 0.895, 0.542) & (0.894, 0.82, 0.819, 0.864, 0.4, NA) \\
        \bottomrule
    \end{tabular}
    \end{adjustbox}
    \label{ACSLresults1}
\end{table}

\begin{table}[ht]
    \centering
    \caption{WCRI and TWCRI values with $t=10,000$. Hierarchical clustering with single linkage (HC-SL) was used.}
\begin{adjustbox}{width=0.95\textwidth}
    \begin{tabular}{lcccccc}
        \toprule
        & K = 2 & K = 3 & K = 4 & K = 5 & K = 6 \\
        \midrule
        p = 100 & \textbf{0.423}, 0.847 & 0.282, 0.847 & 0.249, \textbf{0.998} & 0.225, 0.903 & 0.229, 0.916 \\
              & (0.558, 0.289) & (0.396, 0.22, 0.231) & (0.251, 0.245, 0.249, 0.253) & (0.24, 0.225, 0.217, NA, 0.221) & (0.243, 0.234, NA, 0.226, NA, 0.213) \\
        p = 200 & \textbf{0.413}, 0.826 & 0.261, 0.785 & 0.249, \textbf{0.998}, & 0.217, 0.868 & 0.212, 0.848 \\
              & (0.428, 0.398) & (0.22, 0.317, 0.248) & (0.251, 0.245, 0.249, 0.253) & (0.224, NA, 0.208, 0.217, 0.219) & (0.223, NA, NA, 0.201, 0.207, 0.217) \\
        p = 400 & \textbf{0.425}, 0.851 & 0.289, 0.869 & 0.249, \textbf{0.998} & 0.174, 0.873 & 0.171, 0.856 \\
              & (0.31, 0.541) & (0.241, 0.22, 0.408) & (0.251, 0.245, 0.249, 0.253) & (0.218, 0.224, 0.208, 0.223, 5.7e-5) & (0.219, 0.211, 0.207, 0.219, 4.2e-5, NA) \\
        p = 100, Toeplitz & \textbf{0.447}, 0.894 & 0.252, 0.757 & 0.193, 0.772 & 0.182, \textbf{0.913} & 0.151, 0.91 \\
              & (0.443, 0.451) & (0.212, 0.187, 0.358) & (0.251, 0.239, 0.11, 0.172) & (0.237, 0.233, 8.8e-5, 0.22, 0.223) & (0.239, 0.223, 9e-5, 0.221, 1.1e-4, 0.227) \\
        p = 200, Toeplitz & \textbf{0.393}, 0.787 & 0.275, 0.825 & 0.249, 0.998 & 0.178, 0.89 & 0.149, \textbf{0.899} \\
              & (0.379, 0.408) & (0.258, 0.313, 0.254) & (0.251, 0.245, 0.249, 0.253) & (0.223, 0.214, 0.218, 5.5e-5, 0.235) & (0.222, 6.7e-5, 0.219, 0.216, 1.3e-4, 0.242) \\
        p = 400, Toeplitz & \textbf{0.404}, 0.808 & 0.268, 0.804 & 0.249, \textbf{0.998} & 0.181, 0.906 & 0.148, 0.89 \\
              & (0.286, 0.522) & (0.224, 0.309, 0.271) & (0.251, 0.245, 0.249, 0.253) & (0.222, 0.225, 0.23, 7.8e-5, 0.229) & (0.226, 0.213, 6.1e-5, 0.218, 8.6e-5, 0.233) \\
        p = 100, $\uparrow$ variance & 0.999, 0.999 & 0.499, 0.998 & \textbf{0.249}, 0.997 & 0.199, 0.997 & 0.166, \textbf{0.997} \\
              & (0.999, NA) & (NA, 0.998, 4.8e-5) & (5.1e-5, 0.997, 4.6e-5, 1.3e-4) & (5.1e-5, 0.997, 8.3e-5, 7.4e-5, 6.4e-5) & (5.3e-5, 0.997, 9.6e-5, 1e-4, 6.7e-5, 1.1e-4) \\
        p = 200, $\uparrow$ variance & \textbf{0.413}, \textbf{0.826} & 0.241, 0.724 & 0.179, 0.716 & 0.141, 0.705 & 0.1, 0.605 \\
              & (0.428, 0.398) & (8.4e-5, 0.002, 0.722) & (3.7e-5, 0.046, 0.67, 8e-5) & (3.2e-5, 0.13, 4.7e-5, 0.575, 1.7e-4) & (0.104, 8.9e-5, 7.8e-4, 1e-4, 0.5, 0.001) \\
        p = 400, $\uparrow$ variance & \textbf{0.395}, 0.79 & 0.284, 0.853 & 0.248, \textbf{0.994} & 0.172, 0.864 & 0.169, 0.848 \\
              & (0.233, 0.557) & (0.251, 0.347, 0.255) & (0.249, 0.243, 0.249, 0.253) & (0.217, 0.219, 0.202, 0.226, 5.4e-5) & (0.224, 0.201, 0.204, 0.219, 4e-5, NA) \\
        \bottomrule
    \end{tabular}
    \end{adjustbox}
    \label{ACSLresults12}
\end{table}

\begin{table}[ht]
    \centering
    \caption{Values of the ERICA statistic with $p=1,000$. K-means clustering was used.}
\begin{adjustbox}{width=0.95\textwidth}
    \begin{tabular}{lcccccc}
        \toprule
        & K = 2 & K = 3 & K = 4 & K = 5 & K = 6 \\
        \midrule
        t = 100 & 0.996 & 0.908 & \textbf{0.997} & 0.831 & 0.732 \\
              & (0.993, 1) & (1, 0.901, 0.823) & (0.999, 1, 0.991, 1) & (0.746, 0.774, NA, 0.945, 0.861) & (0.708, NA, 0.584, NA, 0.872, 0.767) \\
        t = 200 & 1 & 0.827 & \textbf{0.99} & 0.722 & 0.625 \\
              & (1, 1) & (0.796, 0.91, 0.775) & (0.992, 0.99, 0.991, 0.989) & (0.799, 0.751, 0.887, 0.564, 0.609) & (0.634, NA, 0.62, 0.762, 0.527, 0.585) \\
        t = 100, Toeplitz & 0.997 & 0.926 & \textbf{0.986} & 0.826 & 0.72 \\
              & (0.994, 1) & (0.952, 0.98, 0.848) & (0.987, 0.991, 0.974, 0.992) & (0.751, 0.79, 0.869, 0.895, NA) & (NA, 0.636, 0.63, 0.819, 0.795, NA) \\
        t = 200, Toeplitz & 1 & 0.835 & \textbf{0.993} & 0.828 & 0.659 \\
              & (1, 1) & (0.86, 0.859, 0.786) & (0.991, 0.996, 0.988, 1) & (0.868, 0.713, 0.876, NA, 0.856) & (0.74, 0.481, 0.572, 0.802, NA, 0.703) \\
        t = 100, $\uparrow$ variance & \textbf{0.992} & 0.835 & 0.653 & 0.754 & 0.629 \\
              & (0.985, 1) & (0.884, 0.867, 0.756) & (0.907, 0.648, 0.473, 0.586) & (0.704, 0.669, NA, 0.832, 0.812) & (0.606, 0.473, 0.606, NA, 0.757, 0.704) \\
        t = 200, $\uparrow$ variance & 1 & 0.789 & \textbf{0.931} & 0.787 & 0.647 \\
              & (1, 1) & (0.727, 0.924, 0.717) & (0.904, 0.965, 0.891, 0.965) & (0.774, NA, 0.732, 0.795, 0.847) & (0.667, NA, NA, 0.592, 0.662, 0.667) \\
        \bottomrule
    \end{tabular}
    \end{adjustbox}
    \label{Kmeans_pbiggern_table1}
\end{table}

\begin{table}[ht]
    \centering
    \caption{WCRI and TWCRI values with $p=1,000$. K-means clustering was used.}
\begin{adjustbox}{width=0.95\textwidth}
    \begin{tabular}{lcccccc}
        \toprule
        & K = 2 & K = 3 & K = 4 & K = 5 & K = 6 \\
        \midrule
        t = 100 & \textbf{0.498}, 0.996 & 0.3, 0.901 & 0.249, \textbf{0.998} & 0.203, 0.814 & 0.178, 0.712 \\
              & (0.606, 0.39) & (0.31, 0.27, 0.321) & (0.309, 0.3, 0.188, 0.2) & (0.231, 0.232, NA, 0.179, 0.172) & (0.219, NA, 0.175, NA, 0.165, 0.153) \\
        t = 200 & \textbf{0.5}, 1 & 0.272, 0.818 & 0.247, \textbf{0.99} & 0.15, 0.754 & 0.128, 0.643 \\
              & (0.535, 0.465) & (0.207, 0.254, 0.356) & (0.258, 0.272, 0.223, 0.237) & (0.207, 0.206, 0.199, 0.045, 0.097) & (0.165, NA, 0.17, 0.171, 0.026, 0.111) \\
        t = 100, Toeplitz & \textbf{0.498}, 0.996 & 0.306, 0.92 & 0.246, \textbf{0.987} & 0.203, 0.813 & 0.175, 0.7 \\
              & (0.606, 0.39) & (0.295, 0.294, 0.3) & (0.306, 0.297, 0.185, 0.198) & (0.232, 0.237, 0.165, 0.179, NA) & (NA, 0.197, 0.189, 0.155, 0.159, NA) \\
        t = 200, Toeplitz & \textbf{0.5}, 1 & 0.277, 0.831 & 0.248, \textbf{0.993} & 0.205, 0.823 & 0.138, 0.694 \\
              & (0.535, 0.465) & (0.223, 0.305, 0.302) & (0.257, 0.273, 0.222, 0.24) & (0.225, 0.196, 0.197, NA, 0.205) & (0.192, 0.009, 0.145, 0.18, NA, 0.168) \\
        t = 100, $\uparrow$ variance & \textbf{0.495}, \textbf{0.991} & 0.277, 0.831 & 0.173, 0.694 & 0.184, 0.738 & 0.129, 0.649 \\
              & (0.601, 0.39) & (0.274, 0.277, 0.279) & (0.281, 0.181, 0.037, 0.193) & (0.218, 0.2, NA, 0.158, 0.162) & (0.181, 0.004, 0.181, NA, 0.143, 0.14) \\
        t = 200, $\uparrow$ variance & \textbf{0.5}, 1 & 0.259, 0.777 & 0.233, \textbf{0.932} & 0.195, 0.783 & 0.161, 0.644 \\
              & (0.535, 0.465) & (0.189, 0.254, 0.333) & (0.235, 0.265, 0.2, 0.231) & (0.201, NA, 0.201, 0.178, 0.203) & (0.173, NA, NA, 0.162, 0.149, 0.16) \\
        \bottomrule
    \end{tabular}
    \end{adjustbox}
    \label{Kmeans_pbiggern_table12}
\end{table}

\begin{table}[ht]
    \centering
    \caption{Values of the ERICA statistic with $p=1,000$. Hierarchical clustering with Ward linkage (HC-WL) was used.}
\begin{adjustbox}{width=0.95\textwidth}
    \begin{tabular}{lcccccc}
        \toprule
        & K = 2 & K = 3 & K = 4 & K = 5 & K = 6 \\
        \midrule
        t = 100 & 0.968 & 0.931 & \textbf{0.995} & 0.801 & 0.735 \\
              & (0.969, 0.967) & (0.96, 0.978, 0.857) & (0.993, 0.996, 1, 0.991) & (0.875, 0.613, 0.714, 0.952, 0.855) & (0.778, 0.596, 0.676, 0.897, 0.778, 0.688) \\
        t = 200 & 0.958 & 0.856 & \textbf{1} & 0.783 & 0.706 \\
              & (0.977, 0.94) & (0.871, 0.938, 0.76) & (1, 1, 1, 1) & (0.92, 0.867, 0.566, 0.686, 0.876) & (0.832, 0.738, NA, 0.527, 0.662, 0.771) \\
        t = 100, Toeplitz & 0.972 & 0.936 & \textbf{0.996} & 0.822 & 0.754 \\
              & (0.971, 0.973) & (0.97, 0.97, 0.87) & (0.993, 0.996, 1, 0.995) & (0.814, 0.606, 0.8, 0.958, 0.932) & (0.748, 0.653, 0.647, 0.728, 0.907, 0.846) \\
        t = 200, Toeplitz & 0.948 & 0.846 & \textbf{1} & 0.792 & 0.702 \\
              & (0.965, 0.931) & (0.868, 0.932, 0.74) & (1, 1, 1, 1) & (0.855, 0.572, 0.697, 0.952, 0.885) & (0.721, 0.608, 0.568, 0.676, 0.871, 0.768) \\
        t = 100, $\uparrow$ variance & 0.967 & 0.92 & \textbf{0.995} & 0.802 & 0.735 \\
              & (0.974, 0.96) & (0.945, 0.976, 0.839) & (0.993, 0.996, 1, 0.991) & (0.875, 0.713, 0.615, 0.952, 0.855) & (0.778, 0.668, 0.602, 0.897, 0.778, 0.688) \\
        t = 200, $\uparrow$ variance & 0.968 & 0.844 & \textbf{1} & 0.781 & 0.705 \\
              & (0.985, 0.951) & (0.862, 0.911, 0.761) & (1, 1, 1, 1) & (0.92, 0.867, 0.557, 0.689, 0.876) & (0.832, 0.738, NA, 0.521, 0.665, 0.771) \\
        \bottomrule
    \end{tabular}
    \end{adjustbox}
    \label{ACWard_pbiggern_results1}
\end{table}

\begin{table}[ht]
    \centering
    \caption{WCRI and TWCRI values with $p=1,000$. Hierarchical clustering with Ward linkage (HC-WL) was used.}
\begin{adjustbox}{width=0.95\textwidth}
    \begin{tabular}{lcccccc}
        \toprule
        & K = 2 & K = 3 & K = 4 & K = 5 & K = 6 \\
        \midrule
        t = 100 & \textbf{0.484}, 0.968 & 0.308, 0.925 & 0.248, \textbf{0.995} & 0.164, 0.823 & 0.124, 0.748 \\
              & (0.591, 0.377) & (0.297, 0.293, 0.334) & (0.308, 0.298, 0.19, 0.198) & (0.271, 0.085, 0.114, 0.18, 0.171) & (0.241, 0.077, 0.114, 0.17, 0.054, 0.089) \\
        t = 200 & \textbf{0.48}, 0.96 & 0.279, 0.838 & 0.25, \textbf{1} & 0.167, 0.839 & 0.149, 0.747 \\
              & (0.523, 0.437) & (0.226, 0.258, 0.353) & (0.26, 0.275, 0.225, 0.24) & (0.239, 0.238, 0.014, 0.137, 0.21) & (0.216, 0.203, NA, 0.018, 0.125, 0.185) \\
        t = 100, Toeplitz & \textbf{0.485}, 0.971 & 0.31, 0.931 & 0.248, \textbf{0.995} & 0.168, 0.843 & 0.13, 0.78 \\
              & (0.592, 0.379) & (0.3, 0.291, 0.339) & (0.307, 0.299, 0.19, 0.199) & (0.252, 0.054, 0.168, 0.182, 0.186) & (0.194, 0.032, 0.058, 0.153, 0.172, 0.169) \\
        t = 200, Toeplitz & \textbf{0.474}, 0.949 & 0.275, 0.827 & 0.25, \textbf{1} & 0.166, 0.834 & 0.124, 0.745 \\
              & (0.516, 0.433) & (0.225, 0.261, 0.34) & (0.26, 0.275, 0.225, 0.24) & (0.222, 0.028, 0.156, 0.214, 0.212) & (0.169, 0.015, 0.031, 0.148, 0.196, 0.184) \\
        t = 100, $\uparrow$ variance & \textbf{0.484}, 0.969 & 0.304, 0.913 & 0.248, \textbf{0.995} & 0.164, 0.823 & 0.124, 0.747 \\
              & (0.594, 0.374) & (0.293, 0.293, 0.327) & (0.308, 0.298, 0.19, 0.198) & (0.271, 0.114, 0.086, 0.18, 0.171) & (0.241, 0.113, 0.078, 0.17, 0.054, 0.089) \\
        t = 200, $\uparrow$ variance & \textbf{0.484}, 0.969 & 0.276, 0.828 & 0.25, \textbf{1} & 0.168, 0.84 & 0.149, 0.748 \\
              & (0.527, 0.442) & (0.224, 0.25, 0.354) & (0.26, 0.275, 0.225, 0.24) & (0.239, 0.238, 0.013, 0.137, 0.21) & (0.216, 0.203, NA, 0.018, 0.126, 0.185) \\
        \bottomrule
    \end{tabular}
    \end{adjustbox}
    \label{ACWard_pbiggern_results2}
\end{table}

\begin{table}[ht]
    \centering
    \caption{Values of the ERICA statistic with $p=1,000$. Hierarchical clustering with single linkage (HC-SL) was used.}
\begin{adjustbox}{width=0.95\textwidth}
    \begin{tabular}{lcccccc}
        \toprule
        & K = 2 & K = 3 & K = 4 & K = 5 & K = 6 \\
        \midrule
        t = 100 & 0.903 & 0.909 & \textbf{0.998} & 0.829 & 0.755 \\
              & (0.903, 0.903) & (0.851, 0.924, 0.952) & (1, 0.999, 1, 0.994) & (0.953, 0.913, 0.951, 0.817, 0.513) & (0.91, 0.85, 0.819, 0.677, 0.465, 0.81) \\
        t = 200 & 0.887 & 0.9 & \textbf{1} & 0.836 & 0.767 \\
              & (0.94, 0.834) & (0.849, 0.928, 0.923) & (1, 1, 1, 1) & (0.907, 0.937, 0.894, 0.523, 0.92) & (0.892, 0.915, 0.785, 0.523, 0.897, 0.59) \\
        t = 100, Toeplitz & 0.878 & 0.764 & \textbf{0.998} & 0.835 & 0.764 \\
              & (0.89, 0.867) & (0.749, 0.694, 0.849) & (0.997, 1, 1, 0.997) & (0.917, 0.907, 0.879, 0.55, 0.925) & (0.823, 0.864, 0.837, 0.484, 0.725, 0.851) \\
        t = 200, Toeplitz & 0.878 & 0.862 & \textbf{1} & 0.899 & 0.789 \\
              & (0.814, 0.943) & (0.915, 0.736, 0.937) & (1, 1, 1, 1) & (0.878, 0.914, 0.882, NA, 0.922) & (0.914, 0.914, 0.723, 0.832, 0.47, 0.886) \\
        t = 100, $\uparrow$ variance & 0.897 & 0.892 & \textbf{0.998} & 0.817 & 0.757 \\
              & (0.882, 0.912) & (0.822, 0.916, 0.938) & (1, 0.999, 1, 0.994) & (0.947, 0.907, 0.917, 0.805, 0.513) & (0.913, 0.843, 0.794, 0.676, 0.511, 0.81) \\
        t = 200, $\uparrow$ variance & 0.849 & 0.86 & \textbf{1} & 0.828 & 0.752 \\
              & (0.912, 0.786) & (0.785, 0.882, 0.914) & (1, 1, 1, 1) & (0.904, 0.921, 0.897, 0.523, 0.898) & (0.862, 0.866, 0.766, 0.571, 0.879, 0.568) \\
        \bottomrule
    \end{tabular}
    \end{adjustbox}
    \label{ACsingle_pbiggern_results1}
\end{table}

\begin{table}[ht]
    \centering
    \caption{WCRI and TWCRI values with $p=1,000$. Hierarchical clustering with single linkage (HC-SL) was used.}
\begin{adjustbox}{width=0.95\textwidth}
    \begin{tabular}{lcccccc}
        \toprule
        & K = 2 & K = 3 & K = 4 & K = 5 & K = 6 \\
        \midrule
        t = 100 & \textbf{0.451}, 0.903 & 0.295, 0.885 & 0.249, \textbf{0.998} & 0.182, 0.91 & 0.137, 0.827 \\
              & (0.686, 0.216) & (0.519, 0.175, 0.19) & (0.31, 0.299, 0.19, 0.198) & (0.295, 0.274, 0.18, 0.155, 0.005) & (0.282, 0.255, 0.155, 0.121, 0.004, 0.008) \\
        t = 200 & \textbf{0.445}, 0.891 & 0.294, 0.884 & 0.25, \textbf{1} & 0.182, 0.913 & 0.145, 0.872 \\
              & (0.512, 0.379) & (0.454, 0.208, 0.221) & (0.26, 0.275, 0.225, 0.24) & (0.235, 0.257, 0.196, 0.002, 0.22) & (0.231, 0.251, 0.172, 0.002, 0.21, 0.002) \\
        t = 100, Toeplitz & 0.441, 0.883 & 0.249, 0.749 & \textbf{0.249}, \textbf{0.998} & 0.181, 0.905 & 0.139, 0.835 \\
              & (0.614, 0.269) & (0.247, 0.298, 0.203) & (0.309, 0.3, 0.19, 0.199) & (0.284, 0.272, 0.167, 0.005, 0.175) & (0.255, 0.25, 0.15, 0.009, 0.007, 0.161) \\
        t = 200, Toeplitz & \textbf{0.425}, 0.85 & 0.279, 0.839 & 0.25, \textbf{1} & 0.224, 0.898 & 0.147, 0.886 \\
              & (0.586, 0.264) & (0.279, 0.334, 0.225) & (0.26, 0.275, 0.225, 0.24) & (0.228, 0.251, 0.198, NA, 0.221) & (0.237, 0.246, 0.003, 0.183, 0.002, 0.212) \\
        t = 100, $\uparrow$ variance & \textbf{0.444}, 0.888 & 0.287, 0.863 & 0.249, \textbf{0.998} & 0.179, 0.898 & 0.137, 0.822 \\
              & (0.688, 0.2) & (0.501, 0.174, 0.187) & (0.31, 0.299, 0.19, 0.198) & (0.293, 0.272, 0.174, 0.153, 0.005) & (0.283, 0.253, 0.151, 0.121, 0.005, 0.008) \\
        t = 200, $\uparrow$ variance & \textbf{0.429}, 0.858 & 0.279, 0.838 & 0.25, \textbf{1} & 0.18, 0.904 & 0.14, 0.843 \\
              & (0.52, 0.338) & (0.416, 0.202, 0.219) & (0.26, 0.275, 0.225, 0.24) & (0.235, 0.253, 0.197, 0.002, 0.215) & (0.224, 0.238, 0.168, 0.002, 0.206, 0.002) \\
        \bottomrule
    \end{tabular}
    \end{adjustbox}
    \label{ACsingle_pbiggern_results2}
\end{table}

\FloatBarrier
\section{Assessing Clustering via Consensus Across Metrics}

This appendix summarizes the number of clusters selected by the different metrics for the Gaussian mixture (Figure \ref{Fintalsynth1_res1}) and breast cancer (Figure \ref{Fintalparmig1_res1}) datasets across the three clustering techniques. 
The unpopulated entries are treated as zeros. 
The bold entries denote the value of $K$ selected by a metric after equally weighting the results of the three clustering techniques. 
Consensus was formed by equally weighting the results of k-means, HC-WL, and HC-SL.

\renewcommand{\thetable}{\thesection.\arabic{table}}
\renewcommand{\thefigure}{\thesection.\arabic{figure}}

\setcounter{table}{0}
\setcounter{figure}{0}

\begin{table}[ht]
    \centering
    \caption{In the tuple $(a, b, c)$, $a$ ($b$ and $c$, respectively) denotes the number of times that $K=K^{*}$ according to the ERICA statistic (WCRI and TWCRI, respectively) values across the three clustering techniques.} 
\begin{adjustbox}{width=0.65\textwidth}
    \begin{tabular}{lccccccccc}
        \toprule
        Synthetic dataset & K = 2 & K = 3 & K = 4 & K = 5 & K = 6 \\
        \midrule
        t = 10,000, p = 100 & 0, \textbf{3}, 0 &  & \textbf{3}, 0, \textbf{3} &  & \\
        t = 10,000, p = 200 & 0, \textbf{3}, 0 &  & \textbf{3}, 0, \textbf{3} &  & \\
        t = 10,000, p = 400 & 0, \textbf{2}, 0 &  & \textbf{3}, 1, \textbf{3} &  & \\
        t = 10,000, p = 100, Toeplitz & 0, \textbf{3}, 0 &  & \textbf{2}, 0, \textbf{2} & 1, 0, 1 & \\
        t = 10,000, p = 200, Toeplitz & 0, \textbf{3}, 0 &  & \textbf{2}, 0, \textbf{2} &  & 1, 0, 1 \\
        t = 10,000, p = 400, Toeplitz & 0, \textbf{3}, 0 &  & \textbf{3}, 0, \textbf{3} &  & \\
        t = 10,000, p = 100, $\uparrow$ variance & 0, \textbf{2}, 0 &  & \textbf{2}, 1, \textbf{2} & 1, 0, 0 & 0, 0, 1 \\
        t = 10,000, p = 200, $\uparrow$ variance & 0, \textbf{3}, 1 &  & \textbf{2}, 0, \textbf{2} & 1, 0, 0 & \\
        t = 10,000, p = 400, $\uparrow$ variance & 0, \textbf{3}, 0 &  & \textbf{3}, 0, \textbf{3} &  &  \\
        t = 100, p = 1,000 & 0, \textbf{3}, 0 &  & \textbf{3}, 0, \textbf{3} &  &  \\
        t = 200, p = 1,000 & 0, \textbf{3}, 0 &  & \textbf{3}, 0, \textbf{3} &  & \\
        t = 100, p = 1,000, Toeplitz & 0, \textbf{2}, 0 &  & \textbf{3}, 1, \textbf{3} &  &  \\
        t = 200, p = 1,000, Toeplitz & 0, \textbf{3}, 0 &  & \textbf{3}, 0, \textbf{3} &  &  \\
        t = 100, p = 1,000, $\uparrow$ variance & 1, \textbf{3}, 1 &  & \textbf{2}, 0, \textbf{2} &  &  \\
        t = 200, p = 1,000, $\uparrow$ variance & 0, \textbf{3}, 0 &  & \textbf{3}, 0, \textbf{3} &  & \\
        \bottomrule
    \end{tabular}
    \end{adjustbox}
    \label{Fintalsynth1_res1}
\end{table}

\begin{table}[ht]
    \centering
    \caption{In the tuple $(a, b)$, $a$ and $b$, respectively, denote the number of times that $K=K^{*}$ according to the ERICA statistic and TWCRI across the three clustering techniques. 
    }
\begin{adjustbox}{width=0.70\textwidth}
    \begin{tabular}{lccccccccc}
        \toprule
        Dataset & K = 2 & K = 3 & K = 4 & K = 5 & K = 6 & K = 7 & K = 8 \\
        \midrule
        Mainz (full) & \textbf{2}, \textbf{2} &  &  &  & 1, 1 &  &  \\
        Mainz (3G) & 0, \textbf{2} & 1, 0 &  &  & \textbf{2}, 1 &  &  \\
        Transbig (full) & 1, \textbf{2} &  & 0, 1 &  &  & \textbf{2}, 0 &  \\
        Transbig (3G) &   &  &  & 0, \textbf{1} &  & \textbf{2}, \textbf{1} & 1, \textbf{1} \\
        VDX (full) & \textbf{1}, \textbf{2} &  &  & \textbf{1}, 0 & \textbf{1}, 0 & 0, 1 &  \\
        VDX (3G) &  &  &  &  & \textbf{2}, \textbf{2} & 1, 0 & 0, 1 \\
        \bottomrule
    \end{tabular}
    \end{adjustbox}
    \label{Fintalparmig1_res1}
\end{table}

\FloatBarrier
\section{Per-cluster Replicability Results for Breast Cancer Data}

The following tables report the metrics obtained by applying ERICA to different breast cancer datasets. 
In each case we used $B=200$ MCSS iterations, $P=80\%$, and either the full $p=22,283$ genes (full) or the 3G subset of genes (3G). 
The format of the panels is described in Table \ref{template_for_mytable1}. The bold entries represent the metric (ERICA statistic, WCRI, or TWCRI) associated with the most appropriate number of clusters ($K^{*}$) in the dataset. 

\renewcommand{\thetable}{\thesection.\arabic{table}}
\renewcommand{\thefigure}{\thesection.\arabic{figure}}

\setcounter{table}{0}
\setcounter{figure}{0}

\begin{table}[ht]
    \centering
    \caption{The ERICA statistics attained for the Mainz dataset with $t=200$ samples. 
    }
\begin{adjustbox}{width=1.0\textwidth}
    \begin{tabular}{lccccccccc}
        \toprule
        Clustering & K = 2 & K = 3 & K = 4 & K = 5 & K = 6 & K = 7 & K = 8 \\
        \midrule
        K-means & \textbf{0.752} & 0.706 & 0.575 & 0.491 & 0.443 & 0.383 & 0.37 \\
           (full)  & (0.816, 0.688) & (0.694, 0.711, 0.713) & (0.375, 0.567, 0.693, 0.667) & (0.528, 0.329, 0.434, 0.608, & (0.295, 0.536, 0.379, 0.403, & (0.292, 0.473, 0.293, 0.345, & (0.362, NA, 0.358, 0.501, \\
               &  &  &  & 0.56) & 0.562, 0.484) & 0.469, 0.435, 0.379) & 0.298, 0.329, 0.381, 0.361) \\
        HC-WL & 0.764 & 0.717 & 0.63 & 0.634 & \textbf{0.643} & 0.605 & 0.585 \\
             (full) & (0.749, 0.779) & (0.679, 0.752, 0.722) & (0.607, 0.555, 0.763, 0.596) & (0.574, 0.6, 0.809, 0.558, & (0.617, 0.593, 0.575, 0.838, & (0.594, 0.572, 0.532, 0.787, & (0.597, 0.555, 0.48, 0.549, \\
               &  &  &  & 0.631) & 0.59, 0.646) & 0.538, 0.576, 0.639) & 0.792, 0.491, 0.609, 0.613) \\
        HC-SL & \textbf{0.912} & 0.77 & 0.694 & 0.657 & 0.639 & 0.622 & 0.57 \\
             (full) & (0.85, 0.974) & (0.709, 0.972, 0.63) & (0.581, 0.961, 0.631, 0.605) & (0.547, 0.621, 0.936, 0.623, & (0.735, 0.545, 0.918, 0.56, & (0.795, 0.5, 0.902, 0.536, & (0.729, 0.444, 0.897, 0.425, \\
               &  &  &  & 0.562) & 0.625, 0.451) & 0.505, 0.536, 0.583) & 0.492, 0.589, 0.486, 0.501) \\
        K-means  & 0.989 & 0.757 & 0.882 & 0.688 & \textbf{0.819} & 0.772 & 0.701 \\
             (3G) & (0.979, 1) & (0.914, 0.705, 0.653) & (0.991, 0.885, 0.817, 0.838) & (0.947, 0.623, 0.613, 0.66, & (0.932, 0.795, 0.79, 0.808, & (0.847, 0.775, 0.698, 0.882, & (0.768, 0.652, 0.385, 0.586, \\
               &  &  &  & 0.599) & 0.784, 0.81) & 0.597, 0.733, 0.875) & 0.96, 0.595, 0.775, 0.889) \\
        HC-WL & 0.934 & 0.859 & 0.764 & 0.746 & \textbf{0.769} & 0.738 & 0.731 \\
             (3G) & (0.907, 0.961) & (0.942, 0.854, 0.781) & (0.929, 0.709, 0.842, 0.577) & (0.942, 0.651, 0.71, 0.755, & (0.929, 0.654, 0.921, 0.65, & (0.868, 0.641, 0.55, 0.96, & (0.81, 0.717, 0.52, 0.958, \\
               &  &  &  & 0.676) & 0.649, 0.812) & 0.634, 0.675, 0.842) & 0.654, 0.589, 0.725, 0.832) \\
        HC-SL & 0.762 & \textbf{0.817} & 0.813 & 0.789 & 0.688 & 0.651 & 0.595 \\
             (3G) & (0.594, 0.931) & (0.617, 0.948, 0.888) & (1, 0.768, 0.891, 0.593) & (0.9, 0.76, 0.889, 0.831, & (0.76, 0.72, 0.85, 0.779, & (0.691, 0.506, 0.665, 0.827, & (0.583, 0.422, 0.641, 0.799, \\
               &  &  &  & 0.569) & 0.555, 0.467) & 0.853, 0.528, 0.492) & 0.824, 0.428, 0.612, 0.452) \\
        \bottomrule
    \end{tabular}
    \end{adjustbox}
    \label{mainzfull_res1}
\end{table}

\begin{table}[ht]
    \centering
    \caption{WCRI and TWCRI values attained for the Mainz dataset with $t=200$ samples. 
    }
\begin{adjustbox}{width=0.95\textwidth}
    \begin{tabular}{lccccccccc}
        \toprule
        Clustering & K = 2 & K = 3 & K = 4 & K = 5 & K = 6 & K = 7 & K = 8 \\
        \midrule
        K-means & \textbf{0.376}, \textbf{0.753} & 0.234, 0.704 & 0.155, 0.623 & 0.105, 0.525 & 0.078, 0.472 & 0.056, 0.396 & 0.053, 0.372 \\
           (full)  & (0.42, 0.333) & (0.267, 0.127, 0.31) & (0.013, 0.215, 0.135, 0.26) & (0.11, 0.003, 0.108, 0.103, & (0.002, 0.077, 0.054, 0.062, & (0.004, 0.07, 0.019, 0.067, & (0.056, NA, 0.089, 0.067, \\
               &  &  &  & 0.201) & 0.098, 0.179) & 0.079, 0.034, 0.123) & 0.01, 0.062, 0.083, 0.005) \\
        HC-WL & \textbf{0.387}, 0.775 & 0.237, 0.711 & 0.153, 0.612 & 0.123, 0.618 & 0.105, \textbf{0.632} & 0.085, 0.595 & 0.072, 0.58 \\
           (full)  & (0.093, 0.681) & (0.254, 0.127, 0.328) & (0.221, 0.074, 0.08, 0.235) & (0.189, 0.09, 0.076, 0.047, & (0.074, 0.136, 0.083, 0.075, & (0.071, 0.103, 0.079, 0.074, & (0.071, 0.088, 0.014, 0.079, \\
               &  &  &  & 0.214) & 0.056, 0.206) & 0.096, 0.054, 0.115) & 0.071, 0.095, 0.048, 0.11) \\
        HC-SL & \textbf{0.486}, \textbf{0.972} & 0.32, 0.961 & 0.235, 0.942 & 0.182, 0.911 & 0.148, 0.888 & 0.123, 0.864 & 0.105, 0.845 \\
           (full)  & (0.008, 0.964) & (0.01, 0.938, 0.012) & (0.008, 0.908, 0.012, 0.012) & (0.01, 0.009, 0.871, 0.009, & (0.007, 0.008, 0.844, 0.011, & (0.007, 0.012, 0.803, 0.013, & (0.007, 0.013, 0.78, 0.008, \\
               &  &  &  & 0.011) & 0.003, 0.013) & 0.01, 0.008, 0.008) & 0.012, 0.005, 0.009, 0.007) \\
        K-means & 0.498, 0.996 & 0.245, 0.736 & 0.218, 0.872 & 0.134, 0.671 & \textbf{0.136}, \textbf{0.816} & 0.105, 0.737 & 0.083, 0.671\\
           (3G) & (0.176, 0.82) & (0.15, 0.539, 0.045) & (0.123, 0.402, 0.253, 0.092) & (0.118, 0.305, 0.03, 0.178, & (0.111, 0.135, 0.031, 0.335, & (0.101, 0.127, 0.205, 0.035, & (0.092, 0.075, 0.009, 0.105, \\
               &  &  &  & 0.038) & 0.137, 0.064) & 0.104, 0.091, 0.07) & 0.038, 0.202, 0.077, 0.071) \\
        HC-WL & \textbf{0.475}, \textbf{0.951} & 0.285, 0.857 & 0.183, 0.734 & 0.145, 0.729 & 0.117, 0.707 & 0.098, 0.687 & 0.085, 0.683 \\
           (3G)  & (0.167, 0.783) & (0.117, 0.653, 0.085) & (0.116, 0.453, 0.092, 0.072) & (0.113, 0.107, 0.355, 0.083, & (0.111, 0.117, 0.036, 0.302, & (0.104, 0.099, 0.063, 0.038, & (0.093, 0.082, 0.072, 0.038, \\
               &  &  &  & 0.071) & 0.077, 0.06) & 0.24, 0.077, 0.063) & 0.203, 0.076, 0.054, 0.062) \\
        HC-SL & \textbf{0.46}, \textbf{0.921} & 0.303, 0.911 & 0.215, 0.861 & 0.171, 0.857 & 0.135, 0.81 & 0.115, 0.805 & 0.096, 0.774 \\
           (3G)  & (0.017, 0.903) & (0.064, 0.811, 0.035) & (0.005, 0.088, 0.735, 0.032) & (0.004, 0.087, 0.707, 0.033, & (0.003, 0.082, 0.654, 0.031, & (0.006, 0.005, 0.073, 0.033, & (0.008, 0.006, 0.067, 0.031, \\
               &  &  &  & 0.025) & 0.03, 0.007) & 0.653, 0.029, 0.004) & 0.618, 0.004, 0.03, 0.006) \\
        \bottomrule
    \end{tabular}
    \end{adjustbox}
    \label{mainzfull_res2}
\end{table}

\begin{table}[ht]
    \centering
    \caption{The ERICA statistics attained for the VDX dataset with $t=344$ samples. 
    }
\begin{adjustbox}{width=1.0\textwidth}
    \begin{tabular}{lccccccccc}
        \toprule
        Clustering & K = 2 & K = 3 & K = 4 & K = 5 & K = 6 & K = 7 & K = 8 \\
        \midrule
        K-means & 0.932 & 0.522 & 0.445 & \textbf{0.539} & 0.425 & 0.417 & 0.522 \\
             (full) & (0.94, 0.925) & (0.531, 0.599, 0.438) & (0.493, 0.451, 0.454, 0.385) & (0.577, 0.562, 0.519, 0.656, & (0.331, 0.475, 0.421, 0.409, & (0.218, 0.476, 0.472, 0.445, & (0.537, 0.639, 0.469, 0.404, \\
               &  &  &  & 0.385) & 0.586, 0.333) & 0.429, 0.323, 0.561) & 0.52, 0.444, 0.643, NA) \\
        HC-WL & \textbf{0.871} & 0.746 & 0.685 & 0.657 & 0.646 & 0.606 & 0.572 \\
             (full) & (0.874, 0.868) & (0.769, 0.797, 0.672) & (0.694, 0.775, 0.624, 0.649) & (0.636, 0.779, 0.603, 0.728, & (0.592, 0.636, 0.727, 0.632, & (0.587, 0.578, 0.671, 0.584, & (0.567, 0.549, 0.519, 0.621, \\
               &  &  &  & 0.539) & 0.762, 0.528) & 0.596, 0.733, 0.496) & 0.559, 0.593, 0.665, 0.504) \\
        HC-SL & 0.85 & 0.8 & 0.668 & 0.678 & \textbf{0.687} & 0.684 & 0.565 \\
             (full) & (0.991, 0.71) & (0.985, 0.723, 0.692) & (0.517, 0.978, 0.554, 0.624) & (0.508, 0.977, 0.608, 0.682, & (0.521, 0.969, 0.494, 0.73, & (0.442, 0.617, 0.95, 0.531, & (0.526, 0.521, 0.429, 0.953, \\
               &  &  &  & 0.615) & 0.822, 0.587) & 0.8, 0.878, 0.576) & 0.48, 0.559, 0.545, 0.51) \\
        K-means  & 0.992 & 0.975 & 0.866 & 0.77 & \textbf{0.828} & 0.788 & 0.713 \\
             (3G) & (0.988, 0.997) & (0.991, 0.98, 0.954) & (0.979, 0.736, 0.905, 0.847) & (0.939, 0.677, 0.921, 0.721, & (0.864, 0.723, 0.93, 0.813, & (0.693, 0.615, NA, 0.799, & (0.783, 0.545, 0.659, 0.534, \\
               &  &  &  & 0.594) & 0.889, 0.754) & 0.893, 0.861, 0.869) & 0.872, 0.744, 0.782, 0.79) \\
        HC-WL & 0.941 & 0.937 & 0.783 & 0.771 & \textbf{0.783} & 0.734 & 0.692 \\
             (3G) & (0.929, 0.953) & (0.971, 0.958, 0.883) & (0.893, 0.762, 0.874, 0.605) & (0.913, 0.648, 0.924, 0.616, & (0.852, 0.662, 0.909, 0.707, & (0.851, 0.565, 0.642, 0.893, & (0.792, 0.588, 0.574, 0.834, \\
               &  &  &  & 0.756) & 0.768, 0.801) & 0.691, 0.724, 0.771) & 0.645, 0.644, 0.675, 0.685) \\
        HC-SL & 0.794 & 0.63 & 0.604 & 0.688 & 0.642 & \textbf{0.669} & 0.58 \\
             (3G) & (0.833, 0.756) & (0.775, 0.475, 0.64) & (0.501, 0.822, 0.586, 0.509) & (0.561, 0.86, 0.606, 0.442, & (0.643, 0.865, 0.687, 0.659, & (0.633, 0.831, 0.679, 0.727, & (0.748, 0.838, 0.745, 0.399, \\
               &  &  &  & 0.972) & 0.425, 0.577) & 0.45, 0.418, 0.945) & 0.412, 0.325, 0.523, 0.648) \\
        \bottomrule
    \end{tabular}
    \end{adjustbox}
    \label{VDXfull_res1}
\end{table}

\begin{table}[ht]
    \centering
    \caption{WCRI and TWCRI values attained for the VDX dataset with $t=344$ samples. 
    }
\begin{adjustbox}{width=0.95\textwidth}
    \begin{tabular}{lccccccccc}
        \toprule
        Clustering & K = 2 & K = 3 & K = 4 & K = 5 & K = 6 & K = 7 & K = 8 \\
        \midrule
        K-means & \textbf{0.466}, 0.933 & 0.181, 0.545 & 0.114, 0.456 & 0.112, 0.56 & 0.075, 0.45 & 0.065, \textbf{0.461} & 0.075, 0.53 \\
           (full)  & (0.595, 0.338) & (0.321, 0.193, 0.031) & (0.101, 0.241, 0.1, 0.014) & (0.182, 0.091, 0.176, 0.104, & (0.013, 0.136, 0.077, 0.128, & (0.001, 0.11, 0.075, 0.063, & (0.113, 0.091, 0.059, 0.018, \\
               &  &  &  & 0.007) & 0.09, 0.006) & 0.128, 0.003, 0.081) & 0.093, 0.065, 0.091, NA) \\
        HC-WL & \textbf{0.435}, \textbf{0.871} & 0.246, 0.74 & 0.169, 0.677 & 0.129, 0.646 & 0.105, 0.633 & 0.087, 0.61 & 0.072, 0.576 \\
           (full)  & (0.614, 0.257) & (0.263, 0.227, 0.25) & (0.223, 0.155, 0.237, 0.062) & (0.199, 0.142, 0.145, 0.074, & (0.123, 0.081, 0.128, 0.137, & (0.112, 0.077, 0.113, 0.061, & (0.09, 0.071, 0.06, 0.097, \\
               &  &  &  & 0.086) & 0.079, 0.085) & 0.133, 0.083, 0.031) & 0.055, 0.093, 0.081, 0.029) \\
        HC-SL & \textbf{0.494}, \textbf{0.988} & 0.326, 0.979 & 0.24, 0.962 & 0.191, 0.959 & 0.158, 0.949 & 0.132, 0.926 & 0.114, 0.917 \\
           (full)  & (0.982, 0.006) & (0.967, 0.008, 0.004) & (0.007, 0.946, 0.004, 0.005) & (0.007, 0.937, 0.007, 0.003, & (0.006, 0.923, 0.007, 0.004, & (0.008, 0.003, 0.9, 0.006, & (0.004, 0.004, 0.007, 0.886, \\
               &  &  &  & 0.005) & 0.004, 0.005) & 0.002, 0.002, 0.005) & 0.005, 0.003, 0.001, 0.007) \\
        K-means & \textbf{0.496}, 0.993 & 0.326, 0.979 & 0.217, 0.868 & 0.157, 0.787 & 0.139, \textbf{0.834} & 0.128, 0.768 & 0.092, 0.737 \\
           (3G)  & (0.39, 0.609) & (0.296, 0.541, 0.141) & (0.287, 0.173, 0.128, 0.278) & (0.267, 0.163, 0.107, 0.207, & (0.236, 0.098, 0.097, 0.205, & (0.189, 0.084, NA, 0.202, & (0.166, 0.036, 0.09, 0.032, \\
               &  &  &  & 0.041) & 0.134, 0.061) & 0.093, 0.13, 0.07) & 0.088, 0.151, 0.106, 0.064) \\
        HC-WL & \textbf{0.471}, 0.943 & 0.316, 0.95 & 0.2, 0.801 & 0.15, 0.752 & 0.13, \textbf{0.78} & 0.107, 0.754 & 0.086, 0.692 \\
           (3G)  & (0.367, 0.576) & (0.276, 0.537, 0.136) & (0.262, 0.401, 0.094, 0.044) & (0.257, 0.199, 0.094, 0.139, & (0.237, 0.084, 0.092, 0.176, & (0.22, 0.018, 0.08, 0.088, & (0.198, 0.023, 0.073, 0.082, \\
               &  &  &  & 0.061) & 0.12, 0.067) & 0.174, 0.105, 0.067) & 0.163, 0.093, 0.025, 0.031) \\
        HC-SL & \textbf{0.416}, 0.832 & 0.254, 0.763 & 0.187, 0.75 & 0.15, 0.754 & 0.127, 0.767 & 0.105, 0.74 & 0.096, \textbf{0.774} \\
           (3G)  & (0.83, 0.002) & (0.741, 0.016, 0.005) & (0.07, 0.614, 0.061, 0.004) & (0.138, 0.542, 0.063, 0.006, & (0.177, 0.495, 0.073, 0.001, & (0.178, 0.463, 0.073, 0.002, & (0.208, 0.46, 0.069, 0.015, \\
               &  &  &  & 0.02) & 0.014, 0.003) & 0.001, 0.018, 0.002) & 0.002, 9.4e-4, 0.015, 0.001) \\
        \bottomrule
    \end{tabular}
    \end{adjustbox}
    \label{VDXfull_res2}
\end{table}

\begin{table}[ht]
    \centering
    \caption{The ERICA statistics attained for the Transbig dataset with $t=198$ samples. 
    }
\begin{adjustbox}{width=1.0\textwidth}
    \begin{tabular}{lccccccccc}
        \toprule
        Clustering & K = 2 & K = 3 & K = 4 & K = 5 & K = 6 & K = 7 & K = 8 \\
        \midrule
        K-means & \textbf{0.877} & 0.641 & 0.45 & 0.444 & 0.362 & 0.342 & 0.295 \\
             (full) & (0.877, 0.878) & (0.517, 0.844, 0.563) & (0.39, 0.55, 0.456, 0.404) & (0.433, 0.364, 0.525, 0.411, & (0.353, 0.323, 0.434, 0.38, & (0.322, 0.272, 0.485, 0.332, & (0.362, 0.303, 0.246, 0.205, \\
               &  &  &  & 0.488) & 0.269, 0.418) & 0.358, 0.225, 0.406) & 0.309, 0.317, 0.336, 0.286) \\
        HC-WL & 0.894 & 0.713 & 0.61 & 0.597 & 0.571 & \textbf{0.595} & 0.593 \\
             (full) & (0.874, 0.915) & (0.576, 0.81, 0.754) & (0.605, 0.729, 0.607, 0.499) & (0.616, 0.513, 0.732, 0.59, & (0.572, 0.49, 0.459, 0.751, & (0.515, 0.51, 0.9, 0.443, & (0.475, 0.812, 0.524, 0.783, \\
               &  &  &  & 0.538) & 0.614, 0.542) & 0.696, 0.564, 0.538) & 0.438, 0.644, 0.546, 0.529) \\
        HC-SL & 0.853 & 0.835 & 0.724 & 0.652 & 0.637 & \textbf{0.68} & 0.555 \\
             (full) & (0.984, 0.723) & (0.967, 0.619, 0.92) & (0.486, 0.957, 0.694, 0.762) & (0.51, 0.408, 0.928, 0.781, & (0.443, 0.466, 0.4, 0.921, & (0.663, 0.455, 0.374, 0.902, & (0.668, 0.424, 0.328, 0.891, \\
               &  &  &  & 0.633) & 0.937, 0.658) & 1, 0.98, 0.387) & 0.656, 0.762, 0.338, 0.375) \\
        K-means  & 0.998 & 0.736 & 0.866 & 0.727 & 0.69 & \textbf{0.787} & 0.698 \\
             (3G) & (0.997, 1) & (0.783, 0.839, 0.587) & (0.975, 0.739, 0.906, 0.845) & (0.874, 0.648, 0.878, 0.741, & (0.809, 0.529, 0.591, 0.933, & (0.817, 0.572, 0.677, 0.958, & (0.776, 0.555, 0.645, 0.529, \\
               &  &  &  & 0.498) & 0.658, 0.625) & 0.686, 0.723, 0.776) & 0.914, 0.679, 0.584, 0.908) \\
        HC-WL & 0.964 & 0.919 & 0.797 & 0.816 & 0.768 & \textbf{0.793} & 0.709 \\
             (3G) & (0.95, 0.979) & (0.973, 0.918, 0.867) & (0.948, 0.706, 0.898, 0.638) & (0.923, 0.676, 0.938, 0.763, & (0.873, 0.643, 0.895, 0.727, & (0.763, 0.642, 0.654, 0.916, & (0.723, 0.623, 0.651, 0.904, \\
               &  &  &  & 0.781) & 0.636, 0.835) & 0.671, 0.664, 0.843) & 0.65, 0.639, 0.631, 0.807) \\
        HC-SL & 0.698 & 0.694 & 0.63 & 0.685 & 0.628 & 0.684 & \textbf{0.721} \\
             (3G) & (0.549, 0.844) & (0.813, 0.713, 0.557) & (0.93, 0.708, 0.396, 0.488) & (0.586, 1, 0.856, 0.469, & (0.549, 0.409, 0.953, 0.876, & (0.554, 0.418, 0.976, 0.878, & (0.57, 0.501, 0.813, 0.889, \\
               &  &  &  & 0.518) & 0.483, 0.5) & 0.722, 0.62, 0.623) & 0.688, 0.644, 0.538, 0.628) \\
        \bottomrule
    \end{tabular}
    \end{adjustbox}
    \label{Transbigfull_res1}
\end{table}

\begin{table}[ht]
    \centering
    \caption{WCRI and TWCRI values attained for the Transbig dataset with $t=198$ samples. 
    }
\begin{adjustbox}{width=0.95\textwidth}
    \begin{tabular}{lccccccccc}
        \toprule
        Clustering & K = 2 & K = 3 & K = 4 & K = 5 & K = 6 & K = 7 & K = 8 \\
        \midrule
        K-means & \textbf{0.438}, 0.877 & 0.21, 0.63 & 0.114, \textbf{0.458} & 0.091, 0.458 & 0.064, 0.389 & 0.055, 0.386 & 0.039, 0.316 \\
           (full)  & (0.292, 0.585) & (0.117, 0.238, 0.275) & (0.1, 0.136, 0.204, 0.018) & (0.098, 0.064, 0.116, 0.008, & (0.098, 0.037, 0.107, 0.013, & (0.063, 0.009, 0.105, 0.03, & (0.047, 0.09, 0.008, 0.002, \\
               &  &  &  & 0.172) & 0.001, 0.133) & 0.043, 0.001, 0.135) & 0.048, 0.024, 0.095, 0.002) \\
        HC-WL & \textbf{0.452}, \textbf{0.904} & 0.25, 0.75 & 0.157, 0.628 & 0.121, 0.608 & 0.1, 0.601 & 0.081, 0.567 & 0.068, 0.544 \\
           (full)  & (0.229, 0.675) & (0.064, 0.229, 0.457) & (0.094, 0.187, 0.297, 0.047) & (0.093, 0.049, 0.144, 0.25, & (0.086, 0.054, 0.03, 0.136, & (0.08, 0.051, 0.009, 0.033, & (0.076, 0.004, 0.05, 0.011, \\
               &  &  &  & 0.07) & 0.235, 0.057) & 0.13, 0.199, 0.062) & 0.037, 0.12, 0.184, 0.058) \\
        HC-SL & \textbf{0.489}, \textbf{0.978} & 0.318, 0.956 & 0.235, 0.941 & 0.18, 0.902 & 0.148, 0.888 & 0.124, 0.868 & 0.105, 0.84 \\
           (full)  & (0.964, 0.014) & (0.933, 0.018, 0.004) & (0.009, 0.913, 0.01, 0.007) & (0.012, 0.008, 0.867, 0.003, & (0.011, 0.009, 0.006, 0.846, & (0.006, 0.011, 0.007, 0.824, & (0.006, 0.006, 0.008, 0.787, \\
               &  &  &  & 0.009) & 0.004, 0.009) & 0.005, 0.004, 0.007) & 0.006, 0.007, 0.003, 0.013) \\
        K-means & \textbf{0.499}, 0.999 & 0.27, 0.812 & 0.212, 0.851 & 0.149, 0.749 & 0.115, 0.691 & 0.105, \textbf{0.737} & 0.085, 0.682 \\
           (3G)  & (0.322, 0.676) & (0.233, 0.555, 0.023) & (0.226, 0.209, 0.086, 0.328) & (0.203, 0.193, 0.084, 0.25, & (0.179, 0.008, 0.17, 0.075, & (0.14, 0.034, 0.13, 0.077, & (0.125, 0.039, 0.12, 0.05, \\
               &  &  &  & 0.017) & 0.219, 0.037) & 0.114, 0.189, 0.05) & 0.073, 0.12, 0.097, 0.055) \\
        HC-WL & \textbf{0.485}, 0.97 & 0.308, 0.926 & 0.193, 0.773 & 0.158, \textbf{0.794} & 0.125, 0.753 & 0.101, 0.713 & 0.085, 0.687 \\
           (3G)  & (0.307, 0.663) & (0.226, 0.612, 0.087) & (0.22, 0.435, 0.072, 0.045) & (0.214, 0.167, 0.075, 0.285, & (0.194, 0.123, 0.072, 0.235, & (0.142, 0.029, 0.125, 0.074, & (0.116, 0.044, 0.105, 0.068, \\
               &  &  &  & 0.051) & 0.077, 0.05) & 0.213, 0.077, 0.051) & 0.177, 0.045, 0.076, 0.053) \\
        HC-SL & \textbf{0.419}, 0.838 & 0.234, 0.702 & 0.17, 0.682 & 0.151, 0.757 & 0.125, 0.75 & 0.11, 0.771 & 0.096, \textbf{0.774} \\
           (3G)  & (0.011, 0.827) & (0.004, 0.655, 0.042) & (0.004, 0.622, 0.006, 0.049) & (0.115, 0.005, 0.57, 0.021, & (0.102, 0.01, 0.004, 0.562, & (0.092, 0.021, 0.004, 0.55, & (0.095, 0.027, 0.004, 0.552, \\
               &  &  &  & 0.044) & 0.026, 0.043) & 0.058, 0.012, 0.031) & 0.052, 0.013, 0.027, 0.003) \\
        \bottomrule
    \end{tabular}
    \end{adjustbox}
    \label{Transbigfull_res2}
\end{table}

\clearpage
\vskip 0.2in
\bibliography{sample}

@article{ioannidis2005,
  author = {Ioannidis, John P. A.},
  title = {Why Most Published Research Findings Are False},
  journal = {PLoS Medicine},
  year = {2005}
}

@article{baker2016,
  author = {Baker, Monya},
  title = {Is There a Reproducibility Crisis?},
  journal = {Nature},
  year = {2016}
}

@article{openscience2015,
  author = {{Open Science Collaboration}},
  title = {Estimating the Reproducibility of Psychological Science},
  journal = {Science},
  year = {2015}
}

@article{begley2012,
  author = {Begley, C. Glenn and Ellis, Lee M.},
  title = {Raise Standards for Preclinical Cancer Research},
  journal = {Nature},
  year = {2012}
}

@article{liu2022stability,
  author = {Liu, T. and Yu, H. and Blair, R.},
  title = {Stability Estimation for Unsupervised Clustering: A Review},
  journal = {WIREs Data Mining and Knowledge Discovery},
  year = {2022}
}

@inproceedings{bendavid2006,
  author = {Ben-David, Shai and von Luxburg, Ulrike and P{\'a}l, D{\'a}vid},
  title = {A Sober Look at Clustering Stability},
  booktitle = {Proceedings of the 19th Annual Conference on Learning Theory},
  year = {2006}
}

@article{vonluxburg2010,
  author = {von Luxburg, Ulrike},
  title = {Clustering Stability: An Overview},
  journal = {Foundations and Trends in Machine Learning},
  year = {2010}
}

@article{hennig2007,
  author = {Hennig, Christian},
  title = {Cluster-Wise Assessment of Cluster Stability},
  journal = {Computational Statistics \& Data Analysis},
  year = {2007}
}

@article{masoero2023,
  author = {Masoero, Lorenzo and Thomas, Elizabeth and Parmigiani, Giovanni and Tyekucheva, Svitlana and Trippa, Lorenzo},
  title = {Cross-Study Replicability in Cluster Analysis},
  journal = {Statistical Science},
  year = {2023}
}

@article{jain1987bootstrap,
  author = {Jain, Anil K. and Moreau, Joseph V.},
  title = {Bootstrap Technique in Cluster Analysis},
  journal = {Pattern Recognition},
  year = {1987}
}

@article{tibshirani2001gap,
  author = {Tibshirani, Robert and Walther, Guenther and Hastie, Trevor},
  title = {Estimating the Number of Clusters in a Data Set via the Gap Statistic},
  journal = {Journal of the Royal Statistical Society: Series B},
  year = {2001}
}

@article{tibshirani2005prediction,
  author = {Tibshirani, Robert and Walther, Guenther},
  title = {Cluster Validation by Prediction Strength},
  journal = {Journal of Computational and Graphical Statistics},
  year = {2005}
}

@article{moeller2006,
  author = {Moeller, Michael and Radke, Daniel},
  title = {Performance of Data Resampling Methods for Robust Class Discovery Based on Clustering},
  journal = {Intelligent Data Analysis},
  year = {2006}
}

@incollection{mucha2015,
  author = {Mucha, Holger and Bartel, Heiko},
  title = {Resampling Techniques in Cluster Analysis: Is Subsampling Better than Bootstrapping?},
  booktitle = {Data Science, Learning by Latent Structures, and Knowledge Discovery},
  year = {2015}
}

@article{dudoit2002,
  author = {Dudoit, Sandrine and Fridlyand, Jane},
  title = {A Prediction-Based Resampling Method for Estimating the Number of Clusters in a Dataset},
  journal = {Genome Biology},
  year = {2002}
}

@inproceedings{abul2003,
  author = {Abul, O. and others},
  title = {Cluster Validity Analysis Using Subsampling},
  booktitle = {IEEE International Conference on Systems, Man and Cybernetics},
  year = {2003}
}

@article{lange2004,
  author = {Lange, Tilman and Roth, Volker and Braun, Mikio and Buhmann, Joachim M.},
  title = {Stability-Based Validation of Clustering Solutions},
  journal = {Neural Computation},
  year = {2004}
}

@article{dangl2020,
  author = {Dangl, Robert and Leisch, Friedrich},
  title = {Effects of Resampling in Determining the Number of Clusters in a Data Set},
  journal = {Journal of Classification},
  year = {2020}
}

@article{jain2010,
  author = {Jain, Anil K.},
  title = {Data Clustering: 50 Years Beyond K-Means},
  journal = {Pattern Recognition Letters},
  year = {2010}
}

@incollection{masulli2015,
  author = {Masulli, Francesco and Rovetta, Stefano},
  title = {Clustering High-Dimensional Data},
  booktitle = {Clustering High-Dimensional Data},
  publisher = {Springer},
  year = {2015}
}

@article{vanderlaan2003,
  author = {van der Laan, Mark J. and Pollard, Katherine S.},
  title = {A New Algorithm for Hybrid Hierarchical Clustering with Visualization and the Bootstrap},
  journal = {Journal of Statistical Planning and Inference},
  year = {2003}
}

@article{hofmans2015,
  author = {Hofmans, Joeri and Ceulemans, Eva and Steinley, Douglas and Van Mechelen, Iven},
  title = {On the Added Value of Bootstrap Analysis for K-Means Clustering},
  journal = {Journal of Classification},
  year = {2015}
}

@article{minaeibidgoli2014,
  author = {Minaei-Bidgoli, Behrouz and others},
  title = {Effects of Resampling Method and Adaptation on Clustering Ensemble Efficacy},
  journal = {Artificial Intelligence Review},
  year = {2014}
}

@article{kimes2017,
  author = {Kimes, Patrick K. and Liu, Yufeng and Hayes, Daniel N. and Marron, J. S.},
  title = {Statistical Significance for Hierarchical Clustering},
  journal = {Biometrics},
  year = {2017}
}

@article{liu2008,
  author = {Liu, Yufeng and Hayes, Daniel N. and Nobel, Andrew and Marron, J. S.},
  title = {Statistical Significance of Clustering for High-Dimension, Low-Sample Size Data},
  journal = {Journal of the American Statistical Association},
  year = {2008}
}

@article{mcshane2002,
  author = {McShane, Lisa M. and others},
  title = {Methods for Assessing Reproducibility of Clustering Patterns Observed in Analyses of Microarray Data},
  journal = {Bioinformatics},
  year = {2002}
}

@article{grabski2023,
  author = {Grabski, Isabella N. and Street, Kelly and Irizarry, Rafael A.},
  title = {Significance Analysis for Clustering with Single-Cell RNA-Sequencing Data},
  journal = {Nature Methods},
  year = {2023}
}

@article{andrews2018,
  author  = {Andrews, Jeffrey L.},
  title   = {Addressing Overfitting and Underfitting in Gaussian Model-Based Clustering},
  journal = {Computational Statistics \& Data Analysis},
  year    = {2018}
}

@article{ghasemian2019,
  author = {Ghasemian, Amir and Hosseinmardi, Homa and Clauset, Aaron},
  title = {Evaluating Overfit and Underfit in Models of Network Community Structure},
  journal = {IEEE Transactions on Knowledge and Data Engineering},
  year = {2019}
}

@article{sugar2003,
  author = {Sugar, Catherine A. and James, Gareth M.},
  title = {Finding the Number of Clusters in a Dataset: An Information-Theoretic Approach},
  journal = {Journal of the American Statistical Association},
  year = {2003}
}

@article{golalipour2021,
  author = {Golalipour, K. and Akbari, E. and Hamidi, S. and Lee, M. and Enayatifar, R.},
  title = {From Clustering to Clustering Ensemble Selection: A Review},
  journal = {Engineering Applications of Artificial Intelligence},
  year = {2021}
}

@article{dalmaijer2022,
  author = {Dalmaijer, Edwin S. and Nord, Camilla L. and Astle, Duncan E.},
  title = {Statistical Power for Cluster Analysis},
  journal = {BMC Bioinformatics},
  year = {2022}
}

@article{haibekains2012,
  author  = {Haibe-Kains, Benjamin and
             Desmedt, Christine and
             Loi, Sherene and
             Culhane, Aedin C. and
             Bontempi, Gianluca and
             Quackenbush, John and
             Sotiriou, Christos},
  title   = {A Three-Gene Model to Robustly Identify Breast Cancer Molecular Subtypes},
  journal = {Journal of the National Cancer Institute},
  year    = {2012}
}

@article{perou2000,
  author = {Perou, Charles M. and others},
  title = {Molecular Portraits of Human Breast Tumours},
  journal = {Nature},
  year = {2000}
}

@article{sorlie2001,
  author = {S{\o}rlie, Therese and others},
  title = {Gene Expression Patterns of Breast Carcinomas Distinguish Tumor Subclasses with Clinical Implications},
  journal = {Proceedings of the National Academy of Sciences},
  year = {2001}
}

@article{parker2009,
  author = {Parker, Joel S. and others},
  title = {Supervised Risk Predictor of Breast Cancer Based on Intrinsic Subtypes},
  journal = {Journal of Clinical Oncology},
  year = {2009}
}

@article{rousseeuw1987,
  author = {Rousseeuw, Peter J.},
  title = {Silhouettes: A Graphical Aid to the Interpretation and Validation of Cluster Analysis},
  journal = {Journal of Computational and Applied Mathematics},
  year = {1987}
}

@article{batool2021,
  author = {Batool, Farhat and Hennig, Christian},
  title = {Clustering with the Average Silhouette Width},
  journal = {Computational Statistics \& Data Analysis},
  year = {2021}
}

@article{hennig2022,
  author = {Hennig, Christian},
  title = {An Empirical Comparison and Characterisation of Nine Popular Clustering Methods},
  journal = {Advances in Data Analysis and Classification},
  year = {2022}
}

@article{smolkin2003,
  author = {Smolkin, Michael and Ghosh, Debashis},
  title = {Cluster Stability Scores for Microarray Data in Cancer Studies},
  journal = {BMC Bioinformatics},
  year = {2003}
}

\end{document}